%% file: main.tex
\documentclass[10pt]{article} %
\usepackage[accepted]{tmlr}

\input{math_commands.tex}

\usepackage{hyperref}
\usepackage{booktabs}
\usepackage[section,below]{placeins}
\usepackage{graphicx}
\usepackage{float}
\usepackage{url}
\usepackage{xcolor}
\usepackage{array}
\usepackage{framed}   %
\newcommand{\rev}[1]{{#1}}


\title{
Is the Geometry Doing the Work?\\
An Operating-Point Audit of Hierarchy in Hyperbolic Vision--Language Models
}

\author{\name Jaeyoung Kim \email jykim@madidt.com \\
      \addr MADI
      \AND
      \name EunSeok Kim \email eskim@madidt.com \\
      \addr MADI
      \AND
      \name Dongsuk Jang \email dsjang@madidt.com \\
      \addr MADI}

\newcommand\blfootnote[1]{%
  \begingroup
  \renewcommand\thefootnote{}%
  \footnotetext{#1}%
  \endgroup
}

\def\month{11}   %
\def\year{2026}  %
\def\openreview{{\upshape\url{https://openreview.net/forum?id=Rj5ewBz62j}}} %

\begin{document}

\maketitle

\blfootnote{We used large language models as general-purpose assistive tools for code debugging, numerical cross-checking, organization of results, and manuscript polishing. All research questions, experimental design choices, statistical procedures, interpretations, and scientific claims are the authors' responsibility.}

\begin{abstract}
Hyperbolic vision--language models are designed to encode abstraction geometrically: general concepts near the origin, specific ones farther out, and entailment cones representing directed order. We ask whether trained MERU, HyCoCLIP, and PHyCLIP models actually use these mechanisms. We audit seven released checkpoints and matched from-scratch interventions, using diagnostics designed to distinguish active hyperbolic geometry from angular structure and supervision-related effects. All audited converged checkpoints remain near-Euclidean in the dimensionless radius $u=\sqrt{c}\rho$, which measures how strongly embeddings experience hyperbolic geometry: the largest observed image-side value is $0.37$---well below $u\approx0.84$, where local metric distortion reaches $10\%$. Releasing the curvature floor changes curvature and norms but not this regime, with mixed, generally modest downstream shifts. Trained entailment cones are saturated or nearly saturated, so low violation rates can arise from trivially wide cones rather than learned order. A preregistered semantic-traversal analysis detects weak order when retrieval is restricted to the correct branch, but no operative readout over the full hierarchy. Shuffle-controlled tests detect no pair-specific radial ordering in released checkpoints, and no positive result is consistent across all three matched ViT-B seeds. \rev{This non-detection is sensitivity-bounded: the diagnostic reaches 80\% power only at ${\approx}13.5$\,pp of shuffle-controlled excess, while the largest released gap is ${\approx}5$\,pp.} Remaining probes find at most small, conditional radial signals that do not establish a usable hierarchy beyond angular structure. We trace this behavior to a low-curvature shortcut: lowering curvature widens entailment cones and suppresses violations without learning order. In the probed training trajectories, gradient decomposition identifies entailment as the dominant curvature-lowering pressure during collapse. Yet curvature continues to contract when entailment is removed, so the shortcut is not the sole cause. Under our diagnostics, the audited formulations do not demonstrate an operative radial or cone-based hierarchy. We distill the audit into a five-number geometry report for evaluating future hierarchy claims.
\end{abstract}

\input{sections/01_Introduction}
\input{sections/02_Related_work}
\input{sections/03_Audit_setup}
\input{sections/04_Q1}
\input{sections/05_Q2}
\input{sections/06_Q3}
\input{sections/07_Reason}
\input{sections/08_Conclusion}

\subsubsection*{Reproducibility Statement}
Our supplementary material provides the measurement scripts, persisted per-run outputs, released-checkpoint SHA-256 hashes, current-GRIT configurations, and geometry convention documents. From-scratch checkpoints are hosted at \url{https://huggingface.co/jaeyoungkim-cs/hyperbolic-vlm-audit}. Quantitative tables can be re-derived directly from the persisted outputs. Regeneration requirements and provenance limits are documented in the supplementary README and provenance notes (Appendix~\ref{app:reproducibility}).

\subsubsection*{Acknowledgments}
We thank the Action Editor and the anonymous reviewers for their careful reading and constructive feedback. This research was supported by the ``Advanced GPU Utilization Support Program'' funded by the Government of the Republic of Korea (Ministry of Science and ICT and Ministry of Health and Welfare).

\bibliography{main}
\bibliographystyle{tmlr}

\appendix
\input{sections/09_Appendix}

\end{document}

%% file: math_commands.tex
\usepackage{amsmath,amsfonts,bm}

\def\eqref#1{equation~\ref{#1}}

\def\1{\bm{1}}

\DeclareMathAlphabet{\mathsfit}{\encodingdefault}{\sfdefault}{m}{sl}
\SetMathAlphabet{\mathsfit}{bold}{\encodingdefault}{\sfdefault}{bx}{n}

%% file: sections/01_Introduction.tex
\section{Introduction}

Vision-language models are increasingly expected to reason not only about visual similarity, but also about abstraction: an image of a ``golden retriever'' is also an image of a ``dog'', an ``animal'', and an ``entity''. This has motivated a growing line of hyperbolic vision-language models, including MERU~\citep{desai2023hyperbolic}, HyCoCLIP~\citep{pal2025compositional}, and PHyCLIP~\citep{yoshikawa2025phyclip}, which replace Euclidean CLIP-style representations~\citep{radford2021learning} with negatively curved spaces. The motivation is compelling. Hyperbolic spaces have exponential volume growth and are known to embed tree-like structures with low distortion~\citep{bridson2013metric,sarkar2011low,nickel2017poincare,ganea2018hyperbolic}. In this view, general concepts should lie closer to the origin, specific concepts should move toward the boundary, and entailment cones should encode asymmetric abstraction relations.

But do current hyperbolic vision-language models actually instantiate this hierarchy mechanism?

This question is more subtle than asking whether hyperbolic VLMs perform well on standard downstream benchmarks. A model can improve retrieval or classification through better contrastive alignment, compositional supervision, calibration, or regularization, without using nonlocal hyperbolic geometry. Conversely, a model can exhibit taxonomy-like semantic similarity in its angular structure without encoding directed radial hierarchy. Existing evaluations often conflate these possibilities: leaf-level classification does not test abstraction, symmetric taxonomy-distance correlations do not distinguish angular similarity from radial depth, and zero entailment violation can arise from saturated cones rather than learned hierarchy.

We audit three hyperbolic VLM families---MERU, HyCoCLIP, and PHyCLIP---across seven released checkpoints and matched current-GRIT interventions. Our findings are fourfold. First, scalar curvature alone does not characterize the audited operating regime. All converged operating points remain near-Euclidean ($H(u)\approx1$), and unclamping alters $c$ and norms with mixed, generally modest downstream shifts (Section~\ref{sec:geometry}). Second, trained-parent apertures are saturated or nearly saturated, \rev{eliminating the aperture's discriminative role, so low violation rates alone do not establish learned order.} Calibrated traversal reveals only weak, branch-conditioned ordering, and the pair-specific radial test detects nothing in released checkpoints and no positive crossing replicated across seeds (Section~\ref{sec:hierarchy}). Third, gradient diagnostics reveal a low-curvature shortcut. On the probed trajectories, entailment drives curvature down at every collapse-phase step and usually dominates the opposing contrastive gradient after objective weighting, yet removing entailment does not prevent collapse, so the shortcut is the dominant accelerator rather than the sole cause (Section~\ref{sec:why_interventions_fail}). Fourth, standard hierarchy evidence is underdetermined unless angular structure and marginal effects are controlled (Section~\ref{sec:evaluation}).

Rather than a new hyperbolic VLM, we contribute a diagnostic framework and a mechanistic account of why published formulations fail to activate the hierarchy mechanism their geometry motivates, in five parts:
\begin{enumerate}
    \item We show that MERU, HyCoCLIP, and PHyCLIP checkpoints operate in a near-Euclidean regime, and that releasing the curvature floor preserves or lowers the image-side dimensionless radius $\sqrt{c}\rho$.
    \item We show that standard downstream metrics are not diagnostic of the entailment/order mechanism. Retrieval, compositionality, and zero-shot scores show mixed shifts under matched comparison while the audited geometry remains near-Euclidean and the cone/radial diagnostics remain non-operative.
    \item We give a mechanistic explanation for curvature collapse. The entailment objective admits a low-curvature shortcut---reducing curvature widens cone apertures and suppresses violations without learning order. A gradient decomposition (in the two gradient-probed families) shows the entailment term drives curvature down more strongly than the counterfactual depth signal (Section~\ref{sec:why_interventions_fail}).
    \item We provide direct hierarchy diagnostics---radial ordering, cone activity, calibrated semantic traversal, angular/radial decomposition, and shuffle-controlled directed tests---that find partial but non-operative signals, not an operative nonlocal hyperbolic hierarchy (Sections~\ref{sec:parent_child} and~\ref{sec:traversal}).
    \item We identify design requirements for hyperbolic VLMs: curvature-identifying supervision, controlled radial representation learning, entailment objectives that do not reward low-curvature wide-cone shortcuts, and evaluation protocols distinguishing angular organization from radial hierarchy.
\end{enumerate}

\begin{center}\fbox{\parbox{0.955\linewidth}{\small
\rev{\textbf{What this paper shows.} Converged MERU, HyCoCLIP, and PHyCLIP checkpoints, including retrainings with relaxed curvature bounds, remain near-Euclidean ($u=\sqrt{c}\rho\le0.37$; exact tangential distortion $\le2.2\%$). Trained text and region-caption cones are saturated or nearly saturated, so low violation alone does not establish learned order. No pair-specific radial order is detected in released models or replicated across all three seeds of any matched base-size condition, and calibrated traversal finds partial within-branch order but no operative full-hierarchy readout. Gradient probes (six runs) and entailment-off controls in MERU and HyCoCLIP identify entailment as the dominant accelerator of collapse, not its sole cause.\par\smallskip
\textbf{What this does not show.} These results do not show that hierarchy is absent: small box-caption and branch-relative norm signals remain, and the radial test's $80\%$ power at a $13.5$-pp effect quantifies sensitivity, not an upper bound on missed effects. Observing box-image endpoints near $u=0.2$ does not prove that the aperture constant $K$ determines where contraction stops, as testing that requires varying $K$ and training to arrest. Finally, strong downstream performance is still possible in this near-Euclidean regime, and our audit does not rule out other hyperbolic formulations, recipes, or data.}}}\end{center}

%% file: sections/02_Related_work.tex
\section{Related Work}

\subsection{Hyperbolic representations and vision-language models}

Hyperbolic geometry is a natural model for hierarchical data. Its exponential volume growth embeds tree-like structures with low distortion~\citep{sarkar2011low}, and Poincaré or Lorentz spaces have been used to embed symbolic taxonomies and partial orders~\citep{nickel2017poincare,ganea2018hyperbolic}, typically associating hierarchy with a radial structure---general concepts near the origin, specific concepts near the boundary---and encoding asymmetric order through entailment-cone aperture and containment~\citep{ganea2018hyperbolic}. Hierarchy is not unique to hyperbolic geometry, however. Order embeddings model entailment as an asymmetric partial order without negative curvature~\citep{vendrov2015order}, and high-dimensional Euclidean embeddings represent WordNet-like trees competitively~\citep{bansal2021comparing}. This motivates the distinction central to our work: apparent semantic hierarchy in a representation does not by itself imply that the model encodes it through nonlocal hyperbolic geometry.

\looseness=-1 Building on these foundations, recent vision-language models encode abstraction in image-text representations. MERU introduces hyperbolic contrastive learning and motivates the radial coordinate as an abstraction axis~\citep{desai2023hyperbolic}; HyCoCLIP adds box-level compositional supervision and intra-/inter-modal entailment losses~\citep{pal2025compositional}; and PHyCLIP factorizes the representation into product hyperbolic components for taxonomic and compositional structure~\citep{yoshikawa2025phyclip}. These models differ in architecture and supervision but share one geometric hypothesis: curvature and entailment cones should induce directed semantic hierarchy. We audit this hypothesis directly---asking not whether they improve benchmarks, but whether the released checkpoints and matched interventions exhibit the advertised mechanism.

\subsection{Task-level analyses of hyperbolic VLMs}

\citet{ibrahimi2024intriguing} analyze released hyperbolic CLIP/MERU-style checkpoints and find improvements over Euclidean CLIP on spatial awareness, ambiguity resolution, out-of-distribution detection, and taxonomy-distance correlation. However, such task-level differences do not by themselves identify the operating geometric mechanism. In particular, symmetric taxonomy-distance correlations can be driven by angular semantic organization rather than radial hierarchy. We return to this in Section~\ref{sec:taxonomy}, where we decompose the taxonomy signal and find no detectable incremental norm-difference contribution beyond angular distance on the evaluated same-level leaf pairs.

\subsection{Euclidean and hierarchy-aware alternatives}

A parallel line of work studies hierarchy in vision-language models without relying on hyperbolic geometry. Order embeddings model image-caption entailment as a partial order~\citep{vendrov2015order}, and hierarchy-aware CLIP variants introduce explicit label or taxonomy supervision in Euclidean representation spaces~\citep{geng2023hiclip}. EuCLIP~\citep{chou2024embedding} further observes that Euclidean CLIP variants can match or outperform hyperbolic alternatives and reports that the learned curvature consistently collapses to its minimum clamp value in hyperbolic VLMs. Our work is complementary but distinct: whereas EuCLIP observes the collapse, we identify a gradient-level mechanism contributing to it---a low-curvature, wide-cone shortcut in the entailment objective---and provide mechanism-targeted diagnostics that test how curvature and radial scale jointly determine the operating regime and how floor release shifts downstream behavior.

\subsection{Angular objectives and concurrent stabilization methods}

Other recent works modify or stabilize the hyperbolic objective itself. \citet{ramasinghe2024accept} propose an angle-based objective (Accept the Modality Gap, ATMG) that preserves a cross-modal modality gap rather than forcing image and text embeddings close in hyperbolic distance, motivated by the concern that geodesic proximity alignment may disrupt latent hierarchical structure. Angular stabilization is not equivalent to radial hierarchy, however: a model can improve angular alignment without establishing nonlocal curvature, parent-child radial ordering, active cone hierarchy, or monotonic traversal. Norm control in hyperbolic space is a recognized training concern---\citet{guo2022clipped} clip large embedding norms to avoid vanishing gradients at large radii---complementary to our finding that the trained operating point instead sits at small $\sqrt{c}\rho$.

Most directly related is the concurrent work ARGENT~\citep{huynh2026argent}, which independently identifies a related entailment-cone instability. Because cone aperture is inversely coupled to the parent norm, a model can minimize parent norms until the aperture degenerates into a half-space, collapsing the intended hierarchy. ARGENT responds with a method---an adaptive entailment loss that removes the norm-to-cone coupling, paired with a norm regularizer---and reports gains over a HyCoCLIP baseline. Our work is complementary along two axes. First, the two analyses reach the same aperture degeneracy by different routes: the half-aperture $\omega \propto \arcsin\!\bigl(2C/(\sqrt{\kappa}\,\lVert\tilde{y}\rVert)\bigr)$ (ARGENT's curvature $\kappa$ is our $c$) depends on the product $\sqrt{\kappa}\,\lVert\tilde{y}\rVert$, and either factor can drive it to $\pi/2$. ARGENT acts on the parent \emph{norm}, whereas we isolate a gradient-level mechanism that drives the \emph{curvature} factor down (Section~\ref{sec:why_interventions_fail}). Second, our contribution is a set of mechanism-targeted diagnostics rather than a stabilized model. Since ARGENT's checkpoints and code were not available at audit time, we do not audit it. Whether a stabilized entailment loss activates nonlocal hyperbolic geometry or merely reorganizes angular structure within a near-Euclidean regime remains open.

%% file: sections/03_Audit_setup.tex
\section{Audit Setup and Diagnostics}
\label{sec:setup}

Our goal is not to propose a new hyperbolic vision-language model, but to audit whether existing hyperbolic VLM formulations instantiate the geometric mechanism that motivates them. We therefore define a mechanism-targeted diagnostic suite for active hyperbolic hierarchy. These diagnostics are not intended to be a complete benchmark for every possible notion of hierarchy. Rather, they test the specific radial and cone-based mechanism that motivates current hyperbolic VLMs: nonlocal curvature should be used, general concepts should lie closer to the origin than specific concepts, entailment cones should encode asymmetric order, and the representation should support operational movement along hierarchy. Table~\ref{tab:claims_map} summarizes each finding together with the check that supports it, the strength of that evidence, and whether the analysis was planned or registered in advance.

\subsection{Released checkpoints and current-GRIT interventions}
\label{sec:released_current}

We separate two sources of evidence throughout the paper. First, we audit released checkpoints as fixed artifacts. This includes public MERU, HyCoCLIP, and PHyCLIP checkpoints. These analyses ask whether the models released by prior work exhibit these geometry and hierarchy mechanisms under our diagnostics.

Second, we perform controlled interventions on models trained from scratch on a fixed snapshot of the Grounded Image--Text Pairs (GRIT) dataset~\citep{peng2024grounding}. GRIT is distributed as URL-referenced web data, and its accessible subset shrinks over time as source links expire and some samples fail to decode. Documented to contain $20.5$M pairs with $35.9$M box annotations, the snapshot we obtained (crawled 2026-04-13) comprises $2{,}051$ shards holding $13.1$M image--text pairs and $25.0$M parent-box annotations (exact counts in Appendix~\ref{app:grit_snapshot}, Table~\ref{tab:grit_snapshot}). We train on all shards for ${\approx}29$ nominal passes over the snapshot ($500$K iterations $\times$ batch $768 = 384$M sample exposures). Our claims rely on matched within-snapshot comparisons, not on reproducing a particular training scale. Baseline and variant models share the current-GRIT snapshot, optimizer, batch size, data order, hyperparameters, and---except for collapse probes---training budget, differing only in the intervention (Appendix~\ref{app:grit_snapshot}; Table~\ref{tab:training_hparams}).

\begin{figure}[t]
  \centering
  \includegraphics[width=0.6\linewidth]{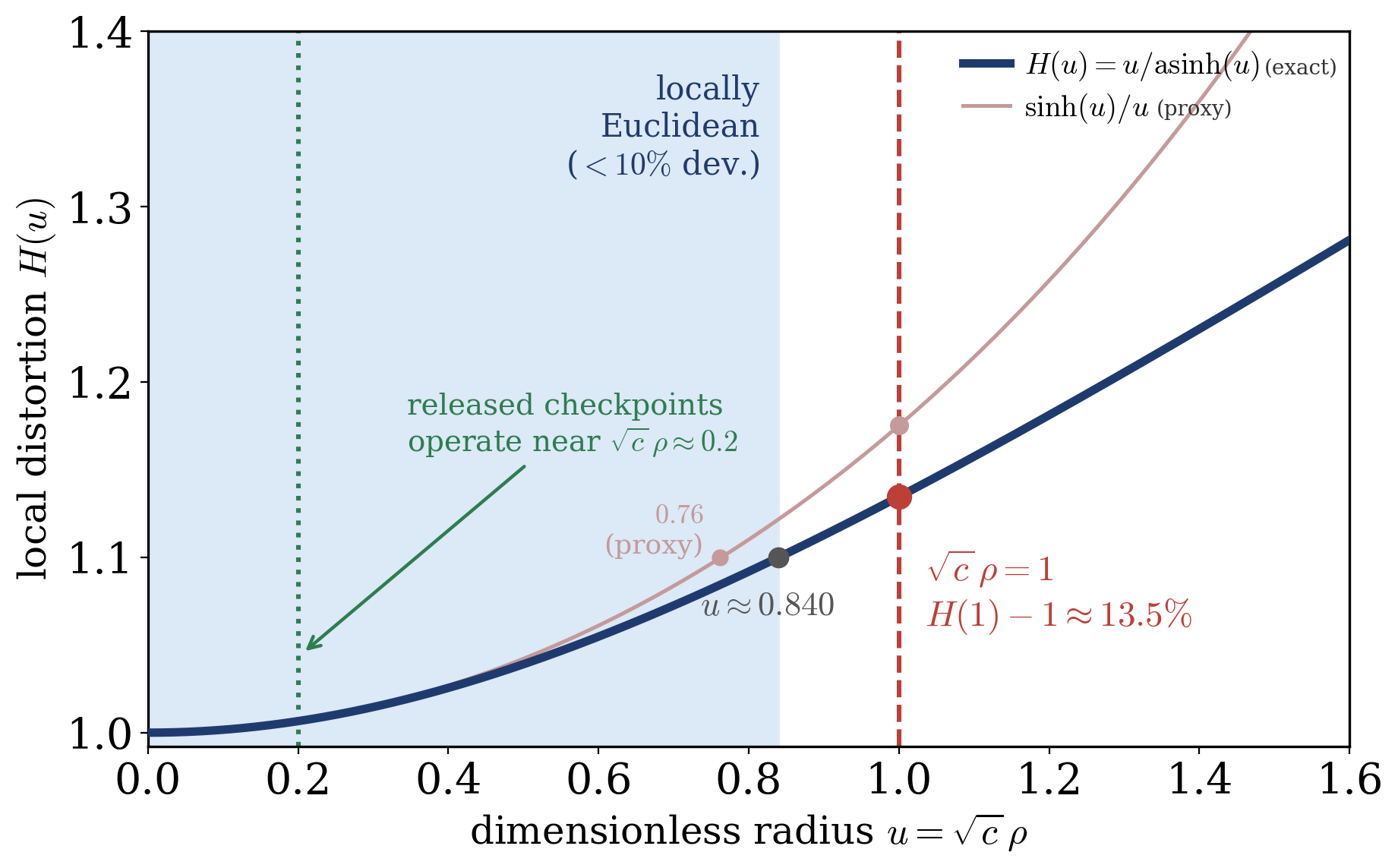}
  \caption{Local distortion factor as a function of the dimensionless spatial
  radius $u=\sqrt{c}\rho$, under the exact convention
  $H(u)=u/\operatorname{asinh}(u)$ and the $\sinh(u)/u$ proxy. The two are
  indistinguishable over the observed checkpoint range ($\sqrt{c}\rho\le0.37$). The shaded
  band is the locally-Euclidean region (deviation $<10\%$), which extends to
  $u\approx0.840$ under the exact convention ($0.76$ under the proxy).
  Released checkpoints operate near $\sqrt{c}\rho\approx0.2$ (dotted vertical),
  and at $\sqrt{c}\rho=1$ (dashed vertical) the stretch is $13.5\%$ exact
  versus $17.5\%$ proxy.}
  \label{fig:hu_curve}
\end{figure}

\subsection{Geometry diagnostics}
\label{sec:geometry_diagnostics}

Curvature alone does not determine whether a representation uses hyperbolic geometry. In Poincaré or Lorentz models, the deviation from local Euclidean behavior depends on the dimensionless product
\begin{equation}
    u = \sqrt{c}\,\rho,
\end{equation}
where $c$ is the curvature parameter and $\rho=\lVert x_{\mathrm{sp}}\rVert_2$ is the Lorentz spatial norm. We write $\sqrt{c}\rho$ throughout. All symbols are collected in Appendix~\ref{app:notation}. We report not only learned curvature, but also radius $\rho$, the product $\sqrt{c}\rho$, and the local distortion factor
\begin{equation}
H(u) \;=\; \frac{\sinh(\hat r)}{\hat r}\bigg|_{\hat r=\operatorname{asinh}(u)} \;=\; \frac{u}{\operatorname{asinh}(u)},
\end{equation}
where $\hat r = \operatorname{asinh}(u)$ is the dimensionless geodesic radius corresponding to the Lorentz spatial norm $\rho$ (so that $\sinh(\hat r) = u$).
$H(u)$ is the factor by which the exponential map at the origin stretches tangential distances at geodesic radius $\hat r$: $H(u) \approx 1$ corresponds to a locally Euclidean regime, while $H(u)$ growing substantially above one signals that the hyperbolic-Euclidean discrepancy becomes order one. We do not interpret $H(u)$ as an estimator of exponential-volume growth. We treat it as a per-sample measure of local metric distortion. Because $H$ is a monotone function of $u=\sqrt{c}\rho$ alone, $H(u)$ and $\sqrt{c}\rho$ are two views of the same quantity, not independent measurements. We report both as the natural unit for different parts of the analysis (the dimensionless radius for the operating point, $H(u)$ for the resulting geodesic deviation). We use $\sqrt{c}\rho > 1$ as a simple marker for whether embeddings reach radii at which this nonlinearity becomes non-negligible. At $u=1$ the stretch exceeds its Euclidean baseline by $H(1)-1 \approx 13.5\%$, and a $10\%$ deviation is reached by $u \approx 0.84$ (Figure~\ref{fig:hu_curve}). The threshold is therefore a conservative marker for substantial nonlinearity, not a discontinuous transition. The proxy $\sinh(u)/u$ treats the spatial coordinate $u$ as the geodesic radius. Its difference from the exact $H(u)$ is $u^{4}/18+O(u^{6})$, below $0.1\%$ for all audited converged checkpoints, whose operating points satisfy $\sqrt{c}\rho\le0.37$. Accordingly, only the final displayed digit can change, with no effect on the conclusions. The exact convention therefore strengthens the near-Euclidean classification.

\rev{A natural question is whether pairwise-distance nonlinearity, rather than radius from the origin, could be geometrically active at $u\approx0.2$--$0.3$. It cannot be large, for the same reason $H(u)\approx1$: for two points within dimensionless radius $u$, the discrepancy between Lorentz geodesic distance and Euclidean distance between spatial components is governed by the same scale, entering at $O(u^{2})$. On the evaluated text and image embeddings of released MERU-B and HyCoCLIP-B, the median ratio of these distances is $0.998$--$0.999$, and the $95$th percentile of the absolute relative deviation is at most $0.26\%$. For illustrative equal-radius pairs separated by $60^\circ$, the signed deviation grows from $-0.17\%$ at $u=0.2$ to $-3.8\%$ at $u=1.0$. The dimensionless radius $\sqrt{c}\rho$ therefore controls both channels. The observed small discrepancy is consistent with the similar behavior of straight-line and exact-geodesic interpolation in the traversal analyses (Appendix~\ref{app:native_radial}).}

These diagnostics distinguish curvature collapse from geometry use. A model may have small curvature and small radius, resulting in locally Euclidean behavior. It may have small curvature but larger radius, preserving the same effective geometry through the product $\sqrt{c}\rho$. Or it may enter a regime where $\sqrt{c}\rho$ is large and $H(u)$ deviates substantially from one, in which case the radial nonlinearity of hyperbolic geometry is functionally engaged. Our claims concern this effective geometry, not curvature as an isolated scalar.

\begin{table}[p]
\centering\small
\caption{\rev{Findings, checks, and limits of the evidence. Claims concern the evaluated models and samples. The planning column preserves the distinction between a written protocol, formal preregistration, and a pair list fixed before the analysis; later sensitivity checks are identified below.}}
\label{tab:claims_map}
\begingroup
\setlength{\tabcolsep}{4pt}
\rev{
\begin{tabular}{@{}>{\raggedright\arraybackslash}p{0.285\linewidth}>{\raggedright\arraybackslash}p{0.285\linewidth}>{\raggedright\arraybackslash}p{0.235\linewidth}>{\raggedright\arraybackslash}p{0.14\linewidth}@{}}
\toprule
Finding & How we checked & Strength and limits & Planning record \\
\midrule
\multicolumn{4}{@{}l}{\emph{Main findings}} \\
\midrule
Audited converged models stay close to Euclidean geometry: local distortion is at most $2.2\%$. & Measure curvature-scaled radius ($u\le0.37$) and local distortion on the evaluated representations. & Direct measurements on the audited samples. & Protocol pre-written \\
\cmidrule(lr){1-4}
Trained text and region-caption cones are almost maximally open. Low violation alone does not show learned order. & Compare cone half-angles with the $90^\circ$ maximum; locate image-region parents relative to the saturation boundary. & Direct measurements in released and retrained models; image-region parents near the boundary with relaxed curvature bounds. & Protocol pre-written \\
\cmidrule(lr){1-4}
Released models show no detectable norm ordering specific to true parent--child pairs; positive results in retrained models do not repeat across all three seeds. & Compare true general--specific class pairs with random re-pairings; add known synthetic signals to check detection. & No detection at the measured sensitivity; weaker effects may remain.\textsuperscript{a} & Protocol pre-written \\
\cmidrule(lr){1-4}
Adding norm to angle gives no detectable improvement in the tested prediction of taxonomy distance. & Predict distance in the concept tree with a linear model of angle, then check the extra contribution of norm. & Sensitivity checked for three released base-size models; this check does not cover retrained models. & Protocol pre-written \\
\cmidrule(lr){1-4}
Tested paths between region and full-image captions show no model-specific advantage in recovering hierarchy. & Compare with retrieval based only on vector length, unrelated target captions, and the Euclidean baseline model. & Descriptive result for the tested pairs, candidate captions, and retrieval rules. & Pre-registered \\
\cmidrule(lr){1-4}
During curvature collapse, the entailment loss pushes curvature downward. & Separate each loss's contribution to curvature updates in six $500{,}000$-step training runs. & A mechanism observed in these runs. & Protocol pre-written \\
\cmidrule(lr){1-4}
Removing the entailment loss slows curvature collapse but does not prevent it. & Retrain two model families without this loss, using three random seeds each. & Result limited to the observed training periods.\textsuperscript{c} & Protocol pre-written \\
\midrule
\multicolumn{4}{@{}l}{\emph{Additional findings with narrower scope}} \\
\midrule
In hyperbolic models, full-image captions tend to have larger norms than their own region captions. & Compare true and randomly re-paired captions while accounting for caption length. & Descriptive result for the fixed pair list. & Pair list fixed pre-run \\
\cmidrule(lr){1-4}
MERU-B shows a small norm difference between general and specific concepts within tree branches. & Shuffle labels within branches after accounting for prompt length. & Difference detected; extra information beyond angle remains unresolved.\textsuperscript{b} & Pre-registered \\
\bottomrule
\end{tabular}
}
\endgroup
\par\smallskip
\begin{minipage}{\linewidth}\footnotesize\rev{\emph{Terms.} An \emph{embedding} is a model's numerical representation; its \emph{norm} is its vector length. A \emph{taxonomy} is a tree of general and specific concepts. A \emph{saturated cone} has the maximum $90^\circ$ half-angle; acceptance still depends on the pair's direction. The \emph{entailment loss} is the training term intended to enforce parent--child order. The radius $u=\sqrt{c}\rho$ combines curvature and vector length (Section~\ref{sec:geometry_diagnostics}).\par\smallskip
\emph{Sensitivity and scope.} \textsuperscript{a}\,A later synthetic check detected a $13.5$-percentage-point signal in $80\%$ of trials; the largest absolute gap in \emph{released} models was about $5$ points. This does not bound undetected effects. \textsuperscript{b}\,The presence test detected signals of $0.8$--$1.1$ standard deviations in $80\%$ of trials. For information beyond angle, detection stayed at or below $28\%$ of trials even for signals of $3$ standard deviations. \textsuperscript{c}\,Training probes ran for $10{,}000$--$11{,}000$ steps; seed~0 was extended to $40{,}000$ (Table~\ref{tab:horizons}).}\end{minipage}
\end{table}

\subsection{Hierarchy diagnostics}
\label{sec:hierarchy_diagnostics}

\looseness=-1 We evaluate three hierarchy notions: lexical taxonomy, visual-semantic granularity, and model-native entailment/order.

\paragraph{Radial ordering.}
Hyperbolic hierarchy is commonly motivated by a radial interpretation: general concepts should lie closer to the origin and specific concepts farther away. For a directed pair $(p,c)$, where $p$ is a parent and $c$ a child, radial consistency is
\begin{equation}
\mathrm{RadialOrder}(p,c)=
\begin{cases}
1, & \rho_c > \rho_p,\\
0, & \text{otherwise},
\end{cases}
\end{equation}
where $\rho_c, \rho_p$ are the child and parent radii and ties are counted as $0$.

\paragraph{Entailment cone activity.}
For models using entailment cones, low violation rate alone is not evidence of hierarchy. It may arise from saturated apertures that make the constraint trivial. We therefore report aperture distributions, active violation rates, directed-pair containment, and, where possible, the gradient contribution of entailment terms (Section~\ref{sec:why_interventions_fail}). When apertures saturate at $\pi/2$, we interpret zero violation as inactive or trivial entailment rather than as successful hierarchy learning.

\paragraph{Traversal robustness.}
If a learned representation supports hierarchy, moving along the proposed radial or geodesic direction should produce graded semantic change. We therefore evaluate traversal trajectories by retrieving nearest captions or labels along interpolation paths. We report monotonicity, perfect traversal rate, collapse rate, and diversity of terminal retrievals as raw statistics. These statistics are descriptive. The primary traversal judgment uses a planted-validated semantic diagnostic evaluated against preregistered effect thresholds (Section~\ref{sec:traversal}; Appendix~\ref{app:traversal_full}).

\paragraph{Product-factor hierarchy.}
For PHyCLIP, our goal is not to evaluate factor disentanglement in general. We ask a narrower question: whether individual factors or the product geometry support directed radial ordering or operative hierarchical traversal. Factor-wise results are therefore interpreted as hierarchy diagnostics, not as a complete analysis of product-factor semantics.

\subsection{Evaluation diagnostics}
\label{sec:evaluation_diagnostics}

Several evaluations can appear hierarchy-sensitive while failing to test the hyperbolic mechanism itself. We therefore separate three sources of signal that hierarchy-looking metrics conflate---angular semantic organization, radial hierarchy, and pair-specific directed structure---and define one diagnostic for each.

\paragraph{Angular/radial taxonomy decomposition.}
A common hierarchy-looking metric is the correlation between taxonomy distance and embedding distance over class labels. This metric is symmetric: it measures whether semantically related labels are close, but does not determine whether hierarchy is encoded radially. To decompose the signal, we compare a cosine-only regression
\begin{equation}
    d_{\mathrm{tree}}(i,j) \sim \beta_0 + \beta_1 d_{\cos}(i,j)
\end{equation}
to a cosine-plus-norm regression
\begin{equation}
    d_{\mathrm{tree}}(i,j)
    \sim
    \beta_0 + \beta_1 d_{\cos}(i,j)
    + \beta_2 \left|\rho_i-\rho_j\right|.
\end{equation}
We report the incremental explanatory power
\begin{equation}
    \Delta R^2_{\mathrm{norm}}
    =
    R^2_{\cos+\mathrm{norm}} - R^2_{\cos}.
\end{equation}
If taxonomy correlation is radial, norm differences should add measurable explanatory power beyond cosine distance. Since pairwise distances are not independent, we use class-level bootstrap or permutation-based procedures rather than treating all class pairs as independent samples.

\paragraph{Shuffle-controlled directed tests.}
Raw directed norm consistency can be confounded by marginal norm distributions or prompt-length effects. For example, if fine-class prompts have slightly larger norms than their CIFAR-defined superclass prompts, many parent-child inequalities may hold even for random pairings. We therefore compare real directed pairs against shuffle-null controls:
\begin{equation}
    \Delta_{\mathrm{pair}}
    =
    \mathrm{Order}_{\mathrm{real}}
    -
    \mathbb{E}_{\mathrm{shuffle}}
    \left[
    \mathrm{Order}_{\mathrm{shuffle}}
    \right].
\end{equation}
A raw ordering score is not interpreted as hierarchy unless it exceeds the shuffle-null distribution. Where prompt length can affect norms, we use length-matched prompts or residualized norms as sensitivity checks. 

As a sanity check, Appendix~\ref{app:positive_controls} applies the same radial and shuffle-controlled diagnostics to a synthetic hyperbolic tree with planted radial and angular hierarchy. The diagnostics recover the planted structure and collapse under radius- or angle-shuffled controls.

\paragraph{Downstream metrics as decoupling probes.}
Retrieval, zero-shot classification, compositional tests, and hierarchy-aware penalties are useful measures of representation quality. However, they do not by themselves establish active hyperbolic hierarchy. We therefore use downstream metrics primarily as decoupling probes: if standard metrics remain comparable or improve while curvature collapses, cones saturate, or radial ordering disappears, then those metrics do not require active hyperbolic hierarchy.

%% file: sections/04_Q1.tex
\section{Curvature Alone Does Not Characterize the Operating Regime}
\label{sec:geometry}
We first ask whether current hyperbolic VLMs operate in a nonlocal hyperbolic regime. The answer is negative across released checkpoints and controlled interventions. Curvature either binds to the published floor or decreases when the floor is released, while the effective geometry remains near-Euclidean.

\subsection{Released checkpoints remain near-Euclidean}
\label{sec:released_geometry}

Across released MERU, HyCoCLIP, and PHyCLIP checkpoints, curvature is at the implementation floor or fixed low-curvature setting, consistent with the curvature collapse reported by~\citet{chou2024embedding}. However, the more important observation is not the scalar value of $c$ alone. The effective operating radius $\sqrt{c}\rho$ remains small across modalities and model families, yielding local distortion factors close to one and zero samples in the nonlocal regime. Curvature and radial scale jointly determine the dimensionless operating coordinate: rescaling $c$ and $\rho$ to preserve $u=\sqrt{c}\rho$ preserves $H(u)$, although absolute geodesic distances retain the global $1/\sqrt{c}$ scale~\citep{sala2018representation,gu2019mixed}. What we add is the empirical finding that trained hyperbolic VLMs \emph{sit} at a near-Euclidean operating point, so it is $\sqrt{c}\rho$, not the scalar $c$, that characterizes them.

\begin{table}[htbp]
\centering
\caption{
Released hyperbolic VLM checkpoints operate in a near-Euclidean regime. To avoid imputing unavailable pooled per-sample distortion summaries, we report directly supported operating-point estimands: the range of per-modality mean $u=\sqrt{c}\rho$ from Table~\ref{tab:released_geometry_full} and the largest observed image-side $u$ from Table~\ref{tab:operating_point_quantiles}. The exact distortion $H(u)=u/\operatorname{asinh}(u)$ is monotone in these same measured coordinates.
}
\label{tab:released_geometry_summary}
\begin{tabular}{lcccc}
\toprule
Model family & Curvature floor & Range of modality-mean $u$ & Max observed image $u$ & \% $u{>}1$ \\
\midrule
MERU & $0.1$ & $[0.173, 0.279]$ & $0.292$ & $0\%$ \\
HyCoCLIP & $0.1$ & $[0.100, 0.201]$ & $0.206$ & $0\%$ \\
PHyCLIP & $0.1$ (per-subspace) & $[0.114, 0.218]$ & $0.367$ & $0\%$ \\
\bottomrule
\end{tabular}
\end{table}

Across all released checkpoints, median local distortion is near one, and the maximum image-side $\sqrt{c}\rho$ is at most $0.37$---far below both the conservative $0.76$ marker and the exact $10\%$-distortion point at $0.84$ (Appendix~\ref{app:operating_point_quantiles}, Table~\ref{tab:operating_point_quantiles}). Thus, no released representation operates in the nonlocal regime ($\sqrt{c}\rho>1$). \rev{The largest observed per-sample distortion across every audited converged checkpoint is $2.17\%$ (at $u=0.3674$), and the largest checkpoint median incurs $1.45\%$, so the criterion would need to tighten below ${\sim}2\%$ before even one sample crosses (Figure~\ref{fig:common_u_axis}).}

Together, these diagnostics limit the interpretation of curvature collapse. 
The issue is not merely that $c$ is small or clamped: the learned embeddings also remain at small dimensionless radius, so the effective operating geometry is close to Euclidean. Notably, the from-scratch current-GRIT baselines converge to the same dimensionless radius as the released checkpoints ($\sqrt{c}\rho \approx 0.20$ for HyCoCLIP at both scales; Tables~\ref{tab:released_geometry_summary}--\ref{tab:clampoff_geometry}), despite differing in initialization and data availability. The near-Euclidean operating point is thus a recurring empirical regime across these runs rather than an artifact of any single one.

\subsection{Unclamping curvature does not change the operating regime}
\label{sec:clampoff_geometry}

To test whether the curvature floor is suppressing a useful geometric degree of freedom, we train matched current-GRIT baselines and curvature-unclamped variants. Here ``baseline'' denotes our numerically stabilized reproduction on the current-GRIT snapshot using the published hyperparameters. All baseline and clampOff runs use the explicit fp32 Lorentz patch documented in Appendix~\ref{app:reproducibility}. Within this shared implementation, baseline and clampOff differ only in the curvature clamp. In the clampOff variants, the lower curvature clamp is relaxed from $0.1$ to $0.001$, while all other training settings are kept fixed. Controlled interventions are run at ViT-S and ViT-B. ViT-L is audited only at the released-checkpoint level (Table~\ref{tab:released_geometry_full}).

\begin{table}[htbp]
\centering
\caption{
Curvature unclamping changes $c$ and radial norms but leaves every run in the near-Euclidean operating regime: $\sqrt{c}\rho$ stays in the $\approx0.2$--$0.3$ band and $H(u)$ near $1$, so the nonlocal-regime fraction is $0\%$ for all. ViT-B rows are mean\,$\pm$\,s.d.\ over three seeds ($0,37,42$). ViT-S rows are single-seed (seed $0$). Per-seed values are in Table~\ref{tab:clampoff_geometry_perseed}; measured on $500$ ImageNet~\citep{deng2009imagenet,russakovsky2015imagenet} and $500$ COCO~\citep{lin2014microsoft} val images (image-side).
}
\label{tab:clampoff_geometry}
\setlength{\tabcolsep}{4pt}
\small
\begin{tabular}{l cc cc cc cc}
\toprule
& \multicolumn{2}{c}{$c$} & \multicolumn{2}{c}{$\rho$} & \multicolumn{2}{c}{$\sqrt{c}\,\rho$} & \multicolumn{2}{c}{$H(u)$} \\
\cmidrule(lr){2-3}\cmidrule(lr){4-5}\cmidrule(lr){6-7}\cmidrule(lr){8-9}
Model & baseline & clampOff & baseline & clampOff & baseline & clampOff & baseline & clampOff \\
\midrule
MERU-S     & $0.1000$ & $0.0419$ & $0.921$ & $1.052$ & 0.291 & 0.215 & 1.014 & 1.008 \\
MERU-B     & $0.1000$ & $0.0289{\scriptstyle\pm.0007}$ & $0.939{\scriptstyle\pm.004}$ & $1.049{\scriptstyle\pm.003}$ & $0.297{\scriptstyle\pm.001}$ & $0.178{\scriptstyle\pm.002}$ & $1.014{\scriptstyle\pm.000}$ & $1.005{\scriptstyle\pm.000}$ \\
HyCoCLIP-S & $0.1000$ & $0.0112$ & $0.635$ & $1.839$ & 0.201 & 0.194 & 1.007 & 1.006 \\
HyCoCLIP-B & $0.1000$ & $0.0093{\scriptstyle\pm.0003}$ & $0.638{\scriptstyle\pm.000}$ & $2.019{\scriptstyle\pm.030}$ & $0.202{\scriptstyle\pm.000}$ & $0.195{\scriptstyle\pm.000}$ & $1.007{\scriptstyle\pm.000}$ & $1.006{\scriptstyle\pm.000}$ \\
PHyCLIP-S  & $0.1000$ & $0.0155$ & $0.644$ & $1.508$ & 0.204 & 0.188 & 1.007 & 1.006 \\
PHyCLIP-B  & $0.1000$ & $0.0144{\scriptstyle\pm.0001}$ & $0.650{\scriptstyle\pm.000}$ & $1.562{\scriptstyle\pm.008}$ & $0.206{\scriptstyle\pm.000}$ & $0.187{\scriptstyle\pm.000}$ & $1.007{\scriptstyle\pm.000}$ & $1.006{\scriptstyle\pm.000}$ \\
\bottomrule
\end{tabular}
\end{table}

\begin{figure}[t]
  \centering
  \includegraphics[width=0.75\linewidth]{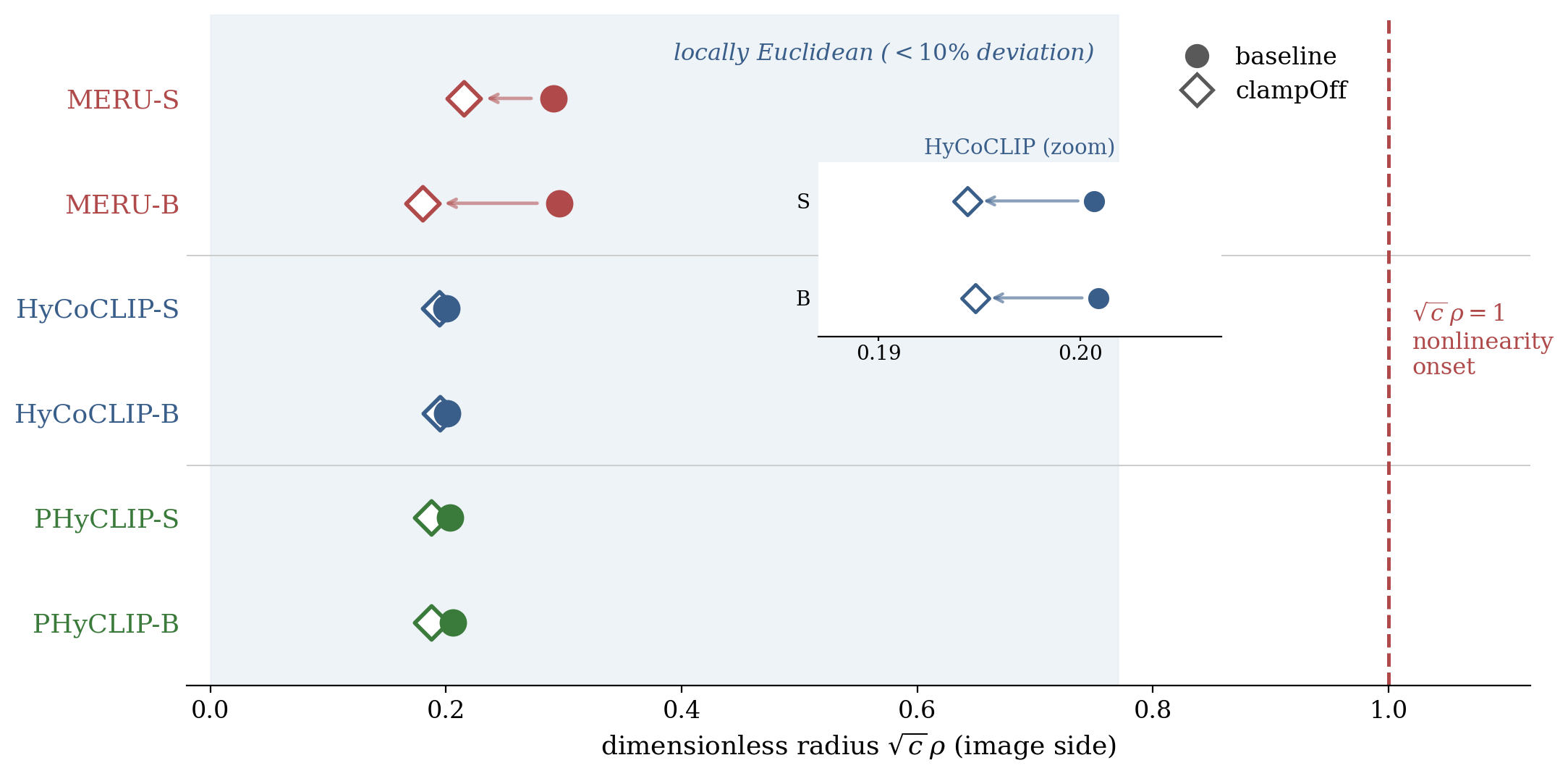}
  \caption{
  Dimensionless radius $\sqrt{c}\rho$ for current-GRIT baseline
  (filled circles) and clampOff (open diamonds) across the three families and
  both ViT sizes (image side; seed-$0$ values from Table~\ref{tab:clampoff_geometry_perseed}). Arrows trace
  the baseline$\rightarrow$clampOff shift. Every model sits at
  $\sqrt{c}\rho\approx0.2$--$0.3$---deep inside the locally-Euclidean region (shaded)
  and far from the threshold $\sqrt{c}\rho=1$ (dashed). HyCoCLIP barely moves (inset), MERU drops
  further, and PHyCLIP lies between.
  }
  \label{fig:radius_scatter}
\end{figure}

Table~\ref{tab:clampoff_geometry} shows the central result. Releasing the curvature floor changes $c$ and embedding norms, but does not move the models into a nonlocal hyperbolic regime. The effect on the dimensionless radius is family-dependent. In HyCoCLIP, $c$ decreases by roughly an order of magnitude and norms increase, preserving $\sqrt{c}\rho$ to within about $3.5\%$. In MERU, the compensation is incomplete and unclamping moves the model to an even lower dimensionless radius. PHyCLIP falls between the two. In every case, however, the model stays well inside the near-Euclidean regime: across all families and sizes in these current-GRIT runs the local distortion factor remains at or below $1.015$ and the nonlocal-regime fraction remains $0\%$ (Figure~\ref{fig:radius_scatter}).

\subsection{Downstream shifts are mixed but generally modest}
\label{sec:downstream_decoupling}

Relaxing the curvature floor tests whether downstream behavior tracks the scalar curvature change. The evaluated metrics show mixed, generally modest shifts while the operating geometry remains near-Euclidean.

\begin{table}[htbp]
\centering
\caption{
Current-GRIT matched comparisons (clampOff minus baseline), averaged over three seeds ($0$, $37$, $42$) and reported as mean$\,\pm\,$standard deviation. \rev{Statistics are calculated before rounding.} Retrieval deltas (COCO, Flickr; R@5) are percentage points. ZSC is zero-shot \emph{ImageNet} top-1 accuracy (percentage points). SC is SugarCrepe overall accuracy in points ($0$--$100$). For the WordNet hierarchical metrics, TIE and LCA are distances (lower is better), while J, PH, and RH are similarity/precision/recall scores on a $0$--$1$ scale (higher is better). All hierarchical deltas are within a few thousandths to about a tenth (TIE is largest, at $+0.084$); for each metric, the three family-level mean deltas have the same sign. Per-seed absolute values are in Tables~\ref{tab:retrieval_hier_full}, \ref{tab:zsc_full}, and \ref{tab:compositional_full}.
}
\label{tab:downstream_delta}
\setlength{\tabcolsep}{4pt}
\resizebox{\textwidth}{!}{%
\begin{tabular}{l cccc c c c cc cc}
\toprule
& \multicolumn{6}{c}{Downstream ($\Delta$)} & \multicolumn{5}{c}{WordNet hierarchy ($\Delta$)} \\
\cmidrule(lr){2-7}\cmidrule(lr){8-12}
Model & T2I COCO & I2T COCO & T2I Flk & I2T Flk & ZSC IN & SC & TIE$\downarrow$ & LCA$\downarrow$ & J$\uparrow$ & PH$\uparrow$ & RH$\uparrow$ \\
\midrule
MERU-B & $-0.04{\scriptstyle\pm0.48}$ & $-0.47{\scriptstyle\pm0.23}$ & $-0.05{\scriptstyle\pm0.17}$ & $+0.00{\scriptstyle\pm0.89}$ & $\rev{-0.46{\scriptstyle\pm0.23}}$ & $\rev{+0.12{\scriptstyle\pm0.13}}$ & $+0.049{\scriptstyle\pm0.039}$ & $\rev{+0.008{\scriptstyle\pm0.021}}$ & $\rev{-0.0038{\scriptstyle\pm0.0026}}$ & $\rev{-0.0016{\scriptstyle\pm0.0018}}$ & $-0.0036{\scriptstyle\pm0.0024}$ \\
HyCoCLIP-B & $+0.96{\scriptstyle\pm0.63}$ & $+1.60{\scriptstyle\pm0.43}$ & $+0.72{\scriptstyle\pm0.40}$ & $+1.00{\scriptstyle\pm0.56}$ & $-0.35{\scriptstyle\pm0.29}$ & $\rev{+1.36{\scriptstyle\pm0.21}}$ & $\rev{+0.084{\scriptstyle\pm0.026}}$ & $\rev{+0.029{\scriptstyle\pm0.027}}$ & $\rev{-0.0053{\scriptstyle\pm0.0015}}$ & $\rev{-0.0033{\scriptstyle\pm0.0020}}$ & $-0.0053{\scriptstyle\pm0.0012}$ \\
PHyCLIP-B & $+0.64{\scriptstyle\pm0.22}$ & $+0.63{\scriptstyle\pm0.48}$ & $+0.43{\scriptstyle\pm0.35}$ & $+0.17{\scriptstyle\pm0.65}$ & $-0.29{\scriptstyle\pm0.61}$ & $\rev{+0.76{\scriptstyle\pm0.80}}$ & $+0.049{\scriptstyle\pm0.030}$ & $+0.003{\scriptstyle\pm0.022}$ & $\rev{-0.0040{\scriptstyle\pm0.0016}}$ & $\rev{-0.0018{\scriptstyle\pm0.0016}}$ & $-0.0040{\scriptstyle\pm0.0009}$ \\
\bottomrule
\end{tabular}%
}
\end{table}

Table~\ref{tab:downstream_delta} reports current-GRIT matched comparisons for MERU-B, HyCoCLIP-B, and PHyCLIP-B. The positive HyCoCLIP-B deltas are consistent across all three seeds, but we interpret them as decoupling evidence: the evaluated metrics show mixed, generally modest shifts as curvature collapses. For HyCoCLIP-B clampOff, curvature drops by an order of magnitude to $c\approx0.009$ and the representation remains near-Euclidean, yet COCO and Flickr~\citep{young2014image} retrieval and SugarCrepe~\citep{hsieh2023sugarcrepe} move in the positive direction relative to baseline, while ImageNet zero-shot ($-0.35{\scriptstyle\pm0.29}$) and VL-Checklist~\citep{zhao2022vlchecklist} shift within the seed spread. (VL-Checklist is seed-variable, with three-seed subtype means of $-0.5$ to $-4.3$ points.) Four of the five WordNet~\citep{miller1995wordnet} hierarchical metrics (TIE, J, PH, RH) shift by small but consistent-sign amounts. For MERU-B these metrics show only small shifts ($\leq0.5$ points, some sign-consistent) across the three matched seeds. PHyCLIP-B likewise shows only small shifts under clampOff, with retrieval and SugarCrepe moving in the positive direction, the ImageNet metric within the seed spread, and four of the five WordNet hierarchy metrics (TIE, J, PH, RH) showing small, consistent-sign shifts across seeds. Across all three families, scalar-curvature collapse is accompanied by mixed, generally modest downstream shifts in this matched setting.

Released checkpoints often obtain higher absolute downstream scores than current-GRIT baselines. We treat them as historical references only, since they may differ in data availability.

Together, these results show that scalar curvature alone does not characterize the measured operating regime: across families, sizes, and floor settings, $u=\sqrt{c}\rho$ stays in the same near-Euclidean range ($\approx 0.2$--$0.3$), whether the curvature change is absorbed by compensating norm growth (HyCoCLIP, which nearly preserves $\sqrt{c}\rho$) or only partially compensated (MERU, where $\sqrt{c}\rho$ falls further; Tables~\ref{tab:released_geometry_summary}--\ref{tab:clampoff_geometry}). This is the first of two distinct failures. Curvature is not the active geometric resource in any audited converged configuration, and no converged run remains in the nonlinear hyperbolic regime. This failure does not by itself imply that hierarchy is absent, because radial depth ordering can in principle be expressed even in a near-Euclidean space~\citep{vendrov2015order,bansal2021comparing}. The two failures are \emph{logically separable}.

We therefore ask, in the next section, whether the hierarchy mechanism the geometry motivates appears anywhere in the representation---in radial ordering, cone activity, or traversal---regardless of curvature.

%% file: sections/05_Q2.tex
\section{Direct Diagnostics Do Not Detect an Operative Hierarchy}
\label{sec:hierarchy}

We distinguish lexical hierarchy, model-native cone/order structure, and traversal. Across these diagnostics, no audited formulation shows evidence of an operative, graded directed radial or cone-based hierarchy.

\begin{figure}[t]
  \centering
  \includegraphics[width=0.82\linewidth]{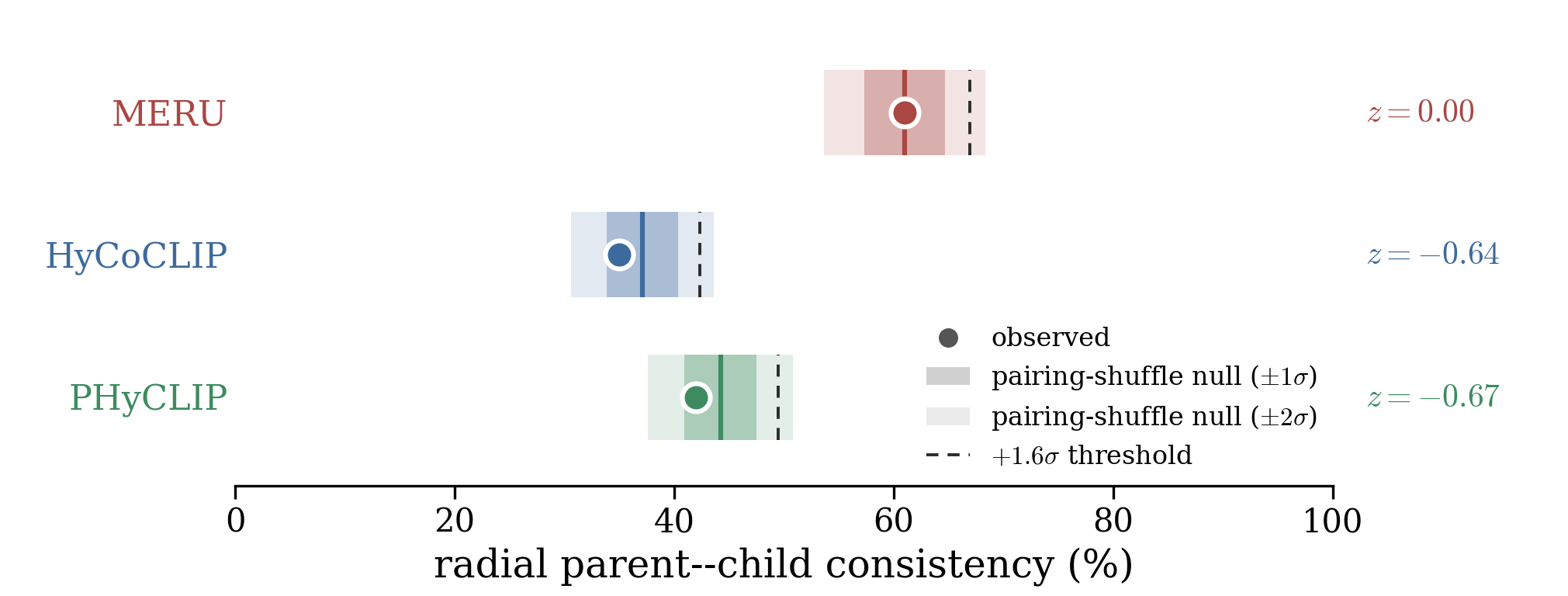}
  \caption{Radial parent--child consistency for released ViT-B checkpoints against each model's
$10{,}000$-permutation pairing-shuffle null ($\pm1\sigma$ and $\pm2\sigma$ bands;
null mean as a vertical tick; one-sided $+1.6\sigma$ detection threshold dashed).
The three hyperbolic models lie within $\pm1\sigma$ of their nulls.
Euclidean CLIP-B is omitted because its released embeddings are unit-normalized,
making norm ordering tie-degenerate rather than a valid radial control
(Table~\ref{tab:parent_child_full}). Raw consistencies span $35$--$61\%$;
none exceeds its null.}
  \label{fig:radial}
\end{figure}

\subsection{Pair-specific radial order is not detected; native signals are non-operative}
\label{sec:parent_child}

Hyperbolic VLMs are motivated by the radial-depth hypothesis that general concepts should be closer to the origin and specific concepts farther away. We evaluate this hypothesis using directed parent--child pairs given by the CIFAR-100~\citep{krizhevsky2009learning} fine$\to$coarse label hierarchy. Each fine class is paired with its CIFAR-defined superclass, yielding 100 (parent$=$coarse, child$=$fine) pairs. For each pair, we ask whether the child embedding has larger norm than the parent embedding, and we compare the resulting consistency against a shuffle-null distribution over randomly permuted pairings, following the criterion of Section~\ref{sec:evaluation_diagnostics}.

A raw norm-ordering rate can be driven by a \emph{marginal} effect, such as prompt length or a global specific-versus-general norm offset, rather than by any learned parent--child relation. The shuffle null removes this marginal component, so the primary estimand is the \emph{pair-specific} excess. Do true pairings carry more radial order than exchangeable pairings with the same norm marginals? This scope is deliberately narrower than radial depth in general. Non-detections do not rule out a purely level-wise or branch-relative norm code.

Across the released ViT-B checkpoints, all three hyperbolic models lie within $1\sigma$ of their pairing-shuffle nulls and satisfy $z<+1.6$ (Figure~\ref{fig:radial}). Euclidean CLIP-B is N/A because unit normalization makes norm ordering tie-degenerate. \rev{Recovering pre-normalization CLIP features (\texttt{project=False}) removes this degeneracy. Across the three released CLIP scales the unnormalized Euclidean control shows directed agreement of $51.0\%$ ($z=+0.71$, ViT-S), $49.0\%$ ($z=+1.82$, ViT-B), and $44.0\%$ ($z=-1.14$, ViT-L). One scale thus crosses the $+1.6$ screen on its own, showing that isolated crossings of this magnitude do not establish hyperbolic radial order. Its taxonomy-side $\Delta R^{2}_{\mathrm{norm}}$ under the same decomposition is $+0.0035$--$+0.0063$ (Mantel $p=0.075$--$0.210$).} All other released scales are likewise non-detections, and the largest absolute gap is about $5$ pp for PHyCLIP-L ($z=-1.59$, reverse direction), with the others within about $2$ pp. In a post-hoc design-matched positive control, the smallest tested planted excess attaining $80\%$ power was $\Delta_{\mathrm{pair}}\approx13.5$ pp (Appendix~\ref{app:sensitivity}). Current-GRIT cells cross $\pm1.6$ in both directions, but no positive crossing is replicated across all three seeds of any ViT-B condition. The largest positive excursion is the unreplicated PHyCLIP-S clampOff run ($9.5$ pp, $z=+2.73$). Separately, calibrated semantic traversal detects partial branch-conditioned ordering but no operative full-hierarchy readout across four confirmatory checkpoints (Section~\ref{sec:traversal}). Full radial results are in Table~\ref{tab:parent_child_full}.

This does not mean the models lack semantic structure. They may still encode angular similarity, with sibling or related concepts close in angle.

A complementary branch-relative test on released ViT-B checkpoints, which the pair-shuffle diagnostic does not adjudicate, detects a small level-related norm signal in MERU-B across $20$ CIFAR superclasses after residualizing token and character length (median $D_b=+0.0147$, Holm-adjusted $p=0.0003$). HyCoCLIP-B and PHyCLIP-B are non-detections ($p=0.605$ and $0.223$). Planted controls give the presence test $80\%$ power for effects of $0.8$--$1.1$ standard deviations, but the unique-beyond-angle comparison remains underpowered (power $\leq0.28$ even at $3$ SD). Thus MERU-B contains detectable branch-relative radial information, but whether it adds information beyond angle remains unresolved and it does not establish an operative radial hierarchy (Appendix~\ref{app:sensitivity}).

\paragraph{A box-family-native relation test partially qualifies this result.}
The external CIFAR-100 taxonomy is not the relation the box-supervised models were trained on, so we repeat the directed radial test on the native GRIT box$\to$full caption structure (hereafter NG2). Each sample pairs a full-image caption (the more specific concept, predicted to have \emph{larger} norm) with a box-level caption (the more general concept, predicted smaller), following HyCoCLIP's compositional convention~\citep{pal2025compositional}. The directed shuffle test uses all $935$ non-identical-caption pairs ($R=500$). The beyond-cosine Mantel analysis uses the first $500$ registry pairs (Table~\ref{tab:native_radial}). Both preserve the full/box norm marginals. We also use length-matched and length-residualized variants. These are fixed-registry conditional diagnostics whose end-to-end size and power are not calibrated.

After these controls, angular distance remains the dominant predictor in every hyperbolic model, and the reported linear radial supplement beyond cosine is numerically small. On the released ViT-B checkpoints $\Delta R^2_{\mathrm{norm}}\le0.00025$ (largest MERU-B; PHyCLIP-B at $0.00001$, $p=0.057$). \rev{Under the clampOff checkpoints it is larger but seed-variable. Per-seed spans and Mantel $p$-values are consolidated in Table~\ref{tab:native_radial} (Appendix).} The conditional directed one-bit excess is consistently positive (full-set $z$ between $+4.7$ and $+6.4$), whereas the unique linear norm increment after cosine varies with seed. \rev{An unnormalized Euclidean CLIP control (all three released scales) on the same registry is strongly reversed ($z$ from $-8.6$ to $-13.0$: specific full captions carry smaller norms), opposite in sign to the box-supervised models, so the positive directed bit is not a caption-length or generic-encoder artifact.} This diagnostic does not adjudicate radial information redundant with angle. Even the largest supplement ($0.00228$) corresponds to only $0.23$ percentage points of additional variance explained in the binary same-pair indicator.

Because the released-versus-clampOff contrast is cross-sectional and the directed $z$ records only a single conditional bit (MERU-B shows it too, $z=+5.3$, without box supervision), neither the one-bit excess nor the numerically small linear increment establishes an operative graded axis (Appendix~\ref{app:native_radial}).

We also examine NG2 with a preregistered grid crossing radial-ray and geodesic interpolation with cosine and model-native retrieval, including a norm-only control (exact Lorentz geodesics for the hyperbolic models, the configured product distance for PHyCLIP, and a straight Euclidean segment for CLIP; Appendix~\ref{app:native_radial}, Tables~\ref{tab:native_traversal} and~\ref{tab:native_graded}). Radial$\times$cosine is identically zero by norm invariance, hence not model evidence. On this fixed registry and retrieval pool, no hyperbolic-model paired-bootstrap interval for the native-distance--minus--norm gain has a positive lower bound (closest: MERU-B released, $[-0.008,+0.053]$), a descriptive conditional non-detection that cannot separate a learned radial code redundant with angular structure from a norm--specificity association in the pool. Euclidean CLIP's norm-only and gain cells are N/A. Geodesic correlations are positive, but mismatched targets remain above the shuffle null and Euclidean CLIP attains the largest value ($0.775$), indicating generic interpolation rather than model-specific hierarchy.

\subsection{Entailment cones are inactive or saturated}
\label{sec:cones}

Entailment cones are intended to encode asymmetric semantic order. However, the observed cone behavior is fragile. Under the default threshold, some trained entailment terms are almost trivially satisfied and provide little effective gradient. The text-side cone is already saturated at $\pi/2$ at baseline in every family, so the text$\to$image violation rate is low ($0$--$0.4\%$ for the current-GRIT HyCoCLIP baselines, $2.7$--$5.1\%$ for MERU; $9$--$14\%$ for PHyCLIP at baseline, where larger image norms place some embeddings just outside the saturated cone). Under clampOff, the PHyCLIP rate collapses ($\le0.2\%$), alongside contraction in $u$ and compression of the text--image angle distribution, while MERU's stays in low single digits and HyCoCLIP's at zero. The diagnostic full-image aperture often saturates too---with MERU near the saturation edge as the exception discussed below. The reverse image$\to$text violation stays near $100\%$ across these aperture regimes, as expected from the measured inner-text/outer-image ordering. No model trains a full-image-parent relation, and a working directed hierarchy would produce this rate just as a modality norm offset would, so we report it as a directionality check rather than evidence for or against hierarchy. The informative reading is on the trained side. Low violation under a saturated parent cone is near-trivial containment, not learned order. When the entailment threshold is tightened, trained-direction diagnostic violation, evaluated at the exact cone boundary ($\eta=1$), rises only to $1.17\%$ for ViT-B and $0.78\%$ for ViT-S, while diagnostic image-cone saturation increases and the operating geometry remains unchanged. The accompanying raw nearest-neighbor traversal statistics vary across scales and pools and are descriptive (Appendix~\ref{sec:eta_intervention}).

\begin{table}[htbp]
\centering
\caption{
A saturated text-side cone makes the trained text$\to$image constraint near-trivial (low t$\to$i violation). The reverse image$\to$text rate stays near $100\%$, as the inner-text/outer-image ordering predicts. Image apertures are diagnostic full-image-parent apertures; trained box-image parents are analyzed in Section~\ref{sec:entailment_shortcut}. All apertures use seed-$0$ checkpoints and the same $256$-sample GRIT batch as Table~\ref{tab:cone_summary_full}.
}
\label{tab:cone_summary}
\small
\begin{tabular}{lp{2.3cm}p{2.3cm}p{7cm}}
\toprule
Model & Image aperture & Text aperture & Interpretation \\
\midrule
MERU-B baseline
& $0.73$ rad
& $\pi/2$ saturated
& Diagnostic image cone non-saturated; trained text-parent cone saturated. \\

MERU-B clampOff
& $\pi/2$ saturated
& $\pi/2$ saturated
& Both saturate; the diagnostic image side sits at $\sqrt{c}\rho$ just below $2K$ at all three seeds (Section~\ref{sec:entailment_shortcut}). \\

MERU-S clampOff
& $1.175$ rad
& $\pi/2$ saturated
& Image-side $\sqrt{c}\rho=0.217>2K$: non-saturated, consistent with the edge criterion (Section~\ref{sec:entailment_shortcut}). \\

HyCoCLIP-B baseline
& $1.42$ rad
& $\pi/2$ saturated
& Diagnostic image cone non-saturated; trained text-parent cone saturated. \\

HyCoCLIP-B clampOff
& $\pi/2$ saturated
& $\pi/2$ saturated
& Both saturate; trained t$\to$i is near-trivial and reverse i$\to$t violation is expected. \\
\bottomrule
\end{tabular}
\end{table}

Table~\ref{tab:cone_summary} summarizes representative cone behavior. In HyCoCLIP, the trained text-side cone is saturated already at baseline (text$\to$image violation $0\%$) and stays saturated under clampOff, so the near-trivial constraint is the baseline condition rather than a clampOff-induced drop. The diagnostic image aperture also saturates under clampOff ($11\%\to100\%$). MERU's diagnostic image apertures are non-saturated at baseline ($0.73$ rad). Under clampOff, MERU-B saturates to $\pi/2$ at all three seeds, whereas MERU-S remains non-saturated ($1.175$ rad). MERU's trained text-parent cone remains saturated throughout. On the trained relations, zero violation under a saturated parent cone is near-trivial containment, not evidence of learned order. Section~\ref{sec:entailment_shortcut} links this behavior to a low-curvature shortcut that widens apertures and suppresses violations without learning semantic order.

The same trained-side pattern appears in PHyCLIP. Its text-parent cone is already saturated, and clampOff compresses the text--image angle distribution, reducing text$\to$image violation per size (PHyCLIP-S $13.8\%\to0.2\%$, PHyCLIP-B $9.5\%\to0.05\%$; Appendix~\ref{app:cone_full}). The simultaneous diagnostic image-aperture saturation ($41$--$45\%\to{\approx}96\%$) is a parallel symptom of curvature collapse, not the cause of the forward-rate drop. Together, these observations extend the aperture-saturation signature across all three families, although the per-loss gradient decomposition is available only for MERU and HyCoCLIP (Section~\ref{sec:entailment_shortcut}).\footnote{Every PHyCLIP factor shows a nonzero image--text norm separation (Cohen's $d \geq 0.2$). This establishes modality separation, not hierarchy relevance. A modality norm gap is not parent--child order. PHyCLIP's directed radial consistency does not exceed its pairing-shuffle null (Section~\ref{sec:parent_child}).} Full per-model diagnostics are in Appendix~\ref{app:cone_full}.

\rev{For fully saturated parent cones, aperture width no longer separates satisfied from violated pairs, and containment is decided by each pair's exterior angle relative to the fixed threshold. Containment can nevertheless remain sensitive to pair identity. On the released and baseline checkpoints, matched pairs satisfy the $\eta=1.0$ containment condition at far higher rates than shuffled pairs ($\Delta_{\mathrm{cone}}$ $+0.24$--$+0.97$; Table~\ref{tab:delta_cone}, Appendix~\ref{app:delta_cone}), except in the two box-image cells where both rates sit at floor (${\approx}0$). Under clampOff, two further cells instead collapse at ceiling (matched $=$ shuffled $=100\%$). Neither low trained-side violation rates nor a positive matched--shuffled gap alone establishes learned entailment. Positive gaps show sensitivity to pairing, but do not distinguish angular alignment from hierarchical order (Section~\ref{sec:evaluation_diagnostics}).}

\begin{figure}[!htbp]
\setlength{\parskip}{0pt}
\setlength{\abovecaptionskip}{3pt}
\centering
\includegraphics[width=\linewidth]{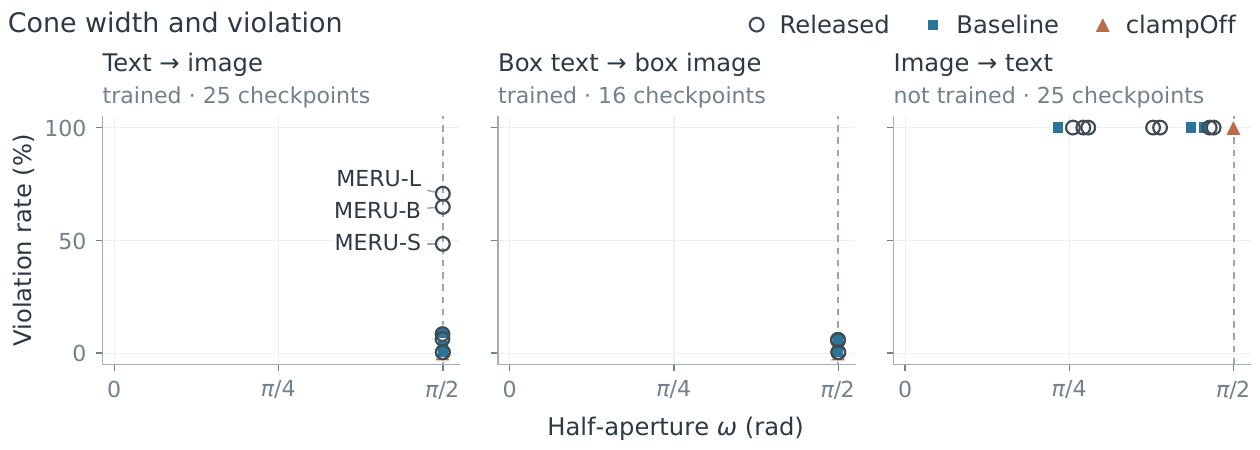}
\vspace{-12pt}
\caption{\rev{Median parent half-aperture versus violation rate ($\eta=1$), grouped by relation. Each point represents one checkpoint, with training seeds counted separately. Dashed lines mark $\pi/2$. Marker styles denote checkpoint settings.}}
\label{fig:cone_width_violation}
\end{figure}

\subsection{Calibrated traversal detects partial but non-operative ordering}
\label{sec:traversal}

A traversal verdict requires semantic order, not merely decreasing norms of finite-pool nearest neighbors. The raw statistic counted strict decreases in retrieved-caption norm, treated plateaus as failures, and scored the shared exact-origin endpoint. Consequently, $50\%$ is not a calibrated chance baseline. In a post-hoc fixed-pool HyCoCLIP-B stress test, a strong planted norm--specificity association (Spearman $0.974$) nevertheless produced only $20.5\%$ strict decrease and $0\%$ perfect paths. This score fell within the $18.7$--$22.8\%$ range of $1{,}000$ random norm assignments (upper-tail add-one $p=0.620$). We therefore retain the raw traversal tables and finite-pool hub examples only as descriptive readout behavior (Appendix~\ref{app:traversal_full}).

We instead use a planted-control-validated semantic traversal over ImageNet--WordNet nodes, with model-native distance, the exact origin excluded, and own-chain versus full-hierarchy candidate pools. After excluding the previously observed HyCoCLIP-S pilot, we preregistered Tier~1 and operative Tier~2 thresholds of $0.12$ and $0.50$, respectively, within fixed multimetric decision rules before running the four previously unobserved confirmatory checkpoints.

In the primary text-source mode, all four confirmatory checkpoints pass own-chain Tier~1 but fail Tier~2. Under full-hierarchy competition only MERU-S passes Tier~1, and none passes Tier~2. Maximum off-chain wrong-hub shares span $86.5$--$99.0\%$ (Table~\ref{tab:traversal_summary}). Thus the models contain partial branch-conditioned order but no operative full-hierarchy readout, rather than no order at all.

\subsection{Threshold changes affect regularization, not operating geometry}
\label{sec:eta}

Finally, we test whether strengthening intra-modal entailment by lowering the threshold $\eta$ alters the measured geometry. Lowering $\eta$ from $1.2$ to $0.7$ changes downstream metrics in a scale-dependent way, but leaves curvature at the floor, the local distortion factor near one, and the cones wider and more saturated (Appendix~\ref{sec:eta_intervention}). The intervention does not recover a nonlocal operating point or a nontrivially active cone. Yet prior work has reported hierarchy-like signals from these same models. The next section asks what those evaluations measure.

%% file: sections/06_Q3.tex
\section{Hierarchy-Looking Evaluations Are Underdetermined}
\label{sec:evaluation}

The previous section shows that the direct diagnostics do not establish an operative hierarchy. We now ask why prior evaluations can nevertheless suggest that hyperbolic VLMs are more hierarchical. We find that several hierarchy-looking metrics are underdetermined. They can capture useful angular semantic organization or bulk norm effects without establishing directed radial hierarchy.

\subsection{Leaf-level taxonomy correlation has no detected unique linear norm increment}
\label{sec:taxonomy}

Prior work reports that hyperbolic VLMs can show stronger correlations between taxonomy distance and embedding distance. We reproduce this qualitative positive signal on CIFAR-100. Hyperbolic checkpoints obtain higher WordNet path-distance correlations than Euclidean CLIP across all sizes. Multiple hyperbolic variants exceed $r \approx 0.45$, with HyCoCLIP-B clampOff attaining the largest three-seed mean ($r = 0.508$) despite its curvature collapsing to $c \approx 0.009$ (Table~\ref{tab:taxonomy_all_models}).

\begin{table}[htbp]
\centering
\caption{
CIFAR-100 leaf-pair taxonomy-distance decomposition. $r$ is the Pearson correlation between cosine distance and WordNet tree distance, and $\Delta R^2_{\mathrm{norm}}$ is the unique linear increment from pairwise absolute norm difference after cosine. Values use manual WordNet disambiguation for $13$ classes; robustness to the naive last-token mapping is reported in Table~\ref{tab:cifar_mapping_robustness}. Current-GRIT HyCoCLIP-B rows average effects over seeds $0/37/42$, with per-seed $p_{\mathrm{perm}}$ ranges. All models are listed in Table~\ref{tab:taxonomy_all_models}. CLIP-B norm entries are N/A because its released output is unit-normalized.
}
\label{tab:taxonomy_decomp}
\begin{tabular}{llccccc}
\toprule
Model & Setting & Taxonomy $r$ & $R^2_{\cos}$ & $R^2_{\mathrm{norm\text{-}only}}$ & $\Delta R^2_{\mathrm{norm}}$ & $p_{\mathrm{perm}}$ \\
\midrule
CLIP-B      & released & 0.370 & 0.137 & N/A & N/A & N/A \\
MERU-B      & released & 0.489 & 0.239 & 0.001 & +0.001 & 0.34 \\
HyCoCLIP-B  & released & 0.465 & 0.217 & 0.005 & +0.001 & 0.50 \\
PHyCLIP-B   & released & 0.456 & 0.208 & 0.002 & +0.000 & 0.74 \\
PHyCLIP-L   & released & 0.456 & 0.208 & 0.003 & +0.001 & 0.50 \\
HyCoCLIP-B  & baseline & 0.462 & 0.214 & 0.001 & +0.002 & 0.12--0.86 \\
HyCoCLIP-B  & clampOff & 0.508 & 0.258 & 0.001 & +0.001 & 0.31--0.56 \\
\bottomrule
\end{tabular}
\end{table}

However, a stronger correlation need not reflect active radial hierarchy. Table~\ref{tab:taxonomy_decomp} tests whether pairwise absolute norm difference adds a unique linear increment beyond cosine, reporting the norm-only $R^2_{\mathrm{norm\text{-}only}}$ and $\Delta R^2_{\mathrm{norm}}=R^2_{\cos+\mathrm{norm}}-R^2_{\cos}$. \rev{For the reported condition-level summaries (averaging the three current-GRIT ViT-B seeds),} cosine distance captures most of the fitted signal on these same-level leaf pairs: the norm-only $R^2$ is small ($\leq0.007$), and $\Delta R^2_{\mathrm{norm}}\leq0.003$ across the decomposed hyperbolic models. Under the manual mapping, Mantel tests detect no significant unique linear contribution from pairwise absolute norm difference in any hyperbolic model or seed.

The mapping choice does not alter the broad pattern. Under the naive mapping, one ViT-B seed cell crosses the threshold, consistent with the per-seed false-positive rate. Pearson $r$ shifts by $0.01$--$0.10$, but the unique linear norm increment remains small under both mappings (Appendix~\ref{app:taxonomy_mapping}; Table~\ref{tab:cifar_mapping_robustness}).

This leaf-pair decomposition tests the unique linear norm increment, not depth coding in general. On released MERU-B, HyCoCLIP-B, and PHyCLIP-B, the registered calibration detected $\Delta R^2_{\mathrm{norm}}=0.020$ in $500/500$ replicates (\rev{one-sided $95\%$} Clopper--Pearson lower bound $0.994$), with a descriptive $80\%$-power transition at $0.0089$--$0.0090$, and observed leaf-pair increments were $0.00018$--$0.00134$. This calibration applies only to effects aligned with the deployed leaf-pair norm feature and these three checkpoints, not to current-GRIT, non-aligned effects, other released rows, or the branch-relative diagnostic (Appendix~\ref{app:sensitivity}; Table~\ref{tab:taxonomy_all_models}).

This reconciles prior positive evidence with our diagnostics. Hyperbolic VLMs can learn useful angular semantic structure, and this can improve symmetric taxonomy-distance correlation. But symmetric semantic distance is not the same as directed radial hierarchy.

\subsection{Directed norm scores require shuffle controls}
\label{sec:shuffle}

Raw directed norm-ordering scores can also mislead. A marginal norm offset, with specific prompts larger-normed than general ones on average, can inflate apparent ordering without learned parent--child structure. Evaluated against the shuffle null instead (Section~\ref{sec:parent_child}), no released hyperbolic checkpoint exceeds the one-sided detection threshold on the CIFAR-100 fine$\to$coarse hierarchy, and no positive crossing is replicated across all three seeds of any current-GRIT ViT-B condition (Figure~\ref{fig:radial}, Table~\ref{tab:parent_child_full}). The native box$\to$caption registry shows a positive conditional one-bit excess ($z\approx+4.7$--$+6.4$) but a numerically small, seed-variable beyond-cosine increment (Table~\ref{tab:native_radial}). Neither establishes an operative graded hierarchy. Directed norm scores therefore require a null control, and because this null preserves level marginals, it does not adjudicate purely level-wise or branch-relative radial codes.

\subsection{Leaf-level benchmarks do not validate hyperbolic hierarchy}
\label{sec:leaf_benchmarks}

Many standard benchmarks are leaf-level discrimination tasks. ImageNet classification evaluates separation among leaf classes, not abstraction across levels, and hierarchy-aware penalties such as TIE or LCA can improve through angular similarity among related leaves rather than directed radial hierarchy. Following the decoupling logic of Section~\ref{sec:evaluation_diagnostics}, such scores cannot by themselves establish active hyperbolic hierarchy, since a model can improve on them while its geometry stays near-Euclidean.

\subsection{Multi-granularity retrieval separates supervision from geometry}
\label{sec:multigranularity}

We evaluate retrieval across WordNet abstraction depths using ancestor-defined query centroids---each formed by averaging the embeddings of an ancestor concept's descendant leaf classes---against ImageNet validation images. This test is more hierarchy-sensitive than leaf-level classification. Coarse queries should retrieve broad semantic groups, while fine queries should retrieve narrow leaf-level classes.

The retrieval picture varies sharply with abstraction depth. At the coarsest levels, all models remain weak in absolute terms. \rev{At depth~$1$ (one qualifying ancestor query)}, even the best AP is $0.128$. At fine depths ($d \geq 11$) the four families are nearly tied at matched scale (ViT-B: AP $0.57$--$0.60$). At coarse depths ($d \leq 5$), however, HyCoCLIP and PHyCLIP substantially outperform CLIP and MERU at every scale, reaching AP $\approx0.33$--$0.34$ at ViT-B compared to $0.18$--$0.19$ for the non-box-supervised models.

This coarse-depth advantage does not by itself identify its mechanism. Across these audited families, the presence of hyperbolic geometry does not predict the gain. MERU remains close to Euclidean CLIP at coarse depths across ViT-S/B/L, whereas HyCoCLIP and PHyCLIP improve substantially (Table~\ref{tab:multigranularity}; Appendix~\ref{app:multigranularity}). These higher-performing models remain near-Euclidean, and their radial, cone, and traversal diagnostics do not establish an operative hierarchy (Section~\ref{sec:hierarchy}). Because this probe does not decompose angle from norm, we do not assign its gain to either channel. We do not isolate box supervision from other pipeline differences. The gain is therefore associated with families using box or compositional supervision rather than causally attributed to box supervision itself.

Across Sections~\ref{sec:taxonomy}--\ref{sec:multigranularity}, each apparent hierarchy signal is compatible with angular semantic organization or marginal effects, or is associated with box/compositional supervision. None uniquely identifies active radial geometry. Together with Sections~\ref{sec:geometry} and~\ref{sec:hierarchy}, this leaves a mechanistic question. Why do interventions intended to install an operative hierarchy fail (Section~\ref{sec:why_interventions_fail})?

%% file: sections/07_Reason.tex
\section{Why Simple Interventions Fail}
\label{sec:why_interventions_fail}

The previous sections show that no converged audited configuration remains in the nonlocal hyperbolic regime or establishes an operative radial/cone hierarchy under our diagnostics. We now ask \emph{why} the natural interventions meant to install hierarchy fail. Two fail \emph{analytically}: pairwise depth ranking has no direct curvature gradient (Section~\ref{sec:ldepth_pairwise}), and naive curvature-coupled depth supervision opens a norm-growth escape route (Section~\ref{sec:ldepth_naive}). The central result (Section~\ref{sec:entailment_shortcut}) is that the hierarchy-motivated entailment objective itself admits a low-curvature shortcut, so curvature collapse is better explained as a property of the loss than as an optimization accident. An entailment-off ablation (Section~\ref{sec:entail0_ablation}) then shows the shortcut is not the only route to low curvature, and activating the entailment term more strongly (Appendix~\ref{sec:eta_intervention}) likewise does not recover a nonlocal operating point or a nontrivially active cone.

\subsection{Pairwise depth ranking has no direct curvature gradient}
\label{sec:ldepth_pairwise}

A natural intervention is to add a pairwise depth-ranking loss encouraging deeper concepts to have larger radii. Let $r=\lVert v\rVert_2$ denote the encoder's pre-exponential-map tangent norm---the quantity such a ranking loss constrains directly, and one that does not depend on the curvature parameter. Defined there,
\begin{equation}
    L_{\mathrm{depth}}
    =
    \max(0, m - (r_{\mathrm{child}}-r_{\mathrm{parent}})),
\end{equation}
so $\partial L_{\mathrm{depth}}/\partial c=0$. In from-scratch training, this loss can change norms but cannot identify curvature. (Defining the same loss on the post-map spatial norm $\rho=\sinh(\sqrt{c}\,r)/\sqrt{c}$ instead couples curvature only through the radial map---precisely the curvature--norm interplay analyzed next.)

\subsection{Naive curvature coupling opens a norm-growth escape route}
\label{sec:ldepth_naive}

A second intervention is to couple depth supervision to the dimensionless radius, for example by replacing the score with $\sqrt{c}\rho$. This creates a direct curvature path, but it also opens an uncontrolled norm-growth path, because $\sqrt{c}\rho$ can be increased either by raising the curvature $c$ or by growing the embedding norm $\rho$, so a model under such supervision can satisfy the depth objective by inflating norms rather than by adjusting curvature, leaving the curvature collapse unaddressed.

This does not mean curvature-coupled objectives are impossible in principle. It shows that curvature-coupled hierarchy supervision must be paired with explicit radial-growth control. Without such control, the depth objective is satisfiable by norm growth alone. This is precisely the path our c-only diagnostic (Section~\ref{sec:entailment_shortcut}) removes, by detaching the norm so that the depth gradient reaches curvature directly.

\subsection{The entailment loss has a low-curvature shortcut}
\label{sec:entailment_shortcut}

The first two interventions fail for fixable reasons---one lacks a curvature gradient, the other lacks norm control. The third reveals a deeper obstacle. The entailment objective \emph{itself} drives curvature down, while a separate c-only counterfactual shows that the depth signal remains weak even after its norm-growth path is blocked.

To separate curvature identification from radial norm growth, we also evaluate a c-only counterfactual, $h=\sqrt{c}\,\mathrm{stopgrad}(\rho)$, which keeps a direct depth-to-curvature gradient while blocking norm changes. Depth supervision is \emph{not} part of the training objective of the probed runs. The counterfactual is defined only for HyCoCLIP, whose depth pairs come from box-level parent captions. Even with the norm-growth path removed, the depth signal is weak and batch-dependent in sign. In a deterministic eight-draw HyCoCLIP-S check, entailment pushes curvature down in every draw while the small depth term pushes it up, and the entailment magnitude exceeds the depth magnitude by $89$--$4500\times$ (Table~\ref{tab:c_only_impl_check}; implementation checks in Appendix~\ref{app:gradient_methods}).

\begin{figure}[t]
  \centering
  \includegraphics[width=\linewidth]{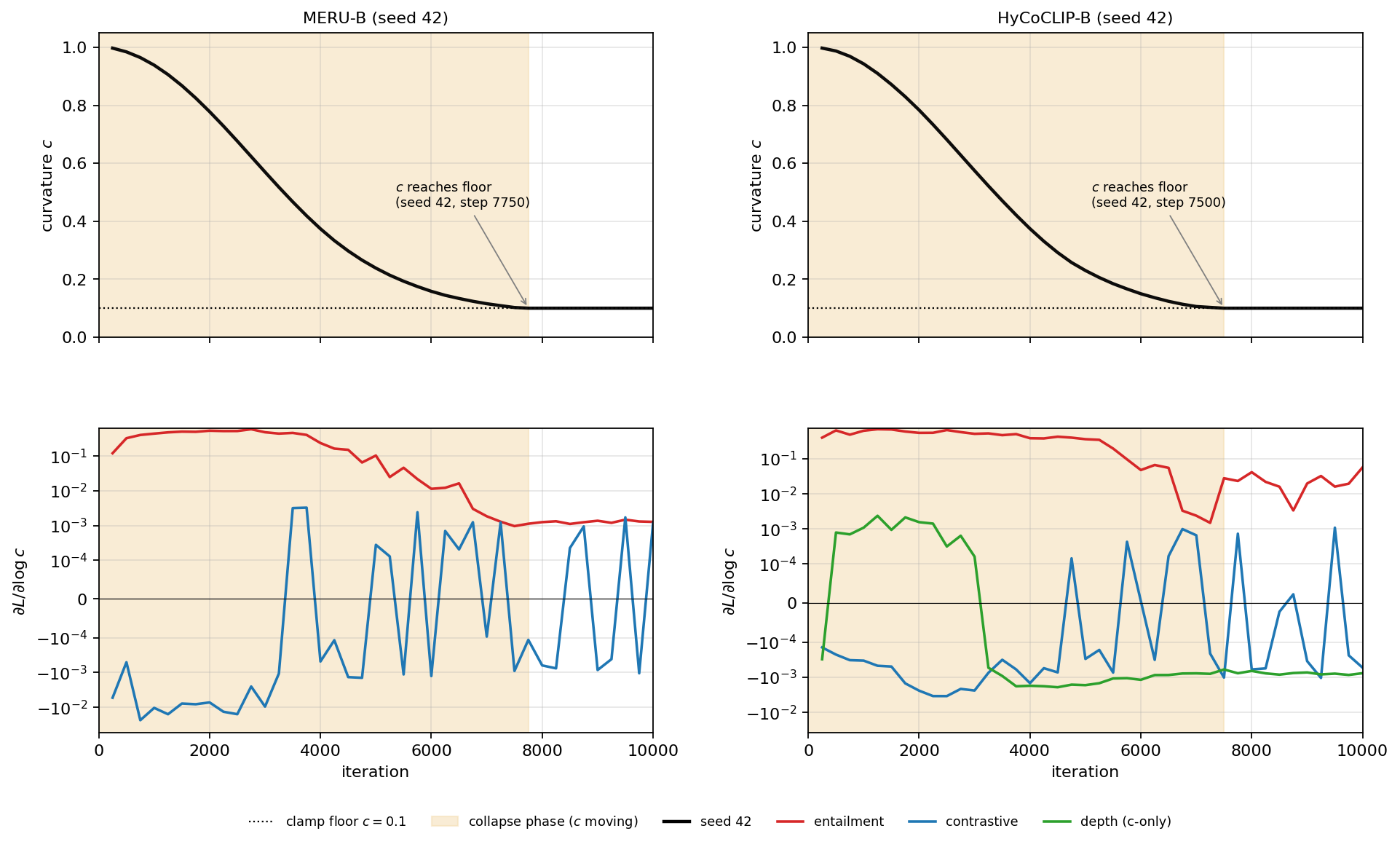}
  \caption{
  Full-objective curvature-gradient attribution for representative ViT-B seed~42
  (first $10$k steps). Top: curvature $c$ falls from $1.0$ to the clamp floor
  ($0.1$); shading marks the collapse phase. Bottom: raw per-loss gradients
  $\partial L/\partial\log c$ on a symlog axis (positive $=$ c-down). In MERU-B
  and HyCoCLIP-B, entailment (red) pushes $c$ down throughout collapse
  ($0.1$--$0.7$) and opposes the smaller contrastive gradient (blue). The c-only
  depth counterfactual (green; HyCoCLIP only) is one to four orders of magnitude
  smaller than entailment. Applying $\lambda_{\mathrm{e}}=0.2$ reduces the
  entailment magnitude but not its sign. Per-seed sign fractions are in Appendix~\ref{app:gradient_methods}.
  }
  \label{fig:gradient}
\end{figure}

We measure the per-loss decomposition directly along the unmodified ViT-B full-objective collapse-probe trajectories of HyCoCLIP-B and MERU-B (Figure~\ref{fig:gradient}), three seeds each ($42$, $37$, and $23$; seed~$23$ replaces the $0$ used elsewhere). It is a mechanism diagnostic rather than a population-level estimate, but the collapse-phase pattern is consistent across all six runs. While $c$ falls from $1.0$ to the floor, the entailment gradient pushes curvature down at every step, exceeds the raw contrastive magnitude at $96.7$--$100\%$ of steps, and dominates after objective weighting at most steps. Per-run medians of the per-step ratio are $48$--$701\times$ raw and $9.6$--$140\times$ after the $\lambda_{\mathrm{e}}=0.2$ weighting. Raw entailment gradients are $0.1$--$0.7$, against a contrastive push of order $10^{-3}$ that points upward while above noise.

Once $c$ reaches the floor, the contrastive gradient decays to noise ($\sim10^{-4}$--$10^{-3}$) with no stable sign. \rev{The entailment gradient remains consistently c-down at every logged post-floor probe. Its magnitude is small in MERU ($O(10^{-3})$) but recovers to $O(10^{-1})$ at later checkpoints in HyCoCLIP.} This residual pressure continues to lower $c$ toward the range observed in the clampOff runs. We do not report a post-floor entailment/contrastive ratio because its contrastive denominator vanishes. The ViT-B depth counterfactual is likewise overwhelmed (roughly $300\times$ smaller than entailment, net c-up but c-down on about one-third of collapse-phase steps; Table~\ref{tab:c_only_impl_check}). MERU-B shows the same collapse-phase pattern despite having no box-level depth term, so the shortcut does not depend on depth supervision. For representative seed~42, $c$ reaches the floor at step $7500$ in HyCoCLIP-B and $7750$ in MERU-B, by which point the entailment pressure has largely subsided, consistent with the shortcut's self-limiting behavior.

This explains why the full objective favors the low-curvature basin. Entailment cones have aperture
\begin{equation}
    \omega(\rho) =
    \arcsin\!\left( \min\!\left\{ 1,\; \frac{2K}{\sqrt{c}\,\rho} \right\} \right).
\end{equation}
As foreshadowed in Section~\ref{sec:cones}, reducing $c$ widens the aperture, which reduces violations in the training (text$\to$image) direction without requiring semantic hierarchy to be learned. The hierarchy-motivated entailment objective therefore contains a low-curvature, wide-cone shortcut, and Figure~\ref{fig:gradient} shows the entailment term driving $c$ downward during collapse, usually more strongly than the contrastive gradient even after objective weighting.

This shortcut is self-limiting, explaining why curvature in the clampOff runs ends above the relaxed $0.001$ clamp rather than running to it. Its c-down pressure comes from violated pairs: it peaks as curvature falls and violations are most active, then weakens as apertures saturate and fewer pairs violate the cone constraint.

In the box-supervised runs, the trained image-parent endpoint is therefore consistent with a balance at the saturation edge. Below it, apertures saturate---clipping the aperture-widening gradient channel and leaving only a weak residual pressure. Above it, violations are active and entailment pushes $c$ down. Whether this balance is a stable attractor---whether the c-down pressure just above the edge is steep enough to dominate the contrastive push---is a dynamical question we leave to future work. The analysis below establishes only \emph{where} the saturation edge lies, and that the box families' trained image-parent coordinate ends near it.

The location of this boundary follows in closed form. With $K$ the fixed constant in the aperture formula ($K=0.1$ throughout, distinct from the numerically equal baseline curvature floor), a cone saturates ($\omega=\pi/2$) exactly when $\sqrt{c}\,\rho \le 2K = 0.2$. The aperture depends on the \emph{parent} embedding of each trained relation, so the coordinate this identity governs is relation-specific. MERU trains a single text$\to$image term (parent: text), while HyCoCLIP and PHyCLIP train four terms (text$\to$image, box-text$\to$box-image, box-image$\to$image, box-text$\to$text) whose parents are the text, box-text, and box-image embeddings.

Under clampOff, every trained relation's parent mean lies at or below the edge. The corresponding parent-cone distributions are fully or nearly fully saturated (saturation $100\%$, except $96.4$--$96.6\%$ for PHyCLIP's box-image factors), clipping the aperture-widening gradient channel. Only one trained parent type approaches the edge: the box-image parent of the box-image$\to$image relation. It lies within $0.013$ of $2K$ in all four box-supervised runs (HyCoCLIP-S/B $0.196$, PHyCLIP-S $0.188$, PHyCLIP-B $0.187$)\footnote{These $\sqrt{c}\rho$ values are measured on the GRIT shards used for the cone diagnostics (Table~\ref{tab:trained_parent_edge}; PHyCLIP values are per-factor). ImageNet/COCO values differ by at most $0.002$ (Table~\ref{tab:clampoff_geometry}), preserving the saturation classification.}, consistent with the cone diagnostics in Section~\ref{sec:cones}.

MERU, which trains no image-parent relation, supports the aperture-clipping picture through its trained text-parent cone, which is below the edge at baseline and farther below under clampOff, but not the image-parent edge-location evidence (Appendix~\ref{app:cone_full}).

Because the half-aperture is a monotone function of the parent $\sqrt{c}\rho$, a parent coordinate at or below the edge necessarily yields a saturated cone. This is an identity, not an independent prediction. The falsifiable observation is where the trained parent coordinates end. In both box-supervised families and at both scales, the trained image-parent coordinate ends within $0.013$ of the $2K$ edge while curvature remains above the relaxed $0.001$ floor. This is consistent with an edge-linked rather than clamp-limited endpoint.

All three implementations share $K=0.1$, however, so we do not observe whether the endpoint moves when the edge is shifted. A $K$-sweep run to arrest would provide a direct causal test of that attribution.

Two further observations sharpen this. Already under the $0.1$ floor at $500$k steps, the HyCoCLIP-B and PHyCLIP-B baselines sit at box-image $\sqrt{c}\rho=0.200$ and $0.205$.

Without entailment, no such arrest appears. All six $\lambda_{\mathrm{e}}{=}0$ runs keep contracting through their probe horizons, and the extended seed-0 runs in both families pass through the edge with no inflection or arrest at $0.2$ (Section~\ref{sec:entail0_ablation}). These controls log the pooled image-side coordinate, which in the entailment-on runs differs from the trained box-image parent coordinate by at most $0.005$. Because the horizons differ ($11$--$40$k versus $500$k), we compare arrest with continued contraction, not endpoint values.

\looseness=1 The evidence for this shortcut is threefold: the aperture identity shows analytically that lowering curvature widens cones and suppresses violations; the trained image-parent coordinate of both box-supervised families, at both scales, ends within $0.013$ of the saturation edge while the entailment-off controls show no arrest there; and the gradient probes show the entailment term pushing curvature down against the opposing contrastive gradient across all three seeds (Appendix~\ref{app:gradient_methods}). \rev{The analytic identity and the endpoint match are consistent with an edge-linked equilibrium. As an endpoint observation, they do not by themselves establish that the arrest location is set by $K$. The decisive test is a $K$-sweep run to arrest that varies the aperture constant and checks whether the endpoint tracks $2K$. Gradient signs corroborate the direction of the entailment pressure.}

This evidence comes from the from-scratch current-GRIT runs. Released checkpoints are pinned at the floor and cannot provide the same intervention. They connect to the mechanism through two facts. The from-scratch baselines and released checkpoints occupy the same near-Euclidean band ($\sqrt{c}\rho\approx0.2$--$0.3$; Section~\ref{sec:released_geometry}). And the closed-form saturation criterion $\sqrt{c}\rho\le2K$, which has no fitted parameters and applies regardless of training history, agrees with the released cone states. Text-parent cones are saturated at $\pi/2$, while the trained box-image parents of the box-supervised families lie at or just above the $0.2$ edge, where the corresponding clampOff coordinate is also observed. The released artifacts are therefore connected to the mechanism through the shared criterion, not through a direct intervention.

The depth term faces a difficulty in either direction. When most active pairs satisfy $\rho_{\mathrm{child}}>\rho_{\mathrm{parent}}$, its gradient is c-up but small (Table~\ref{tab:c_only_impl_check}). When active pairs are wrong-ordered, it can flip c-down and reinforce rather than oppose collapse. Allowing radii to update also reopens the norm-growth path (Section~\ref{sec:ldepth_naive}). A successful objective must therefore jointly learn radial ordering, identify curvature, and control radial growth---three requirements the audited objectives do not satisfy together.

The entailment shortcut is the dominant accelerator on the full-objective trajectory, but not a necessary cause of collapse. As the next section shows, removing entailment still fails to preserve a nonlocal operating point (Section~\ref{sec:entail0_ablation}).

\subsection{Entailment-off training does not stabilize curvature}
\label{sec:entail0_ablation}

The gradient decomposition identifies the entailment term as the dominant curvature-lowering pressure on the full-objective trajectory. We next ask whether it is also necessary for collapse. If so, removing the cone loss should leave curvature high.

We test this by setting $\lambda_{\mathrm{e}}=0$ and retraining otherwise-matched current-GRIT ViT-B collapse probes from scratch for MERU-B and HyCoCLIP-B across three seeds ($0/37/42$). The standard probe horizon is ${\sim}11$k steps, with seed~0 extended to $40$k in both families (per-variant horizons in Appendix~\ref{app:grit_snapshot}). \rev{Table~\ref{tab:horizons} consolidates which analyses rest on full $500$k runs and which on short-horizon probes.} For MERU-B, this yields a \emph{contrastive-only} objective. For HyCoCLIP-B, the \emph{entailment-off} objective retains the box-level compositional and non-entailment alignment terms.

\begin{table}[htbp]
\centering
\caption{
Curvature-collapse summary for ViT-B entailment-off
($\lambda_{\mathrm{e}}=0$) and full-objective
($\lambda_{\mathrm{e}}=0.2$) training under the same protocol and clamp floor
($0.1$). Values are three-seed means ($\{0,37,42\}$ for
$\lambda_{\mathrm{e}}=0$ and $\{23,37,42\}$ for
$\lambda_{\mathrm{e}}=0.2$). The comparison is protocol-matched, not
seed-paired. Curvature reaches the floor in every setting: entailment
accelerates the descent but is not necessary for it. ``Floor step'' is the
first iteration with $c\le0.105$; the stricter $c\le0.1001$ cutoff in
Table~\ref{tab:multiseed_signs} shifts this by at most ${\sim}250$ steps.
$c@k$ denotes curvature at step $k$.
}
\label{tab:entail0}
\begin{tabular}{llcccc}
\toprule
Model & $\lambda_{\mathrm{e}}$ & floor step & $c@3$k & $c@5$k & $c@7$k \\
\midrule
MERU-B     & $0.2$ & $7500$  & $0.570$ & $0.237$ & $0.115$ \\
MERU-B     & $0$   & $10233$ & $0.761$ & $0.423$ & $0.237$ \\
HyCoCLIP-B & $0.2$ & $7250$  & $0.575$ & $0.230$ & $0.107$ \\
HyCoCLIP-B & $0$   & $9833$  & $0.766$ & $0.422$ & $0.230$ \\
\bottomrule
\end{tabular}
\end{table}

\begin{figure}[htbp]
  \centering
  \includegraphics[width=\linewidth]{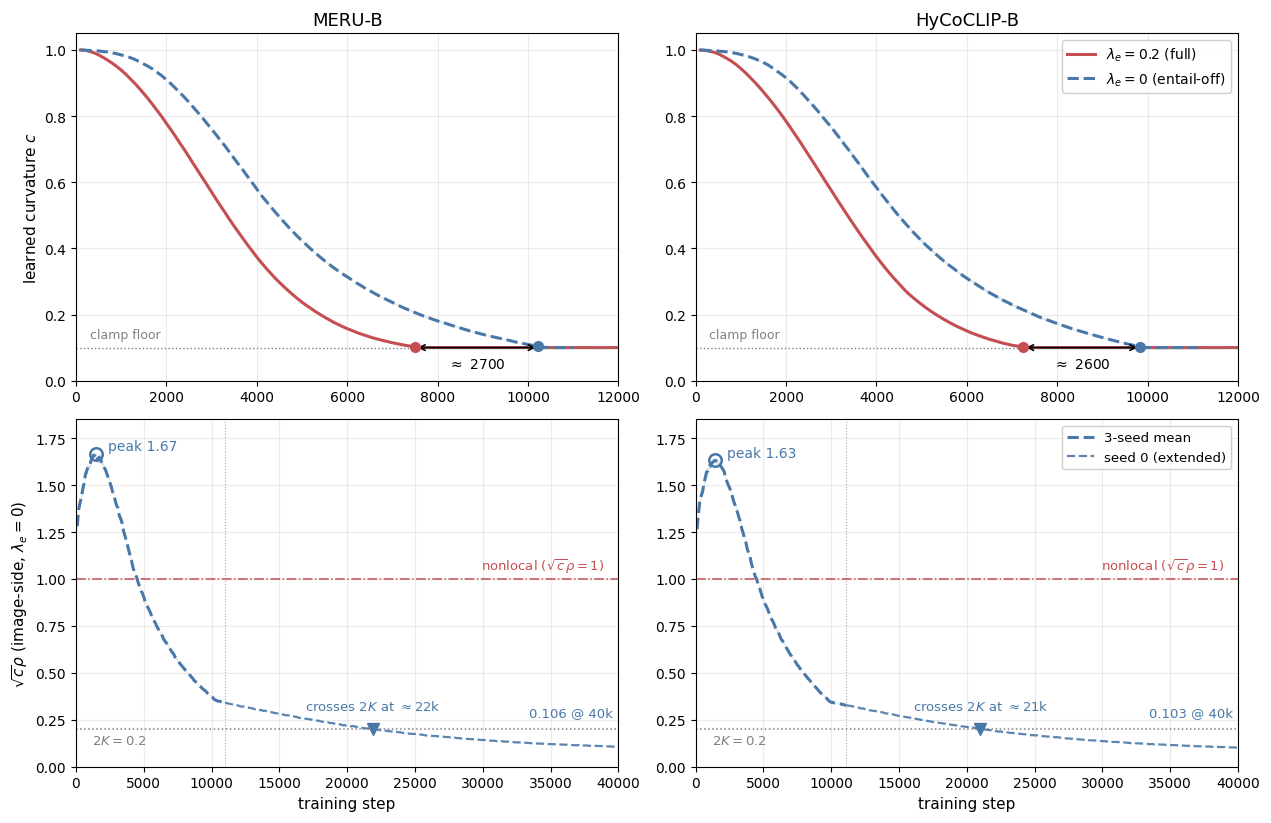}
  \caption{
  \textbf{Removing entailment does not stabilize curvature.}
  \emph{Top:} learned curvature $c$ under the full objective
  ($\lambda_{\mathrm{e}}=0.2$, solid) and entailment-off ablation
  ($\lambda_{\mathrm{e}}=0$, dashed), using ViT-B and the same clamp floor
  ($0.1$). Curvature reaches the floor in both families with or without
  entailment; including entailment accelerates its arrival by
  ${\sim}2600$--$2800$ steps.
  \emph{Bottom:} image-side operating point $\sqrt{c}\rho$ for the
  entailment-off runs. The full-objective collapse probes are omitted because
  they did not log radial norms. The operating point transiently enters the
  nonlocal regime ($\sqrt{c}\rho>1$, peak ${\approx}1.6$--$1.7$) and then
  contracts. Through the ${\sim}11$k probe horizon, curves are three-seed means
  with min--max bands. Beyond the horizon (dotted vertical line), thin dashed
  curves show the seed-0 extensions, which cross the $2K=0.2$ edge at
  ${\approx}21$--$22$k and continue declining to $0.103$--$0.106$ at $40$k
  without arrest (Section~\ref{sec:entail0_ablation}).
  Seed sets are $\{0,37,42\}$ for $\lambda_{\mathrm{e}}=0$ and
  $\{23,37,42\}$ for $\lambda_{\mathrm{e}}=0.2$. The comparison is
  protocol-matched, not seed-paired.
  }
  \label{fig:curv_ablation}
\end{figure}

Removing entailment does not stabilize curvature
(Figure~\ref{fig:curv_ablation}, Table~\ref{tab:entail0}). In every run
curvature still collapses to the clamp floor and the operating point keeps
contracting through the probe horizon. This is consistent with the full-objective gradient decomposition (Figure~\ref{fig:gradient}). The contrastive c-up gradient there carries signal only transiently and is not a persistent restoring force, so once it decays nothing holds curvature up. Entailment is not
necessary for collapse in these settings. What it supplies is a substantial
\emph{acceleration} of it. With the cone loss active
($\lambda_{\mathrm{e}}{=}0.2$) curvature reaches the floor roughly $2600$--$2800$
steps earlier than without it (floor at ${\sim}7.3$--$7.5$k vs.\
${\sim}9.8$--$10.3$k).

The entailment-off runs also expose what the non-entailment objective does on its own. Early in training, $\sqrt{c}\rho$ transiently enters the nonlocal regime in both families and every seed, peaking at ${\approx}1.6$--$1.7$ around step ${\sim}1.3$--$1.8$k (Figure~\ref{fig:curv_ablation}, bottom)\footnote{Operating-point values are the image-side $\sqrt{c}\rho$, computed from the logged mean radial norm (\texttt{rho\_img\_mean}) during training.}. It then contracts as both $c$ and $\rho$ fall. By the ${\sim}11$k probe horizon, all six runs reach $\sqrt{c}\rho=0.33$--$0.35$, with slopes of $-0.035$ to $-0.048$ per $1$k steps and no sign of arrest.

In the seed-0 extensions to $40$k\footnote{\rev{The HyCoCLIP-B extension resumes from its seed-0 $11$k checkpoint, whereas the MERU-B extension is a fresh contiguous run. Both retain the original $500$k cosine learning-rate schedule, and $40$k is the stopping point, not the cosine horizon.}}, the logged image-side coordinate crosses the $2K=0.2$ edge at ${\approx}21$--$22$k and declines smoothly to $0.106$ for MERU-B and $0.103$ for HyCoCLIP-B. Neither trajectory shows a stationary point. Without the cone term, the edge is only a reference landmark and has no dynamical role. Once $c$ reaches the floor, the continued contraction is entirely $\rho$-side.

The non-entailment objective therefore \emph{reaches} the nonlocal operating point but does not \emph{hold} it, whereas the trained image-parent coordinates in the box-supervised entailment-on clampOff runs end near the edge ($0.187$--$0.196$; Section~\ref{sec:entailment_shortcut}). EuCLIP likewise reports entailment-free hyperbolic runs whose scalar curvature ends at the clamp floor \citep{chou2024embedding}. An endpoint alone can conceal the overshoot and subsequent contraction seen in these trajectories.

Two qualifications separate the curvature and norm dynamics. First, curvature collapse is not a direct weight-decay effect. The learned curvature is excluded from weight decay, and the $\lambda_{\mathrm{e}}=0$ probe shows a weak but predominantly c-down contrastive gradient through the collapse phase. The descent is therefore driven rather than a passive parameter drift. The contrastive gradient's sign is trajectory-dependent---c-up on the full-objective trajectory (Figure~\ref{fig:gradient}), but predominantly c-down on the entailment-off trajectory.

Second, the reported runs do not isolate the source of radial-norm contraction. Encoder weight decay ($0.2$) can shrink $\rho$---and hence $\sqrt{c}\rho$---after warmup, while the logit temperature may contribute. We therefore do not attribute the contraction to the objective. This does not affect the main conclusion that the non-entailment objective fails to maintain high curvature or a nonlocal operating point.

The two failures are therefore separable and asymmetric. The contrastive/alignment objective provides only weak, trajectory-dependent curvature pressure and fails to stabilize a nonlocal operating point, while the entailment shortcut adds a much stronger aperture-driven descent, making it the dominant accelerator on the full-objective trajectory but not a necessary cause of collapse. Future models must therefore stabilize a nonlocal operating point under the non-entailment objective, not merely revise the entailment loss (Section~\ref{sec:design_requirements}).

\subsection{\texorpdfstring{\rev{Scope and}}{Scope and} design requirements for future hyperbolic VLMs}
\label{sec:design_requirements}

None of these failures shows that hyperbolic geometry is the wrong tool for vision-language learning. Instead, they clarify why current published formulations leave the radial/cone mechanism dormant, and what a future model must demonstrate to claim otherwise.

\rev{\textbf{Regime limits.} Our from-scratch interventions vary three families, two training scales (ViT-S/B; ViT-L audited at the released level), three seeds, and three objective-side interventions (curvature-floor release, the entailment threshold $\eta$, and entailment removal $\lambda_{\mathrm{e}}=0$), on a fixed 13.1M-pair GRIT snapshot under the published recipes. We hold the optimizer, embedding dimension, and data composition fixed. The low-curvature-shortcut mechanism is therefore established for this recipe family, and its behavior under curvature schedules, other embedding dimensions, or materially different data scales remains unmeasured by our interventions. Two external points provide a data-scale reference: the released checkpoints, trained by the original authors at full scale, and a concurrent hyperbolic VLM trained on the GRIT release for 500k steps, examined next.}

\rev{Concurrent work \citep{kim2026uncha}, inheriting the same cone machinery ($K=0.1$, threshold $\eta$, $\lambda_{\mathrm{e}}=0.2$), reports in its own supplementary that its learned curvature converges to the clamp floor. Applying our operating-point, cone-saturation, and directed-violation diagnostics to its released ViT-S/B checkpoints, we find the same signature: trained text-side parents at $u\approx0.2$ with $\pi/2$-saturated cones, and trained box-image parents above the $2K$ edge with correspondingly \emph{unsaturated} cones, as the parameter-free identity predicts in both directions. Even its highest operating points (band medians $u\approx0.39$--$0.41$, median distortion $2.5$--$2.65\%$) remain far below the $5\%$ marker. Full measurements are in Appendix~\ref{app:uncha_spotcheck}.}

A credible radial/cone hierarchy claim should establish that training stably sustains a nonlocal regime; that trained cones remain active and nontrivially contain directed pairs without a low-curvature or norm-shrinkage escape; and that directed radial ordering survives shuffle and prompt-length controls and supports operational traversal. Evaluation should also separate radial contribution from angular semantic organization. Where training access permits, loss-specific curvature gradients should identify which objective sustains the operating point.

We translate these requirements into a report format that authors and reviewers can adopt directly:

\setlength{\FrameSep}{3pt}\setlength{\FrameRule}{0.4pt}%
\vspace{-4pt}
\begin{framed}\small
\textbf{The five-number geometry report.} A hyperbolic model claiming
radial/cone hierarchy should report, per modality (and per factor for
product spaces):
\begin{enumerate}\itemsep0pt
  \item \textbf{Operating point}: the $\sqrt{c}\rho$ distribution
        (median, p95/p99, maximum, \% above the $10\%$-distortion marker
        $\sqrt{c}\rho>0.84$, and \% above $\sqrt{c}\rho>1$. Our main tables report mean/range summaries and the
$\sqrt{c}\rho>1$ fraction, with medians and upper-tail quantiles in
Appendix~\ref{app:operating_point_quantiles}).
        \hfill\textit{[here: Tables~\ref{tab:released_geometry_summary},
        \ref{tab:clampoff_geometry}]}
  \item \textbf{Saturation state}: cone half-aperture distribution and
        \% saturated at $\pi/2$, with the operating point's position
        relative to the objective's closed-form saturation edge
        ($\sqrt{c}\rho$ vs.\ $2K$ for aperture
        $\arcsin\!\bigl(\min\{1,\,2K/(\sqrt{c}\rho)\}\bigr)$).
        \hfill\textit{[Table~\ref{tab:cone_summary_full}]}
  \item \textbf{Directed violations, both directions}: active violation
        rates at $\eta{=}1$ for parent$\to$child and child$\to$parent
        cones, so trivially satisfied constraints are visible.
        \hfill\textit{[Table~\ref{tab:cone_summary_full}]}
  \item \textbf{Shuffle-controlled radial excess}: directed parent--child
        ordering minus its permuted-pairing null
        ($\Delta_{\mathrm{pair}}$, $z$), not the raw rate.
        \hfill\textit{[Table~\ref{tab:parent_child_full}]}
  \item \textbf{Radial increment beyond angle}:
        $\Delta R^2_{\mathrm{norm}}$ beyond cosine, with Mantel permutation $p$.
        \hfill\textit{[Table~\ref{tab:taxonomy_decomp}]}
\end{enumerate}
\looseness=-1 Where training access permits, additionally attribute the curvature
gradient by loss term ($\partial L/\partial\log c$ sign and magnitude;
Figure~\ref{fig:gradient}). \textbf{Operational hierarchy} complements these
five numbers and is assessed with the planted-control-validated semantic
traversal battery against preregistered thresholds
(Table~\ref{tab:traversal_summary}; Appendix~\ref{app:traversal_full}).
\end{framed}

These diagnostics target active radial/cone hierarchy, not all possible hierarchy representations. Jointly, they address the specific marginal, angular, saturation, and readout alternatives observed in the audited models. Passing all criteria would provide strong, but not conclusive, evidence. A negative result on any one diagnostic does not rule out forms of hierarchy beyond its scope. The report is therefore a practical evidence checklist, not a sufficient condition on its own.

One might object that the audit is self-sealing: if our mechanism (Section~\ref{sec:entailment_shortcut}) explains why no evaluated converged checkpoint sustains a nonlocal operating point, then a negative hierarchy result is guaranteed by construction rather than measured. Two features of the design rule this out.

First, planted controls recover the targeted radial and taxonomy signals, while semantic traversal perfectly recovers a planted hierarchy and detects an injected false hub (Appendices~\ref{app:sensitivity} and~\ref{app:traversal_full}). These controls establish sensitivity only to their stated estimands.

Second, the non-detections have measured, design-specific sensitivity. The directed test reaches $80\%$ power at $\Delta_{\mathrm{pair}}\approx13.5$ pp, compared with a largest released gap of ${\approx}5$ pp. For released MERU-B, HyCoCLIP-B, and PHyCLIP-B, the deployed taxonomy test has a descriptive MDE$_{80}$ of $0.0089$--$0.0090$, and observed increments are $0.00018$--$0.00134$. The latter calibration is not transferred to current-GRIT, and neither sensitivity result is an effect bound (Appendix~\ref{app:sensitivity}).

%% file: sections/08_Conclusion.tex
\section{Conclusion}
\label{sec:conclusion}

We asked whether public hyperbolic vision--language models use the geometry they are built around, and found two distinct failures. First, curvature is not an active geometric resource in any audited converged configuration. Across released checkpoints and converged from-scratch runs in MERU, HyCoCLIP, and PHyCLIP, the dimensionless operating point remains in the near-Euclidean band ($H(u)\approx1$; evaluated maximum $\sqrt{c}\rho\le0.37$). Releasing the curvature floor changes scalar curvature and norms but not this regime. Second, the cone mechanism is non-operative under our diagnostics: trained-parent apertures are saturated or nearly saturated, \rev{eliminating the aperture's discriminative role, so low violation rates alone do not establish learned order}. Semantic traversal detects partial branch-conditioned order but no operative full-hierarchy readout.

At the sensitivity characterized by planted simulations, the shuffle-controlled test detects no pair-specific radial ordering in any released hyperbolic checkpoint. In matched current-GRIT ViT-B runs, no positive result repeats across all three seeds. On a fixed box-to-caption registry, a separate test finds a conditional binary radial signal, but little additional norm information beyond cosine. Neither supports an operative graded hierarchy.

The mechanism analysis helps explain why simple interventions do not repair these failures. Lowering curvature widens entailment cones and reduces violations. All floor-released parent means remain within the analytic saturation region $\sqrt{c}\rho\le2K$, and box-supervised image-parent endpoints lie near its boundary. This proximity does not show that training tracks the boundary. On the probed trajectories, entailment is the dominant curvature-lowering loss pressure. Yet curvature continues to contract when entailment is removed, while weight decay prevents attributing the accompanying norm contraction to the objective alone. Entailment therefore accelerates collapse but is not its sole cause.

These results do not show that hyperbolic geometry is the wrong tool for vision--language learning. They show that the audited published formulations leave the radial/cone mechanism dormant. \rev{Standard positive indicators are instead consistent with angular structure, marginal effects, or box/compositional supervision.} They do not demonstrate active radial depth. To make future claims checkable, we distill our diagnostics into a five-number geometry report (Section~\ref{sec:design_requirements}): operating point, cone-saturation state, directed violations, shuffle-controlled radial excess, and radial increment beyond angle. Paired with the calibrated semantic-traversal battery (Appendix~\ref{app:traversal_full}), these measures let readers judge whether a model's geometry is active, not merely present.

%% file: sections/09_Appendix.tex
\clearpage
\section{Notation}
\label{app:notation}

Table~\ref{tab:notation} collects the symbols used throughout the paper.

\begin{table}[H]
\centering
\caption{Notation used throughout the paper.}
\label{tab:notation}
\begin{tabular}{@{}l p{0.70\textwidth}@{}}
\toprule
Symbol & Meaning \\
\midrule
$c$ & Learned curvature parameter. \\
$\rho = \lVert x_{\mathrm{sp}} \rVert_2$ & Radial coordinate: the Lorentz spatial norm (space components only). The geodesic distance to the origin is $(1/\sqrt{c})\operatorname{asinh}(\sqrt{c}\rho)$. \\
$r = \lVert v \rVert_2$ & Pre-exponential-map tangent norm; independent of $c$, with $\rho=\sinh(\sqrt{c}\,r)/\sqrt{c}$. \\
$u = \sqrt{c}\,\rho$ & Dimensionless operating point. \\
$\sqrt{c}\rho > 1$ & Marker for the nonlocal (operative-hyperbolic) regime. \\
$\hat r = \operatorname{asinh}(u)$ & Dimensionless geodesic radius corresponding to $u$. \\
$H(u) = u/\operatorname{asinh}(u)$ & Local distortion factor (exact tangential stretch of the exponential map at geodesic radius $\hat r$); $H \approx 1$ is near-Euclidean and grows in the nonlocal regime. \\
$K$ & Fixed constant in the cone-aperture formula ($K = 0.1$); a cone saturates when $\sqrt{c}\rho \le 2K$. \\
$\omega(\rho)$ & Entailment-cone half-aperture. \\
$R^2_{\cos}$ & Taxonomy-distance variance explained by cosine (angular) distance. \\
$\Delta R^2_{\mathrm{norm}}$ & Incremental $R^2$ of embedding norm \emph{beyond} cosine distance. \\
$z$ & Shuffle-null standardized score (directed radial test). \\
$\Delta_{\mathrm{pair}}$ & Shuffle-controlled pair-specific excess (radial ordering). \\
$p_{\mathrm{perm}}$ & Mantel / shuffle permutation-test $p$-value. \\
$r_{\mathrm{s}}$ & Per-pair Spearman step--specificity correlation in the NG2 graded readout. \\
OAC & Ordered ancestor coverage in semantic traversal. \\
net & Normalized net progress along the gold ancestor chain. \\
WH & Maximum off-chain wrong-hub share at a scored pre-origin step. \\
fcAUC & False-coalescence area under the traversal curve. \\
$\eta$ & Entailment activation threshold. \\
NG2 & The GRIT box$\to$full-caption relation, native to the box-supervised families' training; evaluated across all models as a controlled relation test. \\
\bottomrule
\end{tabular}
\end{table}

\clearpage  %
\section{Full Geometry Diagnostics}
\label{app:geometry_full}
\subsection{Released-checkpoint per-modality local distortion}

\begin{table}[htbp]
\centering
\caption{
Full released-checkpoint geometry diagnostics. All released checkpoints remain in the near-Euclidean regime, with $0\%$ of samples satisfying $\sqrt{c}\rho>1$. MERU rows use 500 ImageNet validation images and 500 COCO validation captions; HyCoCLIP and PHyCLIP full/box rows use a fixed GRIT-500 sample for within-model modality comparison. PHyCLIP coordinates are aggregated over $64$ factors. HyCoCLIP and PHyCLIP report mean per-sample exact $H(u)$; MERU reports $H$ at the recorded aggregate $u$ because its original persisted output lacks per-sample arrays.
}
\label{tab:released_geometry_full}
\begin{tabular}{lllrrr}
\toprule
Model & Modality & $c$ & $\rho$ & $\sqrt{c}\rho$ & $H(u)$ \\
\midrule
MERU-S       & image      & 0.100 & 0.816 & 0.258 & 1.0109 \\
MERU-S       & text       & 0.100 & 0.547 & 0.173 & 1.0049 \\
MERU-B       & image      & 0.100 & 0.834 & 0.264 & 1.0114 \\
MERU-B       & text       & 0.100 & 0.578 & 0.183 & 1.0055 \\
MERU-L       & image      & 0.100 & 0.882 & 0.279 & 1.0127 \\
MERU-L       & text       & 0.100 & 0.599 & 0.190 & 1.0060 \\
HyCoCLIP-S   & full image & 0.100 & 0.635 & 0.201 & 1.0066 \\
HyCoCLIP-S   & box image  & 0.100 & 0.630 & 0.199 & 1.0065 \\
HyCoCLIP-S   & full text  & 0.100 & 0.389 & 0.123 & 1.0025 \\
HyCoCLIP-S   & box text   & 0.100 & 0.317 & 0.100 & 1.0017 \\
HyCoCLIP-B   & full image & 0.100 & 0.636 & 0.201 & 1.0067 \\
HyCoCLIP-B   & box image  & 0.100 & 0.630 & 0.199 & 1.0065 \\
HyCoCLIP-B   & full text  & 0.100 & 0.385 & 0.122 & 1.0025 \\
HyCoCLIP-B   & box text   & 0.100 & 0.321 & 0.102 & 1.0017 \\
PHyCLIP-B    & full image & 0.100$^\dagger$ & 0.681 & 0.215 & 1.0077 \\
PHyCLIP-B    & box image  & 0.100$^\dagger$ & 0.646 & 0.204 & 1.0069 \\
PHyCLIP-B    & full text  & 0.100$^\dagger$ & 0.442 & 0.140 & 1.0033 \\
PHyCLIP-B    & box text   & 0.100$^\dagger$ & 0.361 & 0.114 & 1.0023 \\
PHyCLIP-L    & full image & 0.100$^\dagger$ & 0.690 & 0.218 & 1.0079 \\
PHyCLIP-L    & box image  & 0.100$^\dagger$ & 0.651 & 0.206 & 1.0071 \\
PHyCLIP-L    & full text  & 0.100$^\dagger$ & 0.442 & 0.140 & 1.0033 \\
PHyCLIP-L    & box text   & 0.100$^\dagger$ & 0.367 & 0.116 & 1.0023 \\
\bottomrule
\end{tabular}

\vspace{2pt}
{\footnotesize $^\dagger$ PHyCLIP uses 64 independently curved subspaces; the value shown is their mean and is not directly comparable to the global curvature of MERU or HyCoCLIP.}
\end{table}

\FloatBarrier  %
\subsection{Operating-point upper tail: \texorpdfstring{$\sqrt{c}\rho$}{sqrt(c)rho} quantiles}
\label{app:operating_point_quantiles}

\begin{figure}[h]
\centering
\includegraphics[width=\linewidth]{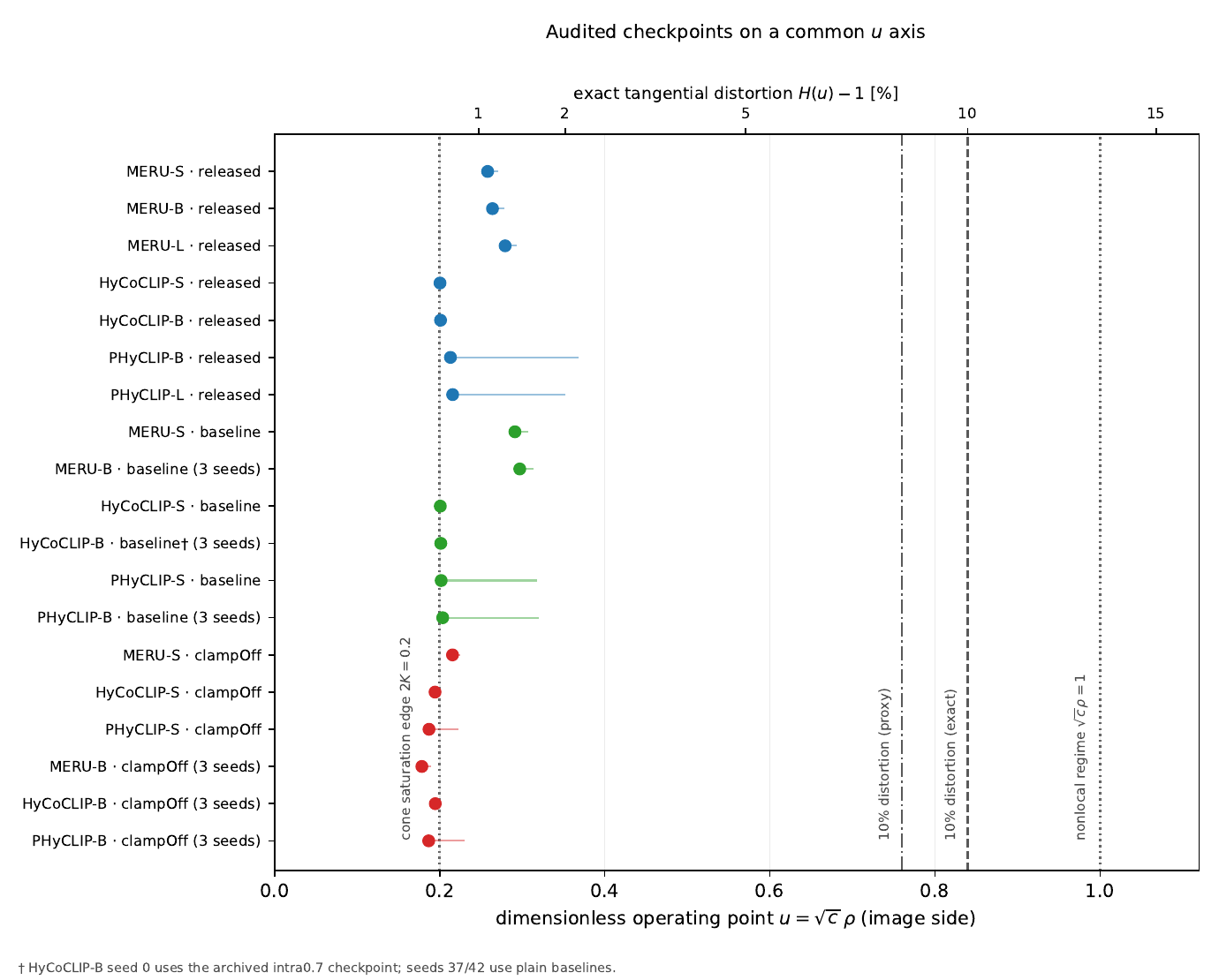}
\caption{\rev{MERU, HyCoCLIP, and PHyCLIP checkpoints on a common $u=\sqrt{c}\rho$ axis. Dots show checkpoint medians, averaged over checkpoints in multi-seed rows; thick segments show their range, and thin lines extend to the per-sample maximum. The top axis gives exact tangential distortion $100[H(u)-1]$. Vertical markers are $2K=0.2$, $0.76$, the exact $10\%$-distortion point $0.840$, and $u=1$. Median ranges in the multi-seed rows are at most $0.004$. The largest observed per-sample distortion is $2.17\%$ ($u=0.3674$); the largest individual checkpoint median incurs $1.45\%$. The marked HyCoCLIP-B row combines a threshold-variant seed-0 checkpoint with plain seed-37/42 baselines, so its range is descriptive rather than controlled seed-only uncertainty.}}
\label{fig:common_u_axis}
\end{figure}

Tables~\ref{tab:released_geometry_summary} and~\ref{tab:clampoff_geometry}
report central tendency and the fraction with $\sqrt{c}\rho>1$.
Table~\ref{tab:operating_point_quantiles} adds the $95$th/$99$th percentiles
and maximum to test whether these summaries hide rare high-radius samples.
Across the $500$ ImageNet images, the largest observed value is $0.367$
(released PHyCLIP-B), below half of both $10\%$-distortion markers
($0.76$ under the proxy and $0.84$ under the exact convention) and well below
the nonlocal marker at $1$ (Figure~\ref{fig:hu_curve}).
For PHyCLIP, the quantiles pool factor-wise operating points, so its wider
tail reflects variation across the $64$ subspaces.

\begin{table}[htbp]
\centering
\small
\caption{
Upper-tail quantiles use a fixed subset of $500$ ImageNet validation images, also used in Table~\ref{tab:clampoff_geometry}. The HyCoCLIP/PHyCLIP full/box means in Table~\ref{tab:released_geometry_full} instead use GRIT-500. Sampling seed~$42$ selects images and is not a model seed. No value exceeds $0.76$, so neither the exact $10\%$-distortion point ($0.84$) nor the nonlocal marker ($1$) is reached; exceedance columns are omitted. \emph{Released} rows are published checkpoints, whereas \emph{baseline}/\emph{clampOff} rows are current-GRIT runs. Baseline and ViT-S rows use model seed~$0$; ViT-B clampOff is reported for seeds $0/37/42$. PHyCLIP quantiles pool its $64$ factor-wise values per image.
}
\label{tab:operating_point_quantiles}
\begin{tabular}{llrrrrr}
\toprule
Model & Setting & Seed & median & p95 & p99 & max \\
\midrule
MERU-S     & released & --- & 0.258 & 0.264 & 0.267 & 0.269 \\
MERU-B     & released & --- & 0.264 & 0.270 & 0.273 & 0.278 \\
MERU-L     & released & --- & 0.279 & 0.287 & 0.290 & 0.292 \\
HyCoCLIP-S & released & --- & 0.200 & 0.203 & 0.204 & 0.205 \\
HyCoCLIP-B & released & --- & 0.201 & 0.204 & 0.205 & 0.206 \\
PHyCLIP-B  & released & --- & 0.213 & 0.257 & 0.279 & 0.367 \\
PHyCLIP-L  & released & --- & 0.216 & 0.262 & 0.287 & 0.351 \\
\midrule
MERU-S     & baseline & 0  & 0.291 & 0.299 & 0.302 & 0.306 \\
MERU-B     & baseline & 0  & 0.296 & 0.305 & 0.308 & 0.310 \\
HyCoCLIP-S & baseline & 0  & 0.201 & 0.203 & 0.204 & 0.207 \\
HyCoCLIP-B & baseline & 0  & 0.202 & 0.205 & 0.206 & 0.206 \\
PHyCLIP-S  & baseline & 0  & 0.202 & 0.239 & 0.257 & 0.317 \\
PHyCLIP-B  & baseline & 0  & 0.204 & 0.242 & 0.263 & 0.314 \\
\midrule
MERU-S     & clampOff & 0  & 0.216 & 0.221 & 0.224 & 0.224 \\
HyCoCLIP-S & clampOff & 0  & 0.194 & 0.197 & 0.197 & 0.198 \\
PHyCLIP-S  & clampOff & 0  & 0.187 & 0.199 & 0.205 & 0.222 \\
MERU-B     & clampOff & 0  & 0.180 & 0.185 & 0.187 & 0.189 \\
MERU-B     & clampOff & 37 & 0.177 & 0.181 & 0.183 & 0.184 \\
MERU-B     & clampOff & 42 & 0.179 & 0.184 & 0.186 & 0.187 \\
HyCoCLIP-B & clampOff & 0  & 0.195 & 0.197 & 0.198 & 0.199 \\
HyCoCLIP-B & clampOff & 37 & 0.194 & 0.197 & 0.197 & 0.198 \\
HyCoCLIP-B & clampOff & 42 & 0.195 & 0.197 & 0.197 & 0.198 \\
PHyCLIP-B  & clampOff & 0  & 0.187 & 0.200 & 0.206 & 0.225 \\
PHyCLIP-B  & clampOff & 37 & 0.187 & 0.199 & 0.206 & 0.225 \\
PHyCLIP-B  & clampOff & 42 & 0.187 & 0.200 & 0.206 & 0.229 \\
\bottomrule
\end{tabular}
\end{table}

\FloatBarrier  %
\subsection{Full per-seed downstream and evaluation results}
\label{app:downstream_full_results}

This section reports the per-seed absolute metrics underlying Table~\ref{tab:downstream_delta} (seeds $0/37/42$, baseline and clampOff), from which each delta can be recomputed. Tables~\ref{tab:retrieval_hier_full}, \ref{tab:zsc_full}, and \ref{tab:compositional_full} report retrieval and WordNet hierarchical metrics, the $16$-task zero-shot suite, and per-subtype compositional results, respectively. Across metrics and models, curvature unclamping produces mixed, generally modest shifts.

\begin{table}[htbp]
\centering
\caption{
Per-seed absolute zero-shot retrieval (Recall@$5$/@$10$, points) and WordNet
hierarchical-classification metrics for current-GRIT baseline and clampOff
(seeds $0/37/42$), underlying the deltas in
Table~\ref{tab:downstream_delta}. TIE/LCA are distances (lower is better);
J/PH/RH range from $0$ to $1$ (higher is better).
}
\label{tab:retrieval_hier_full}
\setlength{\tabcolsep}{3pt}
\resizebox{\textwidth}{!}{%
\begin{tabular}{ll cccc cccc ccccc}
\toprule
& & \multicolumn{4}{c}{Text$\to$Image} & \multicolumn{4}{c}{Image$\to$Text} & \multicolumn{5}{c}{Hierarchical Classification} \\
\cmidrule(lr){3-6}\cmidrule(lr){7-10}\cmidrule(lr){11-15}
& & \multicolumn{2}{c}{COCO} & \multicolumn{2}{c}{Flickr} & \multicolumn{2}{c}{COCO} & \multicolumn{2}{c}{Flickr} & \multicolumn{5}{c}{WordNet} \\
\cmidrule(lr){3-4}\cmidrule(lr){5-6}\cmidrule(lr){7-8}\cmidrule(lr){9-10}\cmidrule(lr){11-15}
Model / setting & seed & R@5 & R@10 & R@5 & R@10 & R@5 & R@10 & R@5 & R@10 & TIE$\downarrow$ & LCA$\downarrow$ & J$\uparrow$ & PH$\uparrow$ & RH$\uparrow$ \\
\midrule
MERU-B baseline & 0 & 55.28 & 66.40 & 81.38 & 88.28 & 68.52 & 78.82 & 89.10 & 94.60 & 3.872 & 2.330 & 0.7707 & 0.8426 & 0.8440 \\
 & 37 & 55.75 & 67.27 & 82.08 & 88.66 & 69.84 & 79.76 & 90.10 & 94.90 & 3.903 & 2.331 & 0.7685 & 0.8413 & 0.8417 \\
 & 42 & 55.09 & 66.49 & 81.66 & 88.72 & 68.58 & 78.34 & 90.40 & 94.80 & 3.864 & 2.307 & 0.7710 & 0.8438 & 0.8424 \\
\addlinespace[2pt]
MERU-B clampOff & 0 & 55.50 & 66.99 & 81.48 & 88.22 & 68.28 & 79.12 & 88.80 & 94.30 & 3.886 & 2.313 & 0.7687 & 0.8428 & 0.8410 \\
 & 37 & 55.15 & 66.77 & 81.84 & 88.60 & 69.14 & 79.00 & 91.10 & 95.40 & 3.994 & 2.352 & 0.7618 & 0.8378 & 0.8355 \\
 & 42 & 55.34 & 66.56 & 81.64 & 88.80 & 68.12 & 78.40 & 89.70 & 94.00 & 3.907 & 2.325 & 0.7682 & 0.8422 & 0.8409 \\
\addlinespace[2pt]
\midrule
HyCoCLIP-B baseline & 0 & 57.38 & 68.39 & 82.94 & 89.70 & 69.36 & 79.50 & 91.40 & 96.30 & 3.299 & 2.086 & 0.8054 & 0.8684 & 0.8670 \\
 & 37 & 56.31 & 67.28 & 82.60 & 89.54 & 69.34 & 79.74 & 90.50 & 95.10 & 3.395 & 2.125 & 0.8002 & 0.8642 & 0.8640 \\
 & 42 & 57.33 & 68.40 & 83.50 & 89.96 & 70.44 & 80.38 & 91.80 & 95.80 & 3.282 & 2.077 & 0.8070 & 0.8692 & 0.8687 \\
\addlinespace[2pt]
HyCoCLIP-B clampOff & 0 & 57.64 & 69.07 & 84.12 & 90.28 & 71.40 & 80.68 & 92.30 & 95.40 & 3.353 & 2.083 & 0.8017 & 0.8673 & 0.8627 \\
 & 37 & 57.44 & 68.69 & 83.16 & 89.84 & 70.92 & 80.98 & 92.10 & 96.70 & 3.498 & 2.173 & 0.7947 & 0.8599 & 0.8590 \\
 & 42 & 58.82 & 69.53 & 83.92 & 90.74 & 71.62 & 81.00 & 92.30 & 95.30 & 3.378 & 2.118 & 0.8002 & 0.8647 & 0.8620 \\
\addlinespace[2pt]
\midrule
PHyCLIP-B baseline & 0 & 56.84 & 68.18 & 82.92 & 89.78 & 69.94 & 79.96 & 91.70 & 95.40 & 3.346 & 2.111 & 0.8034 & 0.8665 & 0.8661 \\
 & 37 & 57.01 & 67.95 & 82.60 & 89.28 & 70.56 & 79.56 & 90.80 & 95.10 & 3.342 & 2.124 & 0.8042 & 0.8659 & 0.8677 \\
 & 42 & 56.79 & 68.13 & 82.96 & 89.46 & 69.64 & 79.76 & 90.60 & 94.80 & 3.284 & 2.070 & 0.8067 & 0.8697 & 0.8671 \\
\addlinespace[2pt]
PHyCLIP-B clampOff & 0 & 57.34 & 68.41 & 83.64 & 89.98 & 70.98 & 80.50 & 91.20 & 95.60 & 3.375 & 2.106 & 0.8009 & 0.8655 & 0.8629 \\
 & 37 & 57.54 & 68.76 & 82.64 & 89.34 & 70.66 & 80.16 & 91.00 & 95.40 & 3.377 & 2.110 & 0.8002 & 0.8650 & 0.8627 \\
 & 42 & 57.68 & 68.53 & 83.50 & 90.08 & 70.40 & 80.34 & 91.40 & 95.70 & 3.367 & 2.097 & 0.8010 & 0.8660 & 0.8632 \\
\bottomrule
\end{tabular}%
}
\end{table}

\begin{table}[htbp]
\centering
\caption{Per-seed absolute zero-shot classification accuracy (top-1, \%) across 16 datasets. Current-GRIT baseline and clampOff, seeds $0,37,42$.}
\label{tab:zsc_full}
\setlength{\tabcolsep}{3pt}
\resizebox{\textwidth}{!}{%
\begin{tabular}{ll c c c c c c c c c c c c c c c c}
\toprule
Model / setting & seed & IN & C10 & C100 & SUN & Cal & STL & Food & CUB & Cars & Airc & Pets & Flwr & DTD & Euro & RES & C211 \\
\midrule
MERU-B baseline & 0 & 37.32 & 75.31 & 45.11 & 49.45 & 72.24 & 92.80 & 48.89 & 8.44 & 7.22 & 2.42 & 42.40 & 17.78 & 20.59 & 37.14 & 40.36 & 5.05 \\
 & 37 & 37.06 & 73.96 & 42.79 & 50.30 & 72.37 & 93.50 & 49.64 & 10.73 & 6.84 & 2.15 & 41.74 & 21.62 & 22.93 & 40.90 & 41.36 & 4.36 \\
 & 42 & 37.39 & 75.28 & 44.00 & 50.05 & 73.25 & 92.55 & 47.90 & 9.44 & 7.07 & 2.33 & 40.43 & 15.09 & 19.84 & 37.82 & 41.41 & 4.91 \\
\addlinespace[2pt]
MERU-B clampOff & 0 & 37.12 & 70.66 & 44.21 & 49.41 & 72.49 & 93.39 & 46.44 & 11.12 & 6.96 & 2.36 & 41.47 & 17.28 & 22.23 & 34.67 & 37.52 & 4.75 \\
 & 37 & 36.40 & 74.14 & 44.03 & 49.58 & 72.08 & 93.19 & 47.03 & 10.54 & 6.21 & 2.36 & 43.42 & 20.85 & 19.31 & 40.89 & 43.67 & 4.54 \\
 & 42 & 36.87 & 76.02 & 43.21 & 49.86 & 71.76 & 92.66 & 50.58 & 10.43 & 6.43 & 2.09 & 44.79 & 16.58 & 21.33 & 47.27 & 42.21 & 4.72 \\
\addlinespace[2pt]
\midrule
HyCoCLIP-B baseline & 0 & 44.39 & 89.01 & 59.42 & 55.08 & 74.36 & 94.94 & 55.21 & 15.60 & 10.20 & 3.98 & 52.38 & 22.17 & 26.76 & 41.14 & 43.32 & 5.57 \\
 & 37 & 42.53 & 87.94 & 58.63 & 54.44 & 77.43 & 94.12 & 56.57 & 17.59 & 9.11 & 3.86 & 51.01 & 24.56 & 25.80 & 41.80 & 46.98 & 5.45 \\
 & 42 & 44.06 & 89.04 & 59.27 & 55.04 & 76.93 & 95.06 & 59.65 & 18.37 & 9.80 & 3.28 & 53.28 & 26.06 & 24.84 & 39.89 & 49.62 & 5.81 \\
\addlinespace[2pt]
HyCoCLIP-B clampOff & 0 & 43.71 & 87.84 & 57.11 & 54.47 & 75.80 & 94.33 & 57.46 & 16.29 & 10.64 & 3.09 & 51.83 & 26.39 & 26.38 & 42.13 & 42.92 & 5.98 \\
 & 37 & 42.38 & 88.26 & 57.93 & 54.09 & 75.22 & 93.82 & 55.14 & 15.63 & 10.08 & 3.56 & 49.77 & 24.97 & 22.98 & 32.28 & 44.28 & 5.92 \\
 & 42 & 43.83 & 88.66 & 59.20 & 54.32 & 76.56 & 94.16 & 55.67 & 14.27 & 11.01 & 3.99 & 52.02 & 26.80 & 27.07 & 40.70 & 45.60 & 5.41 \\
\addlinespace[2pt]
\midrule
PHyCLIP-B baseline & 0 & 43.38 & 88.68 & 59.48 & 55.41 & 77.69 & 94.10 & 58.72 & 16.19 & 9.67 & 3.63 & 53.00 & 25.36 & 25.90 & 41.59 & 46.92 & 5.36 \\
 & 37 & 43.63 & 88.16 & 58.45 & 55.28 & 75.07 & 95.10 & 58.51 & 17.79 & 10.29 & 3.16 & 52.68 & 23.81 & 27.93 & 47.18 & 45.27 & 5.60 \\
 & 42 & 44.25 & 88.70 & 60.74 & 55.92 & 77.31 & 95.19 & 59.20 & 15.86 & 10.55 & 3.17 & 52.99 & 25.29 & 27.61 & 40.17 & 48.66 & 5.63 \\
\addlinespace[2pt]
PHyCLIP-B clampOff & 0 & 43.56 & 88.17 & 58.90 & 55.14 & 76.45 & 93.99 & 55.10 & 14.64 & 9.21 & 4.23 & 54.13 & 24.26 & 25.64 & 38.01 & 46.24 & 5.37 \\
 & 37 & 43.55 & 87.08 & 58.18 & 54.50 & 75.67 & 94.94 & 55.90 & 14.69 & 9.26 & 3.45 & 50.82 & 26.25 & 26.17 & 39.39 & 45.19 & 5.68 \\
 & 42 & 43.27 & 88.16 & 58.34 & 53.77 & 77.67 & 95.00 & 57.40 & 15.47 & 10.59 & 2.75 & 50.40 & 23.35 & 27.18 & 41.87 & 46.45 & 5.73 \\
\bottomrule
\end{tabular}%
}
\end{table}

\begin{table}[htbp]
\centering
\caption{Per-seed absolute compositional (hard-negative) accuracy (points). VL-CheckList-Object: Location (Center/Mid/Margin), Size (Large/Medium/Small). SugarCrepe: Replace (Obj/Att/Rel), Swap (Obj/Att), Add (Obj/Att), overall. Current-GRIT baseline and clampOff, seeds $0,37,42$.}
\label{tab:compositional_full}
\setlength{\tabcolsep}{3pt}
\resizebox{\textwidth}{!}{%
\begin{tabular}{ll c c c c c c c c c c c c c c}
\toprule
& & \multicolumn{6}{c}{VL-CheckList-Object} & \multicolumn{8}{c}{SugarCrepe} \\
\cmidrule(lr){3-8}\cmidrule(lr){9-16}
Model / setting & seed & Loc-C & Loc-M & Loc-Mg & Sz-L & Sz-M & Sz-S & Rep-O & Rep-A & Rep-R & Swp-O & Swp-A & Add-O & Add-A & SC-All \\
\midrule
MERU-B baseline & 0 & 62.90 & 60.30 & 61.60 & 64.00 & 60.60 & 58.80 & 90.68 & 79.57 & 69.84 & 57.14 & 63.96 & 80.02 & 71.68 & 77.47 \\
 & 37 & 62.10 & 58.10 & 57.10 & 62.90 & 57.60 & 55.50 & 88.92 & 81.09 & 69.77 & 60.82 & 66.52 & 80.80 & 72.69 & 77.89 \\
 & 42 & 54.40 & 53.20 & 54.00 & 56.10 & 54.10 & 50.70 & 89.95 & 79.19 & 70.06 & 55.10 & 64.41 & 78.90 & 75.29 & 77.31 \\
\addlinespace[2pt]
MERU-B clampOff & 0 & 60.80 & 60.50 & 61.80 & 63.40 & 61.10 & 55.20 & 89.53 & 79.95 & 69.20 & 57.96 & 67.42 & 80.21 & 72.83 & 77.63 \\
 & 37 & 64.60 & 62.80 & 61.00 & 65.30 & 62.50 & 59.30 & 89.65 & 79.57 & 69.49 & 60.41 & 67.72 & 81.96 & 71.10 & 78.10 \\
 & 42 & 57.50 & 56.30 & 55.70 & 59.50 & 56.60 & 56.00 & 90.13 & 79.44 & 69.35 & 53.06 & 62.01 & 80.75 & 73.27 & 77.29 \\
\addlinespace[2pt]
\midrule
HyCoCLIP-B baseline & 0 & 67.50 & 68.10 & 69.30 & 69.60 & 65.00 & 66.40 & 90.80 & 77.66 & 65.79 & 58.37 & 64.41 & 82.06 & 73.41 & 77.34 \\
 & 37 & 75.00 & 74.30 & 70.90 & 75.70 & 71.00 & 71.20 & 91.16 & 80.71 & 65.58 & 58.78 & 64.56 & 81.57 & 70.81 & 77.35 \\
 & 42 & 76.30 & 73.80 & 71.30 & 77.10 & 71.30 & 71.50 & 90.92 & 78.81 & 65.22 & 57.14 & 63.66 & 81.28 & 74.42 & 77.15 \\
\addlinespace[2pt]
HyCoCLIP-B clampOff & 0 & 69.00 & 68.70 & 70.60 & 70.80 & 66.80 & 69.50 & 90.62 & 81.60 & 69.56 & 60.82 & 65.17 & 82.25 & 74.13 & 78.68 \\
 & 37 & 67.80 & 67.60 & 67.90 & 67.90 & 64.90 & 69.90 & 91.40 & 80.84 & 67.78 & 59.18 & 69.07 & 83.37 & 72.98 & 78.94 \\
 & 42 & 69.80 & 68.70 & 68.00 & 70.90 & 65.20 & 68.10 & 90.86 & 81.60 & 67.85 & 53.06 & 65.17 & 83.27 & 72.69 & 78.31 \\
\addlinespace[2pt]
\midrule
PHyCLIP-B baseline & 0 & 76.60 & 73.00 & 73.10 & 76.70 & 70.70 & 72.00 & 91.10 & 80.33 & 65.50 & 60.41 & 64.86 & 81.09 & 74.42 & 77.57 \\
 & 37 & 73.30 & 70.80 & 68.80 & 73.80 & 67.70 & 66.90 & 90.98 & 80.33 & 66.64 & 60.82 & 66.52 & 82.10 & 70.09 & 77.79 \\
 & 42 & 77.60 & 74.80 & 70.80 & 79.50 & 70.00 & 70.80 & 91.16 & 79.70 & 68.14 & 60.41 & 66.82 & 81.04 & 72.69 & 78.01 \\
\addlinespace[2pt]
PHyCLIP-B clampOff & 0 & 78.20 & 75.80 & 75.10 & 77.80 & 75.00 & 74.30 & 90.74 & 82.11 & 68.92 & 60.41 & 65.77 & 83.75 & 74.86 & 79.16 \\
 & 37 & 76.20 & 72.60 & 71.70 & 75.90 & 70.70 & 71.20 & 91.16 & 80.20 & 67.71 & 59.18 & 62.91 & 83.46 & 75.00 & 78.47 \\
 & 42 & 77.20 & 75.40 & 75.40 & 78.10 & 73.20 & 74.60 & 91.22 & 80.20 & 66.71 & 56.73 & 62.91 & 82.69 & 75.14 & 78.02 \\
\bottomrule
\end{tabular}%
}
\end{table}

\FloatBarrier
\subsection{Cone aperture and entailment violation diagnostics}
\label{app:cone_full}

Table~\ref{tab:trained_parent_edge} lists the trained entailment relations,
their parent embeddings, and the parent operating coordinates under clampOff.
A trained cone's aperture depends only on its parent.

\paragraph{MERU and the edge-location evidence.} MERU trains no image-parent relation. Its sole trained parent (text) is already below the edge at baseline ($0.188$) and moves farther below it under clampOff ($0.130$ for MERU-B and $0.158$ for MERU-S), rather than arresting at the edge. The image-side coordinates ($0.178$--$0.182$ for MERU-B and $0.217$ for MERU-S) apply only to the hypothetical full-image-parent diagnostic; MERU does not optimize this cone. Consistent with the aperture identity, the MERU-B diagnostic cone is saturated, whereas the MERU-S cone remains non-saturated at $1.175$ rad because its image-side coordinate exceeds $2K$. MERU therefore supports the aperture-channel-clipping picture through its trained text-parent cone, but not the trained image-parent edge-location evidence.

\rev{\textbf{Figure~\ref{fig:cone_width_violation} measurement details.} The 16 box-text points use aperture medians and violation rates measured together in a GRIT-256 evaluation with box-selection seed~0. The eight HyCoCLIP medians are $1.5708$\,rad, with $100\%$ saturation; the eight PHyCLIP medians are $1.5694$\,rad, with approximately $99.9$--$100\%$ saturation. PHyCLIP aperture summaries pool its 64 component spaces; numerical clipping places its maximum slightly below $\pi/2$.}

\rev{Baseline and clampOff use matched retrainings with the original and relaxed curvature bounds, respectively. Coincident markers are drawn without jitter. The matched-versus-re-paired containment comparisons are reported in Table~\ref{tab:delta_cone} (Appendix~\ref{app:delta_cone}).}

\begin{table}[htbp]
\centering
\caption{
Trained entailment relations, parent embeddings, and parent operating
coordinates $\sqrt{c}\rho$ under clampOff, measured on GRIT
($N{=}256$) using the protocol of Table~\ref{tab:cone_summary_full}.
Distances to the saturation edge $2K{=}0.2$ are given in parentheses.
All relation-level parent means lie at or below the edge. Parent-aperture
saturation is $100\%$, except for PHyCLIP's per-factor box-image cones
($96.4$--$96.6\%$). Only the box-image parent of the
box-image$\to$image relation approaches the edge. MERU trains only
text$\to$image; its image-side coordinates elsewhere in this appendix
describe an untrained diagnostic cone.
}
\label{tab:trained_parent_edge}
\begin{tabular}{lcccc}
\toprule
Run & t$\to$i (text) & bt$\to$bi (box text) & bi$\to$i (box image) & bt$\to$t (box text) \\
\midrule
MERU-S clampOff     & $0.158\;(-0.042)$ & --- & --- & --- \\
MERU-B clampOff     & $0.130\;(-0.070)$ & --- & --- & --- \\
HyCoCLIP-S clampOff & $0.143\;(-0.057)$ & $0.134\;(-0.066)$ & $0.196\;(-0.004)$ & $0.134\;(-0.066)$ \\
HyCoCLIP-B clampOff & $0.144\;(-0.056)$ & $0.127\;(-0.073)$ & $0.196\;(-0.004)$ & $0.127\;(-0.073)$ \\
PHyCLIP-S clampOff  & $0.139\;(-0.061)$ & $0.124\;(-0.076)$ & $0.188\;(-0.012)$ & $0.124\;(-0.076)$ \\
PHyCLIP-B clampOff  & $0.136\;(-0.064)$ & $0.122\;(-0.078)$ & $0.187\;(-0.013)$ & $0.122\;(-0.078)$ \\
\bottomrule
\end{tabular}
\end{table}

\FloatBarrier  %
Table~\ref{tab:cone_summary_full} reports the full cone diagnostics across families, sizes, released checkpoints, and clampOff, including all three current-GRIT ViT-B seeds. The patterns match Section~\ref{sec:cones}. The reverse i$\to$t rate stays near $100\%$, as expected from the measured inner-text/outer-image ordering. This is a directionality check, not a trained relation. Where the trained t$\to$i rate is small, it coincides with a saturated text-parent cone and alone does not establish learned order. The trained bt$\to$bi rates are likewise near-trivial and are reported only as descriptive checks. MERU and HyCoCLIP begin from opposite diagnostic image-cone regimes at baseline ($\approx0.74$ versus $\approx1.45$ rad), but both saturate to $\pi/2$ under clampOff at ViT-B. MERU-S clampOff is the exception ($0.75\to1.175$ rad, still non-saturated). Combined with the traversal results, these cone diagnostics provide no evidence of operational hierarchy.

\begin{table}[htbp]
\centering
\caption{
Full cone aperture and entailment-violation diagnostics across families,
sizes, released checkpoints, and clampOff. Image-aperture entries give the
median in radians and the fraction saturated at $\pi/2$ in parentheses.
Saturation is defined by $\omega\geq\pi/2-0.01$. Violation columns report the
untrained diagnostic i$\to$t direction (parent $=$ full image), the trained
t$\to$i direction (parent $=$ text), and trained bt$\to$bi for the
box-supervised families. Violations use the exact cone boundary
($\eta{=}1.0$), rather than the relation-specific training margins
($\eta{=}0.7$ or $1.2$). PHyCLIP values are factor-wise means over $64$
subspaces. Measurements use a $256$-sample batch from the same GRIT
distribution as Table~\ref{tab:clampoff_geometry_full}. Current-GRIT ViT-B rows report seeds
$0/37/42$; ViT-S rows use seed~$0$.
}
\label{tab:cone_summary_full}
\resizebox{\textwidth}{!}{%
\begin{tabular}{llcccccc}
\toprule
Model & Setting & $c$ & img aper (rad, sat\%) & text aper (rad, sat\%) & i$\to$t viol \% & t$\to$i viol \% & bt$\to$bi viol \% \\
\midrule
MERU-S$^{\ddagger}$ & released & 0.100 & 0.875 (0\%)    & 1.571 (98.8\%)  & 100.0 & 48.4 & -- \\
MERU-B$^{\ddagger}$ & released & 0.100 & 0.852 (0\%)    & 1.571 (98.0\%)  & 100.0 & 64.8 & -- \\
MERU-L$^{\ddagger}$ & released & 0.100 & 0.802 (0\%)    & 1.571 (97.7\%)  & 100.0 & 70.7 & -- \\
HyCoCLIP-S & released & 0.100 & 1.475 (30.5\%) & 1.571 (100\%)   & 100.0 & 0.39 & \rev{0.39} \\
HyCoCLIP-B & released & 0.100 & 1.456 (26.2\%) & 1.571 (100\%)   & 100.0 & 0.39 & \rev{0.39} \\
PHyCLIP-B  & released & 0.100 & 1.219 (24.6\%) & 1.569 (99.4\%)  & 100.0 & 8.5  & \rev{6.0} \\
PHyCLIP-L  & released & 0.100 & 1.185 (21.4\%) & 1.569 (99.5\%)  & 100.0 & 6.3  & \rev{5.6} \\
\midrule
MERU-S     & baseline & 0.100 & 0.751 (0\%)   & 1.571 (99.2\%)  & 100.0 & 5.1  & -- \\
MERU-S     & clampOff & 0.042 & 1.175 (0\%)   & 1.571 (100\%)  & 100.0 & 5.5  & -- \\
MERU-B     & baseline (s0)  & 0.100 & 0.732 (0\%)   & 1.571 (99.6\%) & 100.0 & 3.1 & -- \\
MERU-B     & baseline (s37) & 0.100 & 0.726 (0\%)   & 1.571 (99.2\%) & 100.0 & 2.7 & -- \\
MERU-B     & baseline (s42) & 0.100 & 0.733 (0\%)   & 1.571 (99.2\%) & 100.0 & 3.1 & -- \\
MERU-B     & clampOff (s0)  & 0.029 & 1.571 (100\%) & 1.571 (100\%)  & 100.0 & 1.6  & -- \\
MERU-B     & clampOff (s37) & 0.028 & 1.571 (100\%)   & 1.571 (100\%)  & 100.0 & 1.2  & -- \\
MERU-B     & clampOff (s42) & 0.029 & 1.571 (100\%)   & 1.571 (100\%)  & 100.0 & 3.1  & -- \\
HyCoCLIP-S & baseline & 0.100 & 1.446 (15.6\%)& 1.571 (100\%)  & 100.0 & 0.0 & 0.8 \\
HyCoCLIP-S & clampOff & 0.011 & 1.571 (100\%) & 1.571 (100\%)  & 100.0 & 0.0 & 0.0 \\
HyCoCLIP-B & baseline (s0)  & 0.100 & 1.422 (10.9\%) & 1.571 (100\%) & 100.0 & 0.0 & 0.0 \\
HyCoCLIP-B & baseline (s37) & 0.100 & 1.420 (12.9\%) & 1.571 (100\%) & 100.0 & 0.0 & 0.4 \\
HyCoCLIP-B & baseline (s42) & 0.100 & 1.423 (10.5\%) & 1.571 (100\%) & 100.0 & 0.4 & 0.4 \\
HyCoCLIP-B & clampOff (s0)  & 0.009 & 1.571 (100\%) & 1.571 (100\%)  & 100.0 & 0.0 & 0.0 \\
HyCoCLIP-B & clampOff (s37) & 0.009 & 1.571 (100\%) & 1.571 (100\%)  & 100.0 & 0.0 & 0.0 \\
HyCoCLIP-B & clampOff (s42) & 0.009 & 1.571 (100\%) & 1.571 (100\%)  & 100.0 & 0.0 & 0.0 \\
PHyCLIP-S  & baseline & 0.100 & 1.417 (44.7\%)& 1.571 (99.5\%) & 100.0 & 13.8 & 6.5 \\
PHyCLIP-S  & clampOff & 0.015 & 1.569 (95.5\%)& 1.569 (100\%)  & 100.0 & 0.20 & 0.05 \\
PHyCLIP-B  & baseline (s0)  & 0.100 & 1.373 (41.7\%) & 1.569 (99.7\%) & 100.0 & 9.5 & 6.1 \\
PHyCLIP-B  & baseline (s37) & 0.100 & 1.372 (40.9\%) & 1.569 (99.7\%) & 100.0 & 9.3 & 6.4 \\
PHyCLIP-B  & baseline (s42) & 0.100 & 1.366 (41.4\%) & 1.569 (99.7\%) & 100.0 & 9.1 & 5.6 \\
PHyCLIP-B  & clampOff (s0)  & 0.014 & 1.569 (95.6\%) & 1.569 (100\%)  & 100.0 & 0.05 & 0.04 \\
PHyCLIP-B  & clampOff (s37) & 0.014 & 1.569 (95.5\%) & 1.569 (100\%)  & 100.0 & 0.10 & 0.04 \\
PHyCLIP-B  & clampOff (s42) & 0.015 & 1.569 (95.7\%) & 1.569 (100\%)  & 100.0 & 0.09 & 0.05 \\
\bottomrule
\end{tabular}}
\par\vspace{3pt}
{\footnotesize\raggedright $^{\ddagger}$Released MERU is RedCaps-trained rather than GRIT-trained, so its GRIT-evaluated t$\to$i rate ($48$--$71\%$) partly reflects domain shift, not learned order. Its i$\to$t rate is still $\approx100\%$.\par}
\end{table}

\clearpage  %
\subsection{Current-GRIT clampOff geometry by modality}

\begin{table}[htbp]
\centering
\caption{
Full modality-wise geometry for current-GRIT MERU and HyCoCLIP baseline and
clampOff runs, with the setting included in the model name. Statistics use $500$ GRIT samples
(processed shards 00000--00001), matching the evaluation distribution of the
cone diagnostics in Table~\ref{tab:cone_summary}. This table complements the
ImageNet/COCO evaluation in Table~\ref{tab:clampoff_geometry}. MERU has no box
modality. PHyCLIP is reported separately because its $64$ independently curved
subspaces require factor-wise aggregation. Its current-GRIT image-side
summaries are in Tables~\ref{tab:clampoff_geometry}
and~\ref{tab:clampoff_geometry_perseed}, and its factor-aware relation and cone
diagnostics are in Tables~\ref{tab:trained_parent_edge}
and~\ref{tab:cone_summary_full}. Every reported model and modality
remains near-Euclidean, with a nonlocal-regime fraction of $0\%$. Rows use
representative seed~$0$; per-seed ViT-B image-side values are in
Table~\ref{tab:clampoff_geometry_perseed}. MERU's image-side coordinates do not
define its trained cone parent, which is the text side
(Table~\ref{tab:trained_parent_edge}).
}
\label{tab:clampoff_geometry_full}
\begin{tabular}{llrrrr}
\toprule
Model & Modality & $c$ & $\rho$ & $\sqrt{c}\rho$ & $H(u)$ \\
\midrule
MERU-S baseline      & image      & 0.100 & 0.929 & 0.294 & 1.0141 \\
MERU-S baseline      & text       & 0.100 & 0.591 & 0.187 & 1.0058 \\
MERU-S clampOff      & image      & 0.042 & 1.061 & 0.217 & 1.0077 \\
MERU-S clampOff      & text       & 0.042 & 0.773 & 0.158 & 1.0041 \\
MERU-B baseline      & image      & 0.100 & 0.947 & 0.299 & 1.0145 \\
MERU-B baseline      & text       & 0.100 & 0.595 & 0.188 & 1.0058 \\
MERU-B clampOff & image      & 0.029 & 1.060 & 0.182 & 1.0055 \\
MERU-B clampOff & text       & 0.029 & 0.759 & 0.130 & 1.0028 \\
\midrule
HyCoCLIP-S baseline  & full image & 0.100 & 0.637 & 0.201 & 1.0067 \\
HyCoCLIP-S baseline  & box image  & 0.100 & 0.631 & 0.200 & 1.0066 \\
HyCoCLIP-S baseline  & full text  & 0.100 & 0.392 & 0.124 & 1.0026 \\
HyCoCLIP-S baseline  & box text   & 0.100 & 0.313 & 0.099 & 1.0016 \\
HyCoCLIP-S clampOff  & full image & 0.011 & 1.848 & 0.195 & 1.0063 \\
HyCoCLIP-S clampOff  & box image  & 0.011 & 1.853 & 0.196 & 1.0063 \\
HyCoCLIP-S clampOff  & full text  & 0.011 & 1.350 & 0.143 & 1.0034 \\
HyCoCLIP-S clampOff  & box text   & 0.011 & 1.266 & 0.134 & 1.0030 \\
\midrule
HyCoCLIP-B baseline  & full image & 0.100 & 0.639 & 0.202 & 1.0067 \\
HyCoCLIP-B baseline  & box image  & 0.100 & 0.632 & 0.200 & 1.0066 \\
HyCoCLIP-B baseline  & full text  & 0.100 & 0.383 & 0.121 & 1.0024 \\
HyCoCLIP-B baseline  & box text   & 0.100 & 0.320 & 0.101 & 1.0017 \\
HyCoCLIP-B clampOff  & full image & 0.009 & 2.058 & 0.195 & 1.0063 \\
HyCoCLIP-B clampOff  & box image  & 0.009 & 2.065 & 0.196 & 1.0063 \\
HyCoCLIP-B clampOff  & full text  & 0.009 & 1.515 & 0.144 & 1.0034 \\
HyCoCLIP-B clampOff  & box text   & 0.009 & 1.337 & 0.127 & 1.0027 \\
\bottomrule
\end{tabular}
\end{table}

\begin{table}[t]
\centering
\caption{Per-seed absolute image-side geometry ($500$ ImageNet and $500$ COCO val images), underlying the summary in Table~\ref{tab:clampoff_geometry}. ViT-B rows are reported for all three seeds ($0$, $37$, $42$). ViT-S rows are single-seed (seed $0$). MERU's image-side values are diagnostic coordinates---its trained cone parent is the text side (Table~\ref{tab:trained_parent_edge}). The nonlocal-regime fraction is $0\%$ in every row. MERU-B clampOff is seed-consistent: $\sqrt{c}\rho=0.176$--$0.180$ across seeds, below $2K{=}0.2$ and saturated. The GRIT per-modality / aperture breakdown is Table~\ref{tab:clampoff_geometry_full}.}
\label{tab:clampoff_geometry_perseed}
\setlength{\tabcolsep}{5pt}
\begin{tabular}{ll cccc}
\toprule
Model / setting & seed & $c$ & $\rho$ (img) & $\sqrt{c}\rho$ & $H(u)$ \\
\midrule
MERU-S baseline & 0 & 0.1000 & 0.9206 & 0.2911 & 1.0138 \\
\addlinespace[2pt]
MERU-S clampOff & 0 & 0.0419 & 1.0523 & 0.2154 & 1.0076 \\
\midrule
MERU-B baseline & 0 & 0.1000 & 0.9362 & 0.2961 & 1.0143 \\
 & 37 & 0.1000 & 0.9430 & 0.2982 & 1.0145 \\
 & 42 & 0.1000 & 0.9371 & 0.2964 & 1.0143 \\
\addlinespace[2pt]
MERU-B clampOff & 0 & 0.0294 & 1.0506 & 0.1800 & 1.0054 \\
 & 37 & 0.0281 & 1.0515 & 0.1762 & 1.0051 \\
 & 42 & 0.0291 & 1.0451 & 0.1784 & 1.0053 \\
\midrule
HyCoCLIP-S baseline & 0 & 0.1000 & 0.6348 & 0.2007 & 1.0066 \\
\addlinespace[2pt]
HyCoCLIP-S clampOff & 0 & 0.0112 & 1.8390 & 0.1944 & 1.0062 \\
\midrule
HyCoCLIP-B baseline & 0 & 0.1000 & 0.6379 & 0.2017 & 1.0067 \\
 & 37 & 0.1000 & 0.6377 & 0.2017 & 1.0067 \\
 & 42 & 0.1000 & 0.6374 & 0.2016 & 1.0067 \\
\addlinespace[2pt]
HyCoCLIP-B clampOff & 0 & 0.0090 & 2.0531 & 0.1949 & 1.0063 \\
 & 37 & 0.0094 & 2.0022 & 0.1945 & 1.0062 \\
 & 42 & 0.0095 & 2.0017 & 0.1947 & 1.0063 \\
\midrule
PHyCLIP-S baseline & 0 & 0.1000 & 0.6440 & 0.2036 & 1.0068 \\
\addlinespace[2pt]
PHyCLIP-S clampOff & 0 & 0.0155 & 1.5084 & 0.1877 & 1.0058 \\
\midrule
PHyCLIP-B baseline & 0 & 0.1000 & 0.6501 & 0.2056 & 1.0070 \\
 & 37 & 0.1000 & 0.6500 & 0.2055 & 1.0070 \\
 & 42 & 0.1000 & 0.6507 & 0.2058 & 1.0070 \\
\addlinespace[2pt]
PHyCLIP-B clampOff & 0 & 0.0143 & 1.5700 & 0.1875 & 1.0058 \\
 & 37 & 0.0144 & 1.5612 & 0.1873 & 1.0058 \\
 & 42 & 0.0145 & 1.5539 & 0.1872 & 1.0058 \\
\bottomrule
\end{tabular}
\end{table}

\clearpage  %
\section{GRIT Snapshot and Training Details}
\label{app:grit_snapshot}

\begin{table}[htbp]
\centering\small
\caption{\rev{Which analyses rest on which horizons. Only the entailment-off no-arrest control uses short-horizon probes. Its design compares arrest against continued contraction at matched horizons, not endpoints.}}
\label{tab:horizons}
\rev{
\begin{tabular}{p{4.3cm}p{4.2cm}p{4.6cm}}
\toprule
Analysis & Runs & Horizon \\
\midrule
Gradient decomposition (Fig.~\ref{fig:gradient}, sign tables) & $6$ full-objective collapse probes ($2$ families $\times3$ seeds) & $500$k (full); the analysis window (collapse phase ${\sim}7.5$--$7.8$k and post-floor steps) lies inside the full runs \\
Entailment-off no-arrest control & $\lambda_{\mathrm{e}}=0$, $2$ families $\times3$ seeds & $10$--$11$k probes; seed-$0$ extended to $40$k in both families \\
$\eta$ intervention & $2$ scales & $500$k \\
Geometry, cone, radial, traversal, downstream tables & released + baseline + clampOff & released artifacts and $500$k runs \\
\bottomrule
\end{tabular}
}
\end{table}

\subsection{Current-GRIT snapshot}

\begin{table}[htbp]
\centering
\caption{
Statistics of the fixed current-GRIT snapshot used for controlled interventions. GRIT is URL-derived. The effective corpus is smaller than the original release due to dead links. All 2{,}051 shards were enumerated. The per-shard breakdown is provided in the released artifact (\texttt{data\_provenance/raw\_json/grit\_snapshot\_count\_per\_shard.csv}).
}
\label{tab:grit_snapshot}
\begin{tabular}{lr}
\toprule
Quantity & Value \\
\midrule
Download / crawl date            & 2026-04-13 (file mtime) \\
Total shards (\texttt{.tar})     & 2{,}051 (all enumerated) \\
Processed corpus size            & 731 GB \\
Image--text pairs                & documented 20.5M / obtained $13{,}064{,}747$ \\
Parent-box annotations           & documented 35.9M / obtained $25{,}018{,}042$ \\
Mean parent boxes per pair       & 1.915 ($25{,}018{,}042 / 13{,}064{,}747$) \\
Pairs per shard                  & min 5{,}611, mean 6{,}370, max 6{,}540 \\
Nominal passes over snapshot      & ${\approx}29$ ($384$M exposures $/$ 13.06M pairs) \\
Mapper                           & \texttt{GroundedDatasetTarMapper} \\
Train image transform            & RandomResizedCrop(224, scale=(0.5,1.0)) + ToTensor \\
Decoding failures                & \texttt{webdataset warn\_and\_continue} \\
\bottomrule
\end{tabular}
\end{table}

A ``parent-box annotation'' is counted as one \texttt{parent\{NNN\}.txt} grounding entry per sample (the \texttt{box\_text} source consumed by \texttt{GroundedDatasetTarMapper}), matching the per-sample box unit used elsewhere in this paper. The documented $35.9$M figure is reproduced from the original release and may aggregate box annotations under a slightly different convention.

\FloatBarrier
\subsection{Training hyperparameters}

\begin{table}[htbp]
\centering
\caption{
Training configurations for current-GRIT interventions. All comparisons use
the same snapshot and differ only in the variable listed in the final column.
For PHyCLIP, clampOff applies the same floor relaxation
($[0.1,10]\to[0.001,10]$) independently to the product-space curvatures.
}
\label{tab:training_hparams}
\begin{tabular}{lcccc}
\toprule
Family & Size & Variant & Curv. clamp & Changed variable \\
\midrule
MERU & S/B & baseline & [0.1, 10] & none \\
MERU & S/B & clampOff & [0.001, 10] & curvature floor \\
MERU & B & $\lambda_{\mathrm{e}}{=}0$ & [0.1, 10] & entailment weight ($\to 0$) \\
HyCoCLIP & S/B & baseline & [0.1, 10] & none \\
HyCoCLIP & S/B & clampOff & [0.001, 10] & curvature floor \\
HyCoCLIP & S/B & intra=$0.7$ & [0.1, 10] & entailment threshold \\
HyCoCLIP & B & $\lambda_{\mathrm{e}}{=}0$ & [0.1, 10] & entailment weight ($\to 0$) \\
PHyCLIP & S/B & baseline & [0.1, 10] & none \\
PHyCLIP & S/B & clampOff & [0.001, 10] & curvature floor \\
\bottomrule
\end{tabular}
\end{table}

\noindent Unless noted below, all variants use a global batch size of $768$,
$500{,}000$ iterations, AdamW ($\mathrm{lr}=5\times10^{-4}$,
$\beta=(0.9,0.98)$, weight decay $0.2$)\footnote{The serialized training configurations list \texttt{gain\_bias\_decay}${}=0$ for normalization gains and biases, but the inherited optimizer code (\texttt{phyclip/optim.py}, as in the public MERU and HyCoCLIP implementations) replaces an explicit zero with the regular weight decay, so these parameters are also decayed at $0.2$. The learned curvature and the other explicitly excluded scalars have zero decay. We verified the stored optimizer groups ($0.2$, $0.2$, $0$) in the six full-objective $500$k checkpoints and the two $40$k extensions; the same code path governs every run reported here.}, $4{,}000$ steps of linear warmup
followed by cosine decay, and no gradient accumulation. Each iteration is one
optimizer step over $768$ samples. Training uses AMP with fp16 forward passes
and explicit fp32 casts for hyperbolic operations in
\texttt{phyclip/lorentz.py}.

The $\eta{=}0.7$ interventions use the full $500{,}000$-iteration budget
(seed~$0$, ViT-S and ViT-B). \rev{The only shorter runs are the
$\lambda_{\mathrm{e}}{=}0$ probes: $11.0$--$11.2$k steps for seeds $0/37/42$ in
both families, with seed~$0$ extensions to $40$k
(Section~\ref{sec:entail0_ablation}). The dedicated
$\lambda_{\mathrm{e}}{=}0.2$ collapse probes (seeds $23/37/42$) are full
$500{,}000$-iteration runs; their per-loss gradient probe is logged densely
(every $250$ steps) through the collapse window and at coarser intervals
($1$k--$5$k steps) to $500$k (Table~\ref{tab:horizons}).}

Canonical ViT-S runs use one H100, and canonical ViT-B runs use
$2\times$H200. MERU-B clampOff seeds~$37/42$ were retrained on the unified
$2\times$H200 setup after preliminary single-GPU runs showed hardware
sensitivity; the canonical reruns converge with seed~$0$. ViT-L is evaluated
only through released checkpoints (Table~\ref{tab:released_geometry_full}).

\subsection{Threshold-activation stress test ($\eta$ intervention)}
\label{sec:eta_intervention}

We complement the entailment-off ablation by making the published entailment
constraint stricter rather than removing it. Specifically, we lower the
intra-modal threshold $\eta$ from $1.2$ to $0.7$, which activates more
violations, in matched current-GRIT HyCoCLIP runs at ViT-S and ViT-B
(seed~$0$). Only $\eta$ changes.

\begin{table}[htbp]
\centering
\caption{
Threshold-activation stress test on current-GRIT HyCoCLIP (seed~$0$).
Lowering $\eta$ from $1.2$ to $0.7$ leaves the operating geometry unchanged
while widening and further saturating the cones. Behavioral effects are
scale-dependent and adverse at ViT-B. This single-seed comparison is
decoupling evidence, not a performance claim. The ZSC mean uses the zero-shot
suite of this $\eta$ study and is not directly comparable in absolute value
with the $16$-task suite in Table~\ref{tab:zsc_full}. The violation rate uses
the exact-boundary diagnostic at $\eta=1.0$, not the training margin. The two
traversal rows are raw fixed-pool descriptors and carry no semantic-traversal
verdict.
}
\label{tab:eta_intervention}
\begin{tabular}{lcccccc}
\toprule
& \multicolumn{3}{c}{ViT-B} & \multicolumn{3}{c}{ViT-S} \\
\cmidrule(lr){2-4}\cmidrule(lr){5-7}
Quantity & $\eta{=}1.2$ & $\eta{=}0.7$ & $\Delta$ & $\eta{=}1.2$ & $\eta{=}0.7$ & $\Delta$ \\
\midrule
\multicolumn{7}{l}{\emph{Operating geometry (unchanged)}} \\
\quad $c$                       & 0.100 & 0.100 & $0$      & 0.100 & 0.100 & $0$ \\
\quad $\sqrt{c}\rho$ (image)    & 0.202 & 0.201 & $-0.001$ & 0.201 & 0.201 & $0$ \\
\quad $H(u)$                    & 1.007 & 1.007 & $0$      & 1.007 & 1.007 & $0$ \\
\quad \% $\sqrt{c}\rho{>}1$     & 0\%   & 0\%   & $0$      & 0\%   & 0\%   & $0$ \\
\midrule
\multicolumn{7}{l}{\emph{Entailment-loss engagement and cone geometry}} \\
\quad image aper.\ (rad)        & 1.422 & 1.469 & $+0.05$  & 1.446 & 1.491 & $+0.05$ \\
\quad image sat.\ \%            & 10.9  & 25.8  & $+14.9$  & 15.6  & 32.8  & $+17.2$ \\
\quad t$\to$i diagnostic viol.\ \% & 0.0 & 1.2 & $+1.2$ & 0.0 & 0.8 & $+0.8$ \\
\midrule
\multicolumn{7}{l}{\emph{Behavior (scale-dependent)}} \\
\quad raw norm mono.\ (INet)    & 16.5  & 8.2   & $-8.3$   & 15.2  & 17.4  & $+2.2$ \\
\quad raw collapse by step 15 (INet) & 1.5 & 100 & $+98.5$ & 10.0 & 7.0 & $-3.0$ \\
\quad COCO t2i R@5              & 57.4  & 54.0  & $-3.4$   & 51.5  & 52.9  & $+1.4$ \\
\quad ZSC mean                  & 41.4  & 39.4  & $-2.0$   & 38.0  & 37.2  & $-0.8$ \\
\bottomrule
\end{tabular}
\end{table}

The operating geometry is unchanged at both scales: curvature remains at the
floor, $\sqrt{c}\rho\approx0.20$, $H(u)=1.007$, and no embedding reaches the
nonlocal regime. The stricter threshold instead widens the cones and raises
image-aperture saturation from $10.9\%$ to $25.8\%$ at ViT-B and from
$15.6\%$ to $32.8\%$ at ViT-S, while the diagnostic t$\to$i violation rises
only to ${\sim}1\%$. Stronger engagement therefore produces wider, more
saturated cones without changing the operating regime.

Behavioral effects are mixed across scales. At ViT-B, raw norm monotonicity
falls from $16.5\%$ to $8.2\%$, raw collapse by step~15 rises from
$1.5\%$ to $100\%$, and COCO retrieval and ZSC decrease by $3.4$ and $2.0$
points. ViT-S changes are smaller and mixed, including a $2.2$-point raw
monotonicity increase and a $1.4$-point COCO increase. This scale dependence
matches Section~\ref{sec:eta}.

These results are consistent with the low-curvature shortcut of
Section~\ref{sec:entailment_shortcut}: tightening the entailment constraint
increases cone saturation but does not recover a nonlocal operating point or
a nontrivially active cone. The raw fixed-pool traces are descriptive and do not establish semantic ordering.

These current-GRIT interventions are structural rather than performance
comparisons. ViT-B results use three seeds and are reported as seed means
without formal significance tests; ViT-S and $\eta$ interventions use seed~$0$.
They test whether a geometric constraint changes the operating geometry,
hierarchy diagnostics, or downstream behavior under matched training.

\section{Full Traversal Results}
\label{app:traversal_full}

A planted-radius stress test showed that strict retrieved-norm decrease rates
are not calibrated semantic measures: the score was $20.5\%$ with no perfect
paths and lay near the centre of $1{,}000$ randomized assignments
(upper-tail add-one $p=0.620$). We therefore report these fixed-pool rates
descriptively and use the semantic diagnostic below for the primary traversal
judgment.

\subsection{Semantic traversal protocol}
\label{app:semantic_traversal_protocol}

The semantic diagnostic evaluates source-conditioned ImageNet/WordNet order. For each of $200$ source classes, it follows the gold leaf-to-root chain while moving at constant intrinsic speed, $q(t)=\operatorname{Exp}_0((1-t)\operatorname{Log}_0(x))$, over 20 scored points with $0\le t<1$.  The exact-origin query at $t=1$ is neither generated nor scored. Retrieval uses one Lorentz distance for MERU/HyCoCLIP and the configured factor-wise product distance for PHyCLIP.  The \emph{own-chain} pool contains only the source's gold ancestors; the \emph{full-hierarchy} pool contains all unambiguous ImageNet leaves and ancestors.  Ordered ancestor coverage (OAC), Kendall $\tau_b$, normalized net progress, wrong-hub share, and false-coalescence AUC treat plateaus as ties and permit convergence to a genuine common ancestor.  Source bootstrap intervals and semantic permutations keep steps within a source trajectory together.

\subsection{Semantic positive control and preregistered decisions}
\label{app:semantic_traversal_registration}

Eight adversarial unit tests passed.  A dedicated planted Lorentz traversal toy
reached OAC $=1.000$, and an injected false hub was detected.  This toy checks
the semantic traversal implementation; it is distinct from the balanced
radial tree in Table~\ref{tab:positive_control_tree}, which checks the broader
parent--child, taxonomy, sector, and radial diagnostics and is not a traversal
calibration certificate.

Before the confirmatory checkpoints were run, the decision rule was frozen at
commit \texttt{cb1af9a}.  Tier~1 required OAC and net progress at least $0.12$
(approximately one rung of the mean WordNet chain), permutation $p\le.01$, and
bootstrap lower bounds above the permutation reference.  Tier~2 required OAC,
$\tau_b$, and net progress above $0.50$ with lower bounds above $0.50$; the
full-hierarchy condition also required wrong-hub share at most $0.50$ and
false-coalescence AUC at most $0.10$. HyCoCLIP-S was designated as a pilot because its result preceded registration
and was excluded from the four-checkpoint confirmatory decision.

\subsection{Confirmatory semantic traversal results}
\label{app:semantic_traversal_results}

Table~\ref{tab:traversal_summary} reports the preregistered primary text-source
results.  All four checkpoints passed own-chain Tier~1 but failed Tier~2; only
MERU-S passed full-hierarchy Tier~1, and no checkpoint passed full-hierarchy
Tier~2.  Thus all four map to ``trace order plus global readout failure,'' not
to absence of all order.  All OAC, net-progress, and $\tau_b$ statistics
reported here had permutation $p=0.001$;
failures arose from plateaus, skipped rungs, and off-chain absorption rather
than systematic reversal.

\begin{table}[htbp]
\centering
\caption{
Confirmatory text-source semantic traversal on $200$ ImageNet/WordNet sources.
OAC, net, WH, and fcAUC denote ordered ancestor coverage, normalized net
progress, maximum off-chain wrong-hub share, and false-coalescence AUC.
``Map 2'' denotes own-chain Tier~1 detection with full-hierarchy Tier~2 failure.
Selected full-hierarchy estimates with $95\%$ source-bootstrap CIs are MERU-S
OAC $0.178\,[0.168,0.188]$ and net $0.266\,[0.223,0.308]$, and PHyCLIP-B net
$0.009\,[0.005,0.014]$.
}
\label{tab:traversal_summary}
\resizebox{0.89\textwidth}{!}{%
\begin{tabular}{lccc ccccc l}
\toprule
& \multicolumn{3}{c}{Own chain} & \multicolumn{5}{c}{Full hierarchy} & \\
\cmidrule(lr){2-4}\cmidrule(lr){5-9}
Model & OAC & net & $\tau_b$ & OAC & net & $\tau_b$ & WH & fcAUC & Mapping \\
\midrule
MERU-S      & .281 & .706 & .747 & .178 & .266 & .351 & .990 & .170 & 2 \\
MERU-B      & .266 & .728 & .741 & .133 & .043 & .121 & .990 & .374 & 2 \\
HyCoCLIP-B  & .195 & .189 & .360 & .132 & .044 & .103 & .935 & .053 & 2 \\
PHyCLIP-B   & .179 & .172 & .287 & .121 & .009 & .042 & .865 & .056 & 2 \\
\bottomrule
\end{tabular}}
\end{table}

The registered secondary image-source mode is weaker
(Table~\ref{tab:traversal_summary_image}). MERU-S, MERU-B, and HyCoCLIP-B pass
own-chain Tier~1; PHyCLIP-B is inconclusive because its net-progress interval
includes $0.12$. No image-source condition passes Tier~2. Under full-hierarchy
competition, MERU-S is Tier~1-inconclusive and the other three fail. These
results do not strengthen the primary text-source claim.

\begin{table}[htbp]
\centering
\caption{
Confirmatory secondary image-source semantic traversal on the same $200$
ImageNet/WordNet sources.  Columns follow Table~\ref{tab:traversal_summary}.
PHyCLIP-B own-chain net is $0.102\,[0.067,0.136]$, so Tier~1 is inconclusive;
MERU-S full-hierarchy OAC is $0.109\,[0.095,0.121]$, likewise yielding an
inconclusive Tier~1 decision.  ``Inc.'' denotes an inconclusive registered
decision, not a failure.
}
\label{tab:traversal_summary_image}
\resizebox{0.89\textwidth}{!}{%
\begin{tabular}{lccc ccccc cc}
\toprule
& \multicolumn{3}{c}{Own chain} & \multicolumn{5}{c}{Full hierarchy} & \multicolumn{2}{c}{Tier~1} \\
\cmidrule(lr){2-4}\cmidrule(lr){5-9}\cmidrule(lr){10-11}
Model & OAC & net & $\tau_b$ & OAC & net & $\tau_b$ & WH & fcAUC & Own & Full \\
\midrule
MERU-S      & .266 & .550 & .606 & .109 & .138 & .179 & .990 & .148 & Pass & Inc. \\
MERU-B      & .274 & .584 & .627 & .077 & .052 & .104 & .990 & .274 & Pass & Fail \\
HyCoCLIP-B  & .189 & .131 & .262 & .082 & .016 & .035 & .710 & .031 & Pass & Fail \\
PHyCLIP-B   & .175 & .102 & .156 & .062 & .001 & .003 & .760 & .043 & Inc. & Fail \\
\bottomrule
\end{tabular}}
\end{table}

The excluded HyCoCLIP-S pilot produced the same qualitative mapping:
own-chain Tier~1 detection without Tier~2 and full-hierarchy failure. Two
numerical-sensitivity runs (Lorentz $\epsilon=10^{-8}$ and float32 scoring)
caused no Tier-boundary reversals; full pilot values are provided in the
supplementary.

\FloatBarrier
\subsection{Additional fixed-pool readouts}
\label{app:terminal_collapse}

The following tables report additional fixed-pool measurements. Because they
depend on ties, pool composition, and shared endpoints, they are descriptive
rather than semantic pass/fail tests.

\begin{table}[htbp]
\centering
\caption{
Strict retrieved-norm decrease rates for current-GRIT baselines
($\eta=1.2$). Ties, pool composition, and the shared endpoint prevent a
calibrated semantic interpretation. PHyCLIP factor-wise results are reported
separately in Table~\ref{tab:phyclip_factor_traversal_full}.
}
\label{tab:traversal_full}
\begin{tabular}{lccc}
\toprule
Model & ImageNet & COCO & Flickr30k \\
\midrule
MERU-S & 15.7\% & 15.2\% & 13.7\% \\
MERU-B & 12.6\% & 14.5\% & 9.5\% \\
HyCoCLIP-S & 15.2\% & 18.0\% & 14.2\% \\
HyCoCLIP-B & 16.5\% & 18.9\% & 14.9\% \\
\bottomrule
\end{tabular}
\end{table}

\begin{table}[htbp]
\centering
\caption{
Strict retrieved-norm decrease rates across PHyCLIP's $64$ product factors for
the released PHyCLIP-B and PHyCLIP-L checkpoints. These fixed-pool values lack
a calibrated semantic reference.
}
\label{tab:phyclip_factor_traversal_full}
\begin{tabular}{llcc}
\toprule
Model & Dataset & Mean rate & Factor range \\
\midrule
PHyCLIP-B & ImageNet & 0.213 & [0.177, 0.258] \\
PHyCLIP-B & COCO & 0.256 & [0.218, 0.307] \\
PHyCLIP-B & Flickr30k & 0.251 & [0.205, 0.290] \\
PHyCLIP-L & ImageNet & 0.218 & [0.180, 0.266] \\
PHyCLIP-L & COCO & 0.259 & [0.227, 0.297] \\
PHyCLIP-L & Flickr30k & 0.252 & [0.213, 0.302] \\
\bottomrule
\end{tabular}
\end{table}

\FloatBarrier  %
\paragraph{Terminal retrieval records.}

The terminal nearest-neighbor records are also fixed-pool descriptions. In the COCO runs the candidate pool contains $25{,}014$ captions, whereas every trajectory uses the same exact-origin query at $t=1$.  Nearest neighbour retrieval at that endpoint therefore returns the same minimum-radius caption for every source (up to ties).  The resulting $1/200$ terminal count is a deterministic endpoint-and-pool consequence, not an independent traversal statistic.

Table~\ref{tab:terminal_collapse_exhibit_released} shows the stored records
step by step for released checkpoints. Selection and screening rules are given
in the caption. These examples visualize individual nearest-neighbour
sequences but are not used for calibrated inference. In particular, recurrence
of a caption across sources can reflect the shared candidate pool and endpoint
rule and is not interpreted as independent model evidence.

\begin{table}[htbp]
\centering
\caption{
Step-by-step retrieved captions on the \textbf{released} MERU-B and
HyCoCLIP-B checkpoints. The examples are the first two COCO val2017 source
images by index. Captions are truncated to $80$ characters and screened for
PII and junk (no redactions). ``Cos'' denotes cosine similarity to the source
image.  Step~20 is the common exact-origin query at $t=1$;
over the fixed $25{,}014$-caption pool this returns the same minimum-radius
caption for all $200$ sources (up to ties).  The table is a qualitative record,
not a statistical result.
}
\label{tab:terminal_collapse_exhibit_released}
\small
\begin{tabular}{llp{0.46\textwidth}rr}
\toprule
Model & Steps & Retrieved caption ($\le80$ chars) & Norm & Cos \\
\midrule
\multicolumn{5}{l}{\emph{Source A: COCO 000000465718.jpg (idx 0)}} \\
MERU-B & 0--4 & This workstation features three desktop monitors with a single keyboard\ldots & 0.571 & 0.790 \\
       & 5--8 & a laptop computer a keyboard and two monitors & 0.496 & 0.783 \\
       & 9--17 & Not the biggest workspace in the world, but it works & 0.396 & 0.749 \\
       & 18--19 & I do not know what this is supposed to be.. \emph{(pre-origin recurrence)} & 0.384 & 0.679 \\
       & 20 & I do not know what this is supposed to be.. \emph{(exact-origin retrieval)} & 0.384 & 0.679 \\
\addlinespace
HyCoCLIP-B & 0--4 & Two computer screens that are sitting on a desk. & 0.382 & 0.839 \\
           & 5 & A desk has two computer monitors, a keyboard, and a laptop that is all connected. & 0.369 & 0.842 \\
           & 6 & a desk with a monitor and a keyboard & 0.344 & 0.848 \\
           & 7--14 & a desk with a computer a laptop and monitor & 0.327 & 0.849 \\
           & 15--19 & a table that has some computers on it & 0.320 & 0.827 \\
           & 20 & Picture of living room with modern furniture and decor \emph{(exact-origin retrieval)} & 0.318 & 0.532 \\
\midrule
\multicolumn{5}{l}{\emph{Source B: COCO 000000014888.jpg (idx 1)}} \\
MERU-B & 0--2 & A young calf drinks from its mother's udders & 0.613 & 0.752 \\
       & 3--6 & A dairy cow is being milked by machine.. & 0.525 & 0.742 \\
       & 7--8 & A cow is being milked by a machine. & 0.488 & 0.727 \\
       & 9--19 & I do not know what this is supposed to be.. \emph{(pre-origin recurrence)} & 0.384 & 0.682 \\
       & 20 & I do not know what this is supposed to be.. \emph{(exact-origin retrieval)} & 0.384 & 0.682 \\
\addlinespace
HyCoCLIP-B & 0--6 & A dairy cow is being milked by machine.. & 0.391 & 0.828 \\
           & 7--13 & A dairy cow suffering in confinement hooked up to a milking machine. & 0.365 & 0.826 \\
           & 14--18 & a small calf nursing a cow in a pasture & 0.329 & 0.730 \\
           & 19 & a number of people in a bod of water & 0.318 & 0.568 \\
           & 20 & Picture of living room with modern furniture and decor \emph{(exact-origin retrieval)} & 0.318 & 0.417 \\
\bottomrule
\end{tabular}
\end{table}

The grounded Flickr30k quantitative run uses a different candidate construction:
$747{,}260$ mixed items (full images, box crops, captions, and an explicit
synthetic \texttt{[ROOT]}), $200$ image sources, and $50$ interpolation points.
All exact-root queries retrieve the inserted \texttt{[ROOT]}.  Looking instead
at each trajectory's last non-root, non-image retrieved caption gives $13$
distinct captions; the largest caption accounts for $48\%$ of sources and the
largest two for $79\%$.  These are fixed-pool descriptive summaries.  The
separate Flickr exhibit below uses the same mixed pool but a $21$-point run.

The native NG2 record is different again: $935$ fixed registry-v2 relations are
queried against a shared $5{,}000$-caption COCO pool with $10$ query points.  Its
$661$--$715$ distinct retrieved endpoints and ${\le}2.9\%$ largest-endpoint
share are likewise conditional descriptions, not a calibrated comparison
against the COCO or Flickr constructions.

\begin{table}[htbp]
\centering
\caption{
Fixed-pool terminal summaries under three non-comparable constructions. COCO
and Flickr use released HyCoCLIP-B; COCO's $t=1$ endpoint is deterministic,
and Flickr counts the last non-root, non-image captions. NG2 reports the range
across the audited registry-v2 rows. These descriptions have no common
calibrated scale.
}
\label{tab:terminal_collapse_pools}
\resizebox{\textwidth}{!}{%
\begin{tabular}{lllrrl}
\toprule
Retrieval construction & Candidate pool & Trajectories & Pool size & Distinct endpoints & Largest share \\
\midrule
COCO, 21 points & captions only & $200$ & $25{,}014$ & $1/200$ at $t=1$ & $100\%$ at $t=1$ \\
Flickr30k, 50 points & image/box/caption/root & $200$ & $747{,}260$ & $13/200$ last captions & $48\%$ (top-2 $79\%$) \\
NG2, 10 points & shared COCO captions & $935$ & $5{,}000$ & $661$--$715/935$ & ${\le}2.9\%$ \\
\bottomrule
\end{tabular}}
\end{table}

\begin{table}[htbp]
\centering
\caption{
Qualitative $21$-point trajectories for released HyCoCLIP-B over the
$747{,}260$-item grounded Flickr30k mixed pool, separate from the $50$-point
summary in Table~\ref{tab:terminal_collapse_pools}. We show the first two
full-image sources by index, with captions truncated to $80$ characters.
``Norm'' is space-norm and ``Cos'' is source cosine. The exact-root query
retrieves the inserted \texttt{[ROOT]}. This exhibit is illustrative.
}
\label{tab:terminal_collapse_exhibit_flickr}
\small
\begin{tabular}{lp{0.50\textwidth}rr}
\toprule
Steps & Retrieved item ($\le80$ chars) & Norm & Cos \\
\midrule
\multicolumn{4}{l}{\emph{Source A: grounded Flickr30k 3359636318.jpg (idx 517)}} \\
0--6   & $\langle$source image$\rangle$ & 0.641 & 1.000 \\
7--8   & People walk down a city street past a record store. & 0.396 & 0.868 \\
9      & A picture of a storefront with a few people passing by. & 0.370 & 0.853 \\
10     & passersby stare & 0.294 & 0.816 \\
11--12 & Passersby interact & 0.272 & 0.806 \\
13--16 & The \emph{(onset/hub)} & 0.179 & 0.712 \\
17--20 & \texttt{[ROOT]} & 0.000 & 0.000 \\
\midrule
\multicolumn{4}{l}{\emph{Source B: grounded Flickr30k 6959556104.jpg (idx 527)}} \\
0--7   & $\langle$source image$\rangle$ & 0.641 & 1.000 \\
8--9   & Protesters with a sign reading ``ASTI, Save Our Schools'' march outside. & 0.346 & 0.825 \\
10--13 & a protest & 0.260 & 0.792 \\
14--15 & The \emph{(onset/hub)} & 0.179 & 0.687 \\
16--20 & \texttt{[ROOT]} & 0.000 & 0.000 \\
\bottomrule
\end{tabular}
\end{table}

\FloatBarrier  %
\section{Taxonomy Mapping and Robustness}
\label{app:taxonomy}

\subsection{CIFAR-100 to WordNet mapping}
\label{app:taxonomy_mapping}

\begin{table}[htbp]
\centering
\caption{
Manually disambiguated CIFAR-100 to WordNet synset mapping used in the main
taxonomy analysis (Table~\ref{tab:taxonomy_decomp}). For 13 classes, the naive
top-3-synset heuristic (described in Table~\ref{tab:cifar_mapping_robustness})
includes semantically incorrect senses---in several cases excluding the correct
sense entirely---so we manually pin each to the single sense below. The
remaining 87/100 CIFAR classes retain their naive candidate sets.
}
\label{tab:cifar_manual_mapping}
\small
\renewcommand{\arraystretch}{0.9}
\begin{tabular}{lll}
\toprule
CIFAR label    & Manual synset      & Sense / reason \\
\midrule
seal           & \texttt{seal.n.09}          & marine mammal, not sealing wax \\
ray            & \texttt{ray.n.07}           & cartilaginous fish, not light beam \\
turtle         & \texttt{turtle.n.02}        & aquatic reptile, not sweater \\
skunk          & \texttt{skunk.n.04}         & musteline mammal, not pejorative person \\
whale          & \texttt{whale.n.02}         & cetacean, not fictional giant \\
dolphin        & \texttt{dolphin.n.02}       & toothed whale, not dolphinfish \\
sweet pepper   & \texttt{bell\_pepper.n.02}  & bell pepper vegetable, not spice \\
plate          & \texttt{plate.n.04}         & dinner dish, not baseball home plate \\
maple tree     & \texttt{maple.n.02}         & \emph{Acer} tree, not generic tree \\
oak tree       & \texttt{oak.n.02}           & \emph{Quercus} tree, not generic tree \\
palm tree      & \texttt{palm.n.03}          & \emph{Palmae} tree, not generic tree \\
pine tree      & \texttt{pine.n.01}          & coniferous tree, not generic tree \\
willow tree    & \texttt{willow.n.01}        & \emph{Salix} tree, not generic tree \\
\bottomrule
\end{tabular}
\end{table}

\begin{table}[htbp]
\centering
\caption{
Robustness of the taxonomy decomposition to WordNet mapping. We compare the naive last-token mapping (minimum shortest path over the top-three noun-synset candidates) with the manual mapping of Table~\ref{tab:cifar_manual_mapping} on the same $100$ CIFAR classes. Current-GRIT ViT-B rows average $r$ over seeds $0/37/42$ and report Mantel $p$ ranges. Manual disambiguation raises $r$ by $0.01$--$0.10$ while preserving the broad hyperbolic~$>$~Euclidean pattern. Euclidean CLIP outputs are L2-normalized, so their norm-increment Mantel tests are undefined (N/A), while their angular taxonomy correlations remain valid. \rev{Among the reported condition-level summaries with non-degenerate norms (averaging the three ViT-B seeds),} $\Delta R^2_{\mathrm{norm}}\leq0.003$, and no manual-mapping test is significant ($p\geq0.101$). One of the $36$ current-GRIT per-seed tests crosses $0.05$ under the naive mapping without correction. $\Delta r$ uses unrounded values.}
\label{tab:cifar_mapping_robustness}
\small
\renewcommand{\arraystretch}{0.9}
\begin{tabular}{llrrrrr}
\toprule
Model & Setting & naive $r$ & manual $r$ & $\Delta r$ & $p_{\mathrm{perm}}$ (naive) & $p_{\mathrm{perm}}$ (manual) \\
\midrule
CLIP-S      & released & 0.359 & 0.377 & +0.018 & N/A & N/A \\
CLIP-B      & released & 0.358 & 0.370 & +0.012 & N/A & N/A \\
CLIP-L      & released & 0.336 & 0.363 & +0.027 & N/A & N/A \\
\midrule
MERU-S      & released & 0.448 & 0.488 & +0.040 & 0.125 & 0.724 \\
MERU-B      & released & 0.428 & 0.489 & +0.062 & 0.978 & 0.340 \\
MERU-L      & released & 0.409 & 0.446 & +0.037 & 0.467 & 0.977 \\
HyCoCLIP-S  & released & 0.376 & 0.449 & +0.073 & 0.694 & 0.531 \\
HyCoCLIP-B  & released & 0.400 & 0.465 & +0.065 & 0.677 & 0.501 \\
PHyCLIP-B   & released & 0.384 & 0.456 & +0.072 & 0.682 & 0.735 \\
PHyCLIP-L   & released & 0.405 & 0.456 & +0.051 & 0.910 & 0.499 \\
\midrule
MERU-S      & baseline & 0.432 & 0.492 & +0.060 & 0.559 & 0.666 \\
MERU-S      & clampOff & 0.452 & 0.508 & +0.056 & 0.888 & 0.385 \\
HyCoCLIP-S  & baseline & 0.397 & 0.448 & +0.051 & 0.374 & 0.701 \\
HyCoCLIP-S  & clampOff & 0.395 & 0.491 & +0.096 & 0.413 & 0.301 \\
PHyCLIP-S   & baseline & 0.411 & 0.465 & +0.054 & 0.894 & 0.538 \\
PHyCLIP-S   & clampOff & 0.416 & 0.493 & +0.077 & 0.812 & 0.913 \\
\midrule
MERU-B      & baseline & 0.398 & 0.459 & +0.061 & 0.23--0.77 & 0.10--0.80 \\
MERU-B      & clampOff & 0.422 & 0.485 & +0.063 & 0.56--0.75 & 0.12--0.49 \\
HyCoCLIP-B  & baseline & 0.397 & 0.462 & +0.065 & 0.01--0.64 & 0.12--0.86 \\
HyCoCLIP-B  & clampOff & 0.432 & 0.508 & +0.076 & 0.30--0.64 & 0.31--0.56 \\
PHyCLIP-B   & baseline & 0.395 & 0.471 & +0.076 & 0.16--0.80 & 0.23--0.67 \\
PHyCLIP-B   & clampOff & 0.416 & 0.479 & +0.064 & 0.21--0.96 & 0.34--0.90 \\
\bottomrule
\end{tabular}
\end{table}

\clearpage  %
\subsection{Full taxonomy-distance correlation results}

\begin{table}[htbp]
\centering
\caption{
CIFAR-100 WordNet path-distance correlation and angular/radial decomposition across all evaluated configurations, using the manually disambiguated mapping of Table~\ref{tab:cifar_manual_mapping}. Columns report Pearson $r$, the cosine-only and norm-only $R^2$, the incremental $\Delta R^2_{\mathrm{norm}} =R^2_{\cos+\mathrm{norm}}-R^2_{\cos}$, and its Mantel $p_{\mathrm{perm}}$. Euclidean CLIP outputs are L2-normalized, leaving norms constant up to floating-point ulps; its norm-only, incremental, and Mantel cells are therefore undefined and reported as N/A. Its taxonomy $r$ and $R^2_{\cos}$ remain valid. Across the evaluated sizes, hyperbolic checkpoints have higher $r$ than Euclidean CLIP, but no defined manual-mapping Mantel test detects a unique norm contribution beyond cosine ($p_{\mathrm{perm}}>0.05$ throughout). This estimand does not exclude radial information redundant with angular structure. The largest reported hyperbolic seed-mean increment is MERU-B baseline ($\Delta R^2_{\mathrm{norm}}=0.0029$); its underlying per-seed values range from $0.0001$ to $0.0054$. Current-GRIT ViT-B rows average effect sizes over seeds $0/37/42$, while their $p$ columns report the per-seed min--max range. Other rows are single checkpoints. The hyperbolic--Euclidean separation in $r$ is preserved under the naive mapping (Table~\ref{tab:cifar_mapping_robustness}).
}
\label{tab:taxonomy_all_models}
\begin{tabular}{llccccc}
\toprule
Model & Setting & Taxonomy $r$ & $R^2_{\cos}$ & $R^2_{\mathrm{norm\text{-}only}}$ & $\Delta R^2_{\mathrm{norm}}$ & $p_{\mathrm{perm}}$ \\
\midrule
CLIP-S      & released & 0.377 & 0.142 & N/A & N/A & N/A \\
CLIP-B      & released & 0.370 & 0.137 & N/A & N/A & N/A \\
CLIP-L      & released & 0.363 & 0.132 & N/A & N/A & N/A \\
\midrule
MERU-S      & released & 0.488 & 0.238 & 0.001 & $+0.0002$ & 0.724 \\
MERU-B      & released & 0.489 & 0.239 & 0.001 & $+0.0013$ & 0.340 \\
MERU-L      & released & 0.446 & 0.199 & 0.001 & $+0.0000$ & 0.977 \\
HyCoCLIP-S  & released & 0.449 & 0.201 & 0.002 & $+0.0007$ & 0.531 \\
HyCoCLIP-B  & released & 0.465 & 0.217 & 0.005 & $+0.0010$ & 0.501 \\
PHyCLIP-B   & released & 0.456 & 0.208 & 0.002 & $+0.0002$ & 0.735 \\
PHyCLIP-L   & released & 0.456 & 0.208 & 0.003 & $+0.0009$ & 0.499 \\
\midrule
MERU-S      & baseline & 0.492 & 0.242 & 0.002 & $+0.0003$ & 0.666 \\
MERU-S      & clampOff & 0.508 & 0.258 & 0.000 & $+0.0013$ & 0.385 \\
HyCoCLIP-S  & baseline & 0.448 & 0.200 & 0.000 & $+0.0002$ & 0.701 \\
HyCoCLIP-S  & clampOff & 0.491 & 0.241 & 0.000 & $+0.0019$ & 0.301 \\
PHyCLIP-S   & baseline & 0.465 & 0.216 & 0.003 & $+0.0007$ & 0.538 \\
PHyCLIP-S   & clampOff & 0.493 & 0.243 & 0.002 & $+0.0000$ & 0.913 \\
\midrule
MERU-B      & baseline & 0.459 & 0.210 & 0.002 & $+0.0029$ & 0.10--0.80 \\
MERU-B      & clampOff & 0.485 & \rev{0.235} & 0.000 & $+0.0021$ & 0.12--0.49 \\
HyCoCLIP-B  & baseline & 0.462 & 0.214 & 0.001 & $+0.0018$ & 0.12--0.86 \\
HyCoCLIP-B  & clampOff & 0.508 & 0.258 & 0.001 & $+0.0012$ & 0.31--0.56 \\
PHyCLIP-B   & baseline & 0.471 & 0.222 & 0.003 & $+0.0016$ & 0.23--0.67 \\
PHyCLIP-B   & clampOff & 0.479 & 0.230 & 0.002 & $+0.0006$ & 0.34--0.90 \\
\bottomrule
\end{tabular}
\end{table}

\FloatBarrier
\subsection{Radial parent-child ordering diagnostics}
\label{app:radial_full}

Table~\ref{tab:parent_child_full} reports the radial parent--child ordering
values behind Figure~\ref{fig:radial}, extending the analysis across scales and
to the current-GRIT interventions in all three families. Directed pairs use
the CIFAR-100 fine$\to$coarse label hierarchy, not the WordNet mapping of
Table~\ref{tab:cifar_manual_mapping}, which is used only for the
taxonomy-distance analysis. No released hyperbolic checkpoint reaches the
positive detection threshold. Among the three-seed ViT-B interventions, no
positive crossing is replicated across all three seeds. We do not apply the
single-cell $z$ threshold to seed-averaged $z$-scores because no seed-aware
condition-level null was calibrated. The only seed-consistent departure is PHyCLIP-B
baseline in the reverse direction, which is not evidence for the hypothesized
ordering. ViT-S interventions are single-seed and exploratory; the largest
positive excursion, PHyCLIP-S clampOff ($z=+2.73$), is one such unreplicated
cell.

\begin{table}[htbp]
\centering
\caption{
Full radial parent--child ordering diagnostics. ``Radial consistency'' is the fraction of the $100$ directed CIFAR-100 fine-to-superclass pairs for which the child norm exceeds the parent norm. The shuffle preserves the fine- and coarse-level norm marginals, and $z=(\mathrm{real}-\mathrm{null})/\mathrm{s.d.}$ is computed from $10{,}000$ permutations. Thus $z$ tests pair-specific radial alignment beyond a global level separation; it does not exclude a purely level-wise radial code redundant with angular hierarchy. The directional detection threshold is $z\geq+1.6$; $z\leq-1.6$ denotes a reverse-direction excursion and is not evidence for the hypothesis. Threshold calibration and empirical false-positive rates are reported in Appendix~\ref{app:sensitivity}. No positive current-GRIT cell survives a one-sided Bonferroni reference over the $24$ runs ($z\approx2.9$). Euclidean CLIP outputs are L2-normalized, so their norm ordering, shuffle null, and $z$-score are tie-degenerate and reported as N/A. ViT-S intervention rows are single-seed and exploratory.
}
\label{tab:parent_child_full}
\resizebox{\textwidth}{!}{%
\begin{tabular}{llccccl}
\toprule
Model & Setting & Scale & Radial cons. & Shuffle null & $z$ & Interpretation \\
\midrule
Euclidean CLIP & released & ViT-S & N/A & N/A & N/A & unit-norm/tie-degenerate \\
Euclidean CLIP & released & ViT-B & N/A & N/A & N/A & unit-norm/tie-degenerate \\
Euclidean CLIP & released & ViT-L & N/A & N/A & N/A & unit-norm/tie-degenerate \\
MERU           & released & ViT-S & 63\% & $62.1 \pm 3.5$ & $+0.25$ & near null \\
MERU           & released & ViT-B & 61\% & $61.0 \pm 3.7$ & $-0.00$ & near null \\
MERU           & released & ViT-L & 62\% & $62.3 \pm 3.5$ & $-0.09$ & near null \\
HyCoCLIP       & released & ViT-S & 56\% & $55.9 \pm 2.8$ & $+0.04$ & near null \\
HyCoCLIP       & released & ViT-B & 35\% & $37.1 \pm 3.3$ & $-0.64$ & near null \\
PHyCLIP        & released & ViT-B & 42\% & $44.2 \pm 3.3$ & $-0.67$ & near null \\
PHyCLIP        & released & ViT-L & 18\% & $22.9 \pm 3.1$ & $-1.59$ & near null \\
\midrule
MERU           & baseline (s0)  & ViT-B & 82\% & $78.1 \pm 3.0$ & $+1.31$ & near null \\
MERU           & baseline (s37) & ViT-B & 84\% & $77.1 \pm 3.5$ & $+1.99$ & above null \\
MERU           & baseline (s42) & ViT-B & 85\% & $83.1 \pm 3.0$ & $+0.66$ & near null \\
MERU           & clampOff (s0)  & ViT-B & 80\% & $74.9 \pm 3.4$ & $+1.50$ & near null \\
MERU           & clampOff (s37) & ViT-B & 82\% & $79.6 \pm 3.1$ & $+0.77$ & near null \\
MERU           & clampOff (s42) & ViT-B & 83\% & $80.5 \pm 3.5$ & $+0.72$ & near null \\
MERU           & baseline (s0)  & ViT-S & 77\% & $73.6 \pm 3.4$ & $+0.98$ & near null \\
MERU           & clampOff (s0)  & ViT-S & 69\% & $69.5 \pm 3.5$ & $-0.14$ & near null \\
\addlinespace[2pt]
HyCoCLIP       & baseline (s0)  & ViT-B & 38\% & $35.9 \pm 3.1$ & $+0.70$ & near null \\
HyCoCLIP       & baseline (s37) & ViT-B & 34\% & $39.4 \pm 3.1$ & $-1.73$ & below null \\
HyCoCLIP       & baseline (s42) & ViT-B & 22\% & $28.2 \pm 3.2$ & $-1.97$ & below null \\
HyCoCLIP       & clampOff (s0)  & ViT-B & 74\% & $75.8 \pm 3.5$ & $-0.52$ & near null \\
HyCoCLIP       & clampOff (s37) & ViT-B & 67\% & $70.9 \pm 3.4$ & $-1.17$ & near null \\
HyCoCLIP       & clampOff (s42) & ViT-B & 63\% & $61.1 \pm 3.7$ & $+0.51$ & near null \\
HyCoCLIP       & baseline (s0)  & ViT-S & 47\% & $48.9 \pm 3.3$ & $-0.58$ & near null \\
HyCoCLIP       & clampOff (s0)  & ViT-S & 59\% & $62.4 \pm 3.5$ & $-0.97$ & near null \\
\addlinespace[2pt]
PHyCLIP        & baseline (s0)  & ViT-B & 17\% & $23.7 \pm 2.9$ & $-2.34$ & below null \\
PHyCLIP        & baseline (s37) & ViT-B & 17\% & $24.2 \pm 3.6$ & $-2.00$ & below null \\
PHyCLIP        & baseline (s42) & ViT-B & 24\% & $29.9 \pm 3.1$ & $-1.93$ & below null \\
PHyCLIP        & clampOff (s0)  & ViT-B & 63\% & $58.5 \pm 2.9$ & $+1.57$ & near null \\
PHyCLIP        & clampOff (s37) & ViT-B & 62\% & $56.6 \pm 3.3$ & $+1.65$ & above null \\
PHyCLIP        & clampOff (s42) & ViT-B & 64\% & $62.8 \pm 3.3$ & $+0.37$ & near null \\
PHyCLIP        & baseline (s0)  & ViT-S & 41\% & $39.0 \pm 3.3$ & $+0.62$ & near null \\
PHyCLIP        & clampOff (s0)  & ViT-S & 69\% & $59.5 \pm 3.5$ & $+2.73$ & above null \\
\bottomrule
\end{tabular}}
\end{table}

\FloatBarrier
\subsection{Mantel permutation and prompt robustness}

We use Mantel-style permutations because pairwise class distances are not independent. For the native directed test we additionally report a length-matched subset and a length-residualized variant (Section~\ref{sec:parent_child}, Appendix~\ref{app:native_radial}). Across those controls the reported linear norm contribution beyond cosine remains numerically small.

\FloatBarrier
\subsection{Directed radial test on the GRIT box-to-full-caption relation (NG2)}
\label{app:native_radial}

Table~\ref{tab:native_radial} reports the native-relation radial test of
Section~\ref{sec:parent_child}. Each pair consists of a full-image caption
(the predicted more-specific, larger-norm item) and a box caption (the
predicted more-general, smaller-norm item). The directed statistic asks whether
true pairs satisfy this ordering more often than a sample-level re-pairing null
that preserves the full/box norm marginals. It therefore measures a
pair-specific one-bit excess, not a purely marginal full-versus-box separation.

All NG2 results in this subsection are conditional on the fixed $935$-pair
registry; the retrieval readouts additionally condition on a shared
$5{,}000$-caption candidate pool.  The analysis grid and tie rules were frozen
before the final readout, but the collection has not
received an end-to-end size-and-power calibration.  We therefore retain its
statistics, permutation values, and intervals as uncalibrated descriptive
summaries rather than confirmatory tests.

Because this relation has no tree metric, the decomposition instead predicts a
binary same-sample indicator
($y{=}1$ iff a full and box caption come from the same GRIT sample) from cosine
distance and norm difference. Accordingly,
$\Delta R^2_{\mathrm{norm}}$ measures the norm term's unique contribution beyond
cosine to identifying true pairs; it is not a direct depth readout and does not
exclude norm information redundant with angle. Directed $z$ is positive for
every box-trained checkpoint and for MERU-B. Euclidean CLIP-B's L2-normalized
outputs make the norm-ordering and norm-increment quantities undefined. The
recorded unique norm increments range from $0.00001$ to $0.00025$ on released
checkpoints and from $0.00002$ to $0.00228$ under clampOff; angular distance
accounts for most of the fitted variation in this fixed-registry readout.

Three bounds scope this reading. The released-versus-clampOff difference is cross-sectional, not a matched intervention, so it cannot determine whether removing the floor reveals or rescales the supplement. The sample-level null preserves the full/box norm marginals, so a within-sample \emph{visual}-specificity gap that length-matching does not capture could produce the norm ordering without any graded depth coordinate. And the positive directed $z$ records only a single conditional bit---that a full caption tends to sit outside its own box (MERU-B shows it too at $z=+5.3$ despite no box supervision)---not the graded axis the mechanism claims. Neither the one-bit excess nor the numerically small linear increment establishes an operative graded axis.

\begin{table}[htbp]
\centering
\caption{
Conditional NG2 radial test on registry-v2 ($n=935$; $65$ byte-identical pairs removed). Directed $z$ compares true pairs with $R=500$ sample-level re-pairings; the length-matched subset has $n=92$. $\Delta R^2_{\mathrm{norm}}$ is the norm-difference increment beyond cosine for predicting same-sample pairs; Mantel $p$ uses the first $500$ pairs and $9{,}999$ permutations. CLIP-B is N/A because unit norms make the radial quantities tie-degenerate. Results are descriptive and conditional on this registry.
}
\label{tab:native_radial}
\resizebox{0.90\textwidth}{!}{%
\begin{tabular}{llcccc}
\toprule
Model & seed & $z$ (full, $n{=}935$) & $z$ (len-matched, $n{=}92$) & $\Delta R^2_{\mathrm{norm}}$ & Mantel $p$ \\
\midrule
HyCoCLIP-B released & ---  & $+6.39$ & $+4.54$ & $0.00005$ & $0.0003$ \\
HyCoCLIP-B clampOff & 0    & $+5.60$ & $+5.74$ & $0.00174$ & $0.0001$ \\
                    & 37   & $+5.65$ & $+5.00$ & $0.00228$ & $0.0001$ \\
                    & 42   & $+6.24$ & $+5.81$ & $0.00002$ & $0.0084$ \\
\addlinespace[2pt]
PHyCLIP-B released  & ---  & $+6.07$ & $+5.61$ & $0.00001$ & $0.0565$ \\
PHyCLIP-B clampOff  & 0    & $+4.67$ & $+4.48$ & $0.00030$ & $0.0001$ \\
                    & 37   & $+5.45$ & $+4.50$ & $0.00041$ & $0.0001$ \\
                    & 42   & $+5.15$ & $+4.24$ & $0.00021$ & $0.0001$ \\
\addlinespace[2pt]
MERU-B released     & ---  & $+5.25$ & $+1.74$ & $0.00025$ & $0.0001$ \\
CLIP-B (Euclidean)  & ---  & N/A & N/A & N/A & N/A \\
\bottomrule
\end{tabular}}
\end{table}

\paragraph{Traversal along the native radial direction.}
Table~\ref{tab:native_traversal} reports a descriptive fixed-pool counterpart
to the directed test. For each NG2 pair, let $F$ and $B$ denote the full- and
box-caption embeddings, with radii $\rho_F$ and $\rho_B$. The radial ray fixes
the full-caption direction $\hat{x}_F=x_F/\rho_F$ and varies the norm linearly
from $\rho_B$ to $\rho_F$ over $10$ steps ($n=935$ registry-v2 pairs). Thus
the path uses the box-caption norm but not its angular position. At each step,
we retrieve a caption and test whether its norm increases in the predicted
general-to-specific direction.

No hyperbolic model satisfies the strict all-steps criterion. With $10$ points
there are nine transitions, and the heuristic independent fair-sign reference
is $935/2^9=1.83$ perfect paths. Ties and within-trajectory dependence mean
that this is not a calibrated null; the observed $0/935$ is reported
descriptively. Terminal diversity is $661$--$715$ distinct endpoints with
largest shares at most $2.9\%$. Endpoint cosine is higher under clampOff
($0.92$ versus $0.72$--$0.76$ for released checkpoints), but endpoint
proximity does not establish step-wise ordering. CLIP-B's unit-norm outputs
make the proposed radial ray and its direction precheck tie-degenerate, so the
entire Euclidean radial row is N/A. The graded hyperbolic readouts below are
reported conditionally on this fixed NG2 registry and candidate pool.

\begin{table}[htbp]
\centering
\caption{
Radial-ray NG2 traversal over $10$ points. Direction precheck tests $\rho_F>\rho_B$, and strict monotonicity requires all nine retrieved-norm transitions to increase. Endpoint cosine is measured against $F$; terminal columns report the number of distinct endpoints and largest terminal share. Hyperbolic rows use cosine retrieval; native-distance strict monotonicity is also $0\%$ (Table~\ref{tab:native_graded}). CLIP-B is N/A because unit norms define no radial interval. These fixed-pool statistics are descriptive.
}
\label{tab:native_traversal}
\resizebox{\textwidth}{!}{%
\begin{tabular}{lccccc}
\toprule
Model & Dir.\ precheck & Strict mono.\ & Endpoint $\cos(\cdot,F)$ & Distinct term.\ & Top-1 share \\
\midrule
HyCoCLIP-B released & $97.8\%$ & $0\%$ & $0.72$ & $709/935$ & $1.1\%$ \\
PHyCLIP-B released  & $98.6\%$ & $0\%$ & $0.76$ & $710/935$ & $1.1\%$ \\
MERU-B released     & $83.1\%$ & $0\%$ & $0.72$ & $677/935$ & $1.6\%$ \\
HyCoCLIP-B clampOff & $98.4\%$ & $0\%$ & $0.92$ & $661/935$ & $2.5\%$ \\
PHyCLIP-B clampOff  & $97.8\%$ & $0\%$ & $0.92$ & $700/935$ & $2.9\%$ \\
MERU-B clampOff     & $89.1\%$ & $0\%$ & $0.74$ & $715/935$ & $2.2\%$ \\
CLIP-B (Euclidean)  & N/A & N/A & N/A & N/A & N/A \\
\bottomrule
\end{tabular}}
\end{table}

\paragraph{Graded readouts and controls.}
Because strict monotonicity is a knife-edge statistic, we also measure a
graded step--specificity rank correlation. For each pair, $r_{\mathrm{s}}$ is the
Spearman correlation between step index and
$\cos(\mathrm{retrieved},F)-\cos(\mathrm{retrieved},B)$, averaged over the
$935$ registry-v2 pairs. This is a pair-relative angular specificity proxy,
not ground-truth semantic depth. For the hyperbolic models, the frozen grid crosses
\{radial ray, exact Lorentz geodesic\} with
\{cosine, model-native distance\} retrieval. For Euclidean CLIP-B, the
geodesic cell uses $q(t)=(1-t)B+tF$ with cosine or Euclidean-distance
retrieval. The grid also includes a shuffle-step null
($200$ reshuffles, $95\%$ band), a mismatched-target control, a norm-only
component control, and paired
bootstrap intervals ($B=10{,}000$; Table~\ref{tab:native_graded}). These
source-pair bootstrap intervals describe variation conditional on the fixed
registry, embeddings, and candidate pool; their rejection behavior has not
been calibrated.

For the hyperbolic rows, cosine retrieval along the radial ray is invariant to
norm scaling, so
$r_{\mathrm{s}}=0.000$ under fp64. This is a methods fact rather than model evidence
(fp16 leaves tie-breaking residuals of approximately $\pm0.02$).
Native-distance retrieval responds numerically to radial movement: every
hyperbolic model's observed $r_{\mathrm{s,true}}$ ($+0.023$--$+0.132$) lies
above its recorded step-shuffle band. However, no dist--norm-only gap interval
has a lower bound above zero; MERU-B released is closest at
$[-0.008,+0.053]$. We label this a conditional descriptive
non-detection, not a calibrated test of absence or a validated rejection
decision. The comparison does not distinguish a learned radial code
redundant with angular structure from a norm--specificity association in the
retrieval pool.

Under the model-native geodesic controls, $r_{\mathrm{s}}$ is positive under both retrieval
metrics ($0.625$--$0.723$ for the hyperbolic models), but the controls show
why this movement is not specific to hyperbolic hierarchy. Mismatched targets
also have positive correlations ($0.343$--$0.513$), and the true-pair excess
is $0.21$--$0.29$. Euclidean CLIP-B's straight-line path attains the largest
correlation under both metrics ($0.775$); its radial-ray cells are
undefined because its output norms are constant. The evaluated readouts therefore do
not establish a model-specific, operational hyperbolic hierarchy: radial
movement is reproduced by norm proximity, while positive geodesic movement is
generic to interpolation. The norm-only walk is a component control, not the
model's deployed retrieval metric.

\begin{table}[htbp]
\centering
\caption{
Conditional NG2 graded readout on registry-v2 ($n=935$) with a shared $5{,}000$-caption pool. Entries are mean per-pair $r_{\mathrm{s}}$ between step index and $\cos(\mathrm{retrieved},F)-\cos(\mathrm{retrieved},B)$. Radial gap intervals compare native-distance with norm-only retrieval; geodesic intervals compare true with mismatched targets. PHyCLIP native distance averages its $64$ factors. CLIP-B radial cells are N/A because of unit norms, while its Euclidean straight-line cells remain defined. Results are descriptive; bootstrap rejection behavior is not end-to-end calibrated.
}
\label{tab:native_graded}
\resizebox{\textwidth}{!}{%
\begin{tabular}{lcccc ccc}
\toprule
& \multicolumn{4}{c}{Radial ray} & \multicolumn{3}{c}{Geodesic} \\
\cmidrule(lr){2-5}\cmidrule(lr){6-8}
Model & cos & native dist & norm-only & gap CI (dist$-$norm) & cos & native dist & true$-$mismatch [95\% CI] \\
\midrule
CLIP-B (Euclidean)  & N/A & N/A & N/A & N/A & $\mathbf{0.775}$ & $\mathbf{0.775}$ & $+0.166\;[+.133,+.198]$ \\
MERU-B released     & $0.000$ & $+0.132$ & $+0.110$ & $[-.008,+.053]$ & $0.723$ & $0.677$ & $+0.256\;[+.219,+.293]$ \\
MERU-B clampOff     & $0.000$ & $+0.023$ & $+0.071$ & $[-.074,-.021]$ & $0.680$ & $0.625$ & $+0.221\;[+.183,+.260]$ \\
PHyCLIP-B clampOff  & $0.000$ & $+0.076$ & $+0.131$ & $[-.094,-.017]$ & $0.644$ & $0.646$ & $+0.273\;[+.235,+.312]$ \\
PHyCLIP-B released  & $0.000$ & $+0.088$ & $+0.085$ & $[-.036,+.041]$ & $0.643$ & $0.631$ & $+0.287\;[+.247,+.328]$ \\
HyCoCLIP-B clampOff & $0.000$ & $+0.109$ & $+0.313$ & $[-.239,-.170]$ & $0.642$ & $0.633$ & $+0.249\;[+.212,+.287]$ \\
HyCoCLIP-B released & $0.000$ & $+0.047$ & $+0.072$ & $[-.060,+.010]$ & $0.639$ & $0.627$ & $+0.248\;[+.208,+.286]$ \\
\bottomrule
\end{tabular}}
\end{table}

\FloatBarrier
\section{Diagnostic Positive Controls}

\label{app:positive_controls}

We use a synthetic positive control to verify that the diagnostics respond
when radial and pair-specific hierarchy is planted. We construct a balanced
tree with branching factor $4$ and depth $6$, yielding $5{,}461$ nodes and
$5{,}460$ parent--child edges. Nodes are embedded in the Poincar\'e disk with
radius increasing with depth, and angular sectors are assigned so that each
child lies inside its parent's sector.

Table~\ref{tab:positive_control_tree} shows that the diagnostics recover this
structure. Parent--child radial ordering, depth--radius correlation, and chain
monotonicity are perfect in the planted tree. Radius shuffling removes the
depth--radius relation, while angle shuffling removes the pair-specific sector
signal.

The radial pairing null remains high because shuffled children are still marginally deeper than parents. The pair gap therefore tests alignment beyond that level marginal. This control does not adjudicate a purely level-wise radial code, which may be learned or may reflect prompt statistics; it verifies the narrower point that the diagnostics detect planted pair-specific radial and sector structure. This balanced radial tree is distinct from the semantic traversal toy in Appendix~\ref{app:semantic_traversal_registration}. The former checks radial, taxonomy, sector, and tree-chain diagnostics; the latter checks the source-conditioned gold-ancestor traversal scorer and false-hub control.

\begin{table}[htbp]
\centering
\caption{
Synthetic positive control. The diagnostics recover planted radial ordering,
depth--radius correlation, monotonic chains, and angular-sector containment.
Radius and angle shuffling remove the corresponding signals. Pair gap is real
radial order minus its shuffled-pairing null; sector gap is defined
analogously for containment.
}
\label{tab:positive_control_tree}
\resizebox{\textwidth}{!}{%
\begin{tabular}{lcccccc}
\toprule
Embedding & Radial order & Pair gap & Depth--radius Spearman & Chain mono. & Sector gap & $\Delta R^2_{\mathrm{norm}}$ \\
\midrule
Synthetic radial tree & 1.000 & 0.200 & 1.000 & 1.000 & 0.996 & 0.232 \\
Radius-shuffled control & 0.498 & -- & -0.008 & 0.528 & -- & -- \\
Angle-shuffled control & -- & -- & -- & -- & 0.000 & -- \\
\bottomrule
\end{tabular}}
\end{table}

\FloatBarrier

\subsection{Sensitivity of the shuffle-controlled diagnostics}
\label{app:sensitivity}

The control above establishes non-blindness at a perfectly ordered point. We next estimate the effect size detected with $80\%$ probability. Parts~A and B target controlled increments: the radial test uses
\[
\Delta_{\mathrm{pair}}
=
\mathrm{Order}_{\mathrm{real}}
-
\mathbb{E}_{\mathrm{shuffle}}
[\mathrm{Order}_{\mathrm{shuffle}}],
\]
and the taxonomy test uses the incremental
$\Delta R^2_{\mathrm{norm}}$ beyond cosine. A signal confined to the preserved fine/coarse marginals, or redundant with structure already captured by cosine, can therefore yield zero increment. This defines the scope of the Parts~A and B MDEs: they apply to pair-specific or uniquely incremental radial structure, not to every possible radial level code.

\paragraph{Radial directed ordering (Part A).}
In a post-hoc simulation characterization, we plant pair-specific edge margins on the same $100$-pair, $20$-parent design
and use the same coarse-permutation null as
Section~\ref{sec:parent_child}. Each child is placed a margin $m$ above its own
parent anchor with additive noise. The simulation uses $500$ Monte Carlo draws
per setting and $200$ shuffles per draw. At the selected noise level
($\tau=0.005$), the signal-zero one-sided detection rate is $1.6\%$; across
the full noise grid it ranges from $1.6\%$ to $7.6\%$. Power for the
one-sided criterion $z\geq1.6$ reaches $80\%$ at
$\Delta_{\mathrm{pair}}\approx13.5$ percentage points.

For comparison, the observed null SDs of approximately $2.6$--$3.7$
percentage points imply a normal-approximation MDE of roughly $6$--$9$
points. The larger simulated MDE is therefore conservative relative to those
audited null dispersions. The power curve is non-monotonic at very large
$m$: once every child exceeds every parent, both the true and shuffled rates
approach $100\%$ and $\Delta_{\mathrm{pair}}$ returns toward zero. This
illustrates why the statistic does not detect a purely global level shift.

\paragraph{Taxonomy radial contribution (Part B).}
We calibrate the deployed manual-mapping test using $100$ CIFAR classes, $4{,}950$ valid pairs, $999$ node permutations, and the one-sided add-one Mantel $p$-value. Both the exchangeable and cosine-correlated conditional nulls pass the registered size gates for released MERU-B, HyCoCLIP-B, and PHyCLIP-B. At a planted $\Delta R^2_{\mathrm{norm}}=0.020$, all $500$ replicates are detected for each checkpoint; the model-wise \rev{one-sided $95\%$} Clopper--Pearson lower bound is $0.994$.

The descriptive MDE$_{80}$ is $0.0089$ for released MERU-B and $0.0090$ for released HyCoCLIP-B and PHyCLIP-B. Their observed increments are respectively $0.00134$, $0.00096$, and $0.00018$, all non-significant under the deployed test. These comparisons apply only to these three released checkpoints; we do not transfer the MDE or multiplier comparisons to current-GRIT rows.

The planted regressor is aligned with the cosine-orthogonal component of the deployed norm feature. This is therefore a feature-aligned sensitivity characterization, not a bound for all possible radial or depth signals.

\paragraph{Branch-relative radial presence (Part C).}
On the three released ViT-B checkpoints, we evaluate $20$ CIFAR branches after residualizing token and character length. For branch $b$, $D_b$ is the within-branch fine-minus-coarse norm contrast after residualization, with $D_b>0$ denoting a fine-over-coarse contrast. MERU-B has median $D_b=+0.0147$, a $61\%$ fine-over-coarse rate, and AUROC $0.61$; the qualitative direction is stable across four prompt templates. Holm-adjusted $p$-values are $0.0003$, $0.605$, and $0.223$ for MERU-B, HyCoCLIP-B, and PHyCLIP-B, respectively; HyCoCLIP-B points in the reverse direction ($35\%$ fine-over-coarse).

A leave-one-branch-out comparison of cross-fitted angular-only and angular-plus-norm readouts detects no gain from adding norm. Angular-only AUROC is $0.80$--$0.84$, and approximate conditional and high-dimensional sensitivity checks agree. This beyond-angle comparison is underpowered: power remains at most $0.28$ even for a planted $3$-SD effect. In contrast, the presence test reaches $80\%$ power for planted effects of $0.8$--$1.1$ SD. Thus MERU-B contains detectable branch-relative radial information, but whether it adds information beyond angle remains unresolved.

\paragraph{Interpretation.}
These analyses apply to their stated estimands. In Part~A, the largest released pair-specific excess is approximately $5$ points and the largest current-GRIT single-seed excess is $9.5$ points, versus the $13.5$-point MDE. Part~B supports only the released-checkpoint statements above; current-GRIT non-detections have no calibrated power claim. Part~C detects branch-relative radial presence for MERU-B, but its underpowered beyond-angle comparison leaves incremental norm contribution unresolved and does not establish an operative radial hierarchy. None bounds undetected effects or excludes global, angular-redundant, or feature-misaligned radial codes.

\FloatBarrier
\section{Multi-Granularity Retrieval Details}
\label{app:multigranularity}

\subsection{Query construction}

For each WordNet depth level, we construct queries from internal ImageNet ancestors that have at least five descendant leaves and at most 500 descendant leaves. \rev{We report the odd depths $1,3,5,7,9,11,13$ in Table~\ref{tab:multigranularity_full}.}
Query features are computed by averaging descendant leaf class embeddings. Averaging shrinks norms, so coarse centroid queries mechanically sit nearer the origin, and the multi-granularity comparison is accordingly read as a supervision probe, not a radial-geometry readout.
Retrieval is performed over ImageNet validation images, and we report mean average precision over queries at each depth.

\begin{table}[htbp]
\centering
\caption{
Released-checkpoint multi-granularity retrieval. Coarse-retrieval gains are
associated with the box/compositionally supervised families. Across ViT-S/B/L,
MERU remains within $0.02$ AP of Euclidean CLIP at coarse depths, whereas
HyCoCLIP and PHyCLIP improve by $0.14$--$0.15$. This is an association across
model families, not a causal isolation of box supervision. Coarse is the
unweighted mean over depths $1,3,5$ and Fine over depths $11,13$, computed
from unrounded AP. Depth-wise values are in
Table~\ref{tab:multigranularity_full}. No released HyCoCLIP-L or PHyCLIP-S
checkpoint exists.
}
\label{tab:multigranularity}
\begin{tabular}{lcc}
\toprule
Model & Coarse ($d \leq 5$) & Fine ($d \geq 11$) \\
\midrule
CLIP-S & 0.160 & 0.529 \\
CLIP-B & 0.194 & 0.567 \\
CLIP-L & 0.194 & 0.575 \\
\midrule
MERU-S & 0.180 & 0.534 \\
MERU-B & 0.179 & 0.579 \\
MERU-L & 0.203 & 0.583 \\
\midrule
HyCoCLIP-S & \textbf{0.314} & 0.567 \\
HyCoCLIP-B & \textbf{0.333} & 0.602 \\
PHyCLIP-B & \textbf{0.338} & 0.603 \\
PHyCLIP-L & \textbf{0.346} & 0.631 \\
\bottomrule
\end{tabular}
\end{table}

\begin{table}[htbp]
\centering
\caption{
Depth-wise ImageNet multi-granularity retrieval AP across released ViT-S/B/L
checkpoints. Depths $1/3/5/7/9/11/13$ contain
\rev{$1/6/11/27/17/14/6$ qualifying WordNet-ancestor queries;} at most $50$ are
evaluated per depth. Retrieval is over ImageNet validation images.
}
\label{tab:multigranularity_full}
\begin{tabular}{lccccccc}
\toprule
Model & $1$ & $3$ & $5$ & $7$ & $9$ & $11$ & $13$ \\
\midrule
CLIP-S & 0.045 & 0.169 & 0.265 & 0.221 & 0.448 & 0.568 & 0.491 \\
CLIP-B & 0.073 & 0.210 & 0.300 & 0.245 & 0.472 & 0.609 & 0.525 \\
CLIP-L & 0.057 & 0.224 & 0.303 & 0.255 & 0.488 & 0.635 & 0.514 \\
\midrule
MERU-S & 0.063 & 0.182 & 0.294 & 0.236 & 0.464 & 0.575 & 0.493 \\
MERU-B & 0.048 & 0.186 & 0.304 & 0.262 & 0.497 & 0.626 & 0.532 \\
MERU-L & 0.070 & 0.227 & 0.311 & 0.268 & 0.511 & 0.637 & 0.530 \\
\midrule
HyCoCLIP-S & 0.116 & 0.396 & 0.431 & 0.346 & 0.550 & 0.585 & 0.549 \\
HyCoCLIP-B & 0.124 & 0.417 & 0.459 & 0.379 & 0.582 & 0.628 & 0.576 \\
PHyCLIP-B & 0.114 & 0.441 & 0.460 & 0.365 & 0.571 & 0.615 & 0.591 \\
PHyCLIP-L & 0.128 & 0.438 & 0.470 & 0.391 & 0.600 & 0.656 & 0.606 \\
\bottomrule
\end{tabular}
\end{table}

\section{Depth-Supervision Diagnostics}
\label{app:depth_diagnostics}

\subsection{Pairwise depth ranking has no direct curvature gradient}

For a pairwise norm-ranking loss defined on the pre-exponential-map tangent norms $r=\lVert v\rVert_2$ (which do not depend on the curvature parameter),
\[
L_{\mathrm{depth}} = \max(0, m - (r_{\mathrm{child}}-r_{\mathrm{parent}})),
\]
the curvature parameter $c$ does not appear. Therefore
\[
\frac{\partial L_{\mathrm{depth}}}{\partial c} = 0.
\]
This explains why pairwise depth ranking cannot identify curvature by itself; a variant defined on the post-map spatial norm $\rho=\sinh(\sqrt{c}\,r)/\sqrt{c}$ couples curvature only through the radial map (Section~\ref{sec:ldepth_naive}).

\FloatBarrier
\subsection{Full c-gradient diagnostic and multi-seed collapse-phase signs}
\label{app:gradient_methods}

The main text reports the full per-loss curvature-gradient trajectory
(Figure~\ref{fig:gradient}). Here we summarize its multi-seed sign pattern and
a separate c-only depth-gradient check.

\paragraph{Multi-seed collapse-phase sign pattern.}
Table~\ref{tab:multiseed_signs} covers seeds $42$, $37$, and $23$ for MERU-B
and HyCoCLIP-B. The collapse phase ends when
$c\leq0.1001$, with probes every $250$ steps. Entailment pushes curvature down
at every collapse-phase step in all six runs. MERU-B has no defined box-level
depth probe yet shows the same sign pattern, so the shortcut does not depend
on that depth formulation.

The contrastive gradient is reliably c-up only while it carries signal:
four runs are c-up at every $c>0.5$ step, and the two lower-signal MERU runs at
$85\%$ of such steps. After the floor is reached, it decays to sign-unstable
noise of order $10^{-3}$, with c-up fractions of $56$--$74\%$.

\begin{table}[htbp]
\centering
\setlength{\tabcolsep}{3pt}
\caption{
Collapse-phase per-loss gradient signs across three ViT-B seeds. Positive
$\partial L/\partial\log c$ denotes a c-down update.
``Contr.\ c-up (all\,/\,$c{>}0.5$)'' reports all collapse-phase steps and
the high-signal sub-phase before the contrastive gradient reaches its
approximately $10^{-3}$ noise scale. Measurements follow the unmodified
full-objective trajectory with $\lambda_{\mathrm{e}}=0.2$ and are not a
general sign characterization of the contrastive objective.
}
\label{tab:multiseed_signs}
\resizebox{\linewidth}{!}{%
\begin{tabular}{llrrrrcl}
\toprule
Model & Seed & Steps & Floor & Entail c-down & Contr.\ c-up (all\,/\,${>}0.5$) & Entail${>}$contr.\ mag.\ (raw) & Depth probe \\
\midrule
MERU-B     & 42 & 30 & 7750 & $100\%$ & $70\%$ / $100\%$ & $100\%$  & none \\
MERU-B     & 37 & 30 & 7750 & $100\%$ & $70\%$ / $85\%$  & $96.7\%$ & none \\
MERU-B     & 23 & 30 & 7750 & $100\%$ & $67\%$ / $85\%$  & $96.7\%$ & none \\
HyCoCLIP-B & 42 & 29 & 7500 & $100\%$ & $79\%$ / $100\%$ & $100\%$  & counterfactual \\
HyCoCLIP-B & 37 & 29 & 7500 & $100\%$ & $90\%$ / $100\%$ & $100\%$  & counterfactual \\
HyCoCLIP-B & 23 & 28 & 7250 & $100\%$ & $82\%$ / $100\%$ & $100\%$  & counterfactual \\
\bottomrule
\end{tabular}}
\end{table}

\paragraph{Counterfactual c-only depth-gradient check.}
The c-only diagnostic of Section~\ref{sec:entailment_shortcut} uses
\[
h = \sqrt{c}\,\mathrm{stopgrad}(\rho),
\]
which preserves a direct gradient from a depth objective to curvature while blocking that objective from changing feature norms. It is not a hierarchy-learning method: it tests whether a depth signal can identify curvature once the norm-growth path is removed. MERU has no box-level supervision, so the same counterfactual is not comparable there. The implementation was verified to be additive rather than ratio-based, so $\sqrt{c}$ does not cancel, and finite differences confirm a nonzero curvature gradient. The depth gradient's sign depends on whether active pairs already satisfy $\rho_{\mathrm{child}}>\rho_{\mathrm{parent}}$.

The depth gradient is counterfactual in the collapse probes reported here.
Their training configurations contain no depth term, and the logged
depth-loss channel is identically zero across all three HyCoCLIP-B seeds. The
actual probed objective is therefore
contrastive $+\lambda_{\mathrm{e}}\,$entailment, with
$\lambda_{\mathrm{e}}=0.2$. The hypothetical
$\lambda_{\mathrm{depth}}=0.05$ from the coupled variant of
Section~\ref{sec:ldepth_naive} is used only for scale comparison.

At ViT-S, entailment exceeds the counterfactual depth gradient by
$89$--$4500\times$ across eight deterministic draws. At ViT-B, the ratio is
approximately $289$--$330\times$ across the collapse-probe seeds. The depth
gradient is c-up in all ViT-S draws and net c-up at ViT-B, but flips c-down on
roughly one third of ViT-B collapse-phase steps. Its sign is batch-dependent
and its magnitude is too small to offset the entailment pressure.

\begin{table}[htbp]
\centering
\small
\setlength{\tabcolsep}{4pt}
\caption{
Counterfactual c-only depth-gradient check at two scales. ViT-S uses eight
deterministic $16$-sample GRIT draws on the HyCoCLIP-S baseline checkpoint.
ViT-B reports collapse-phase means from the three HyCoCLIP-B trajectories of
Figure~\ref{fig:gradient}. Gradients are with respect to log curvature;
positive denotes a c-down update. Absolute magnitudes vary across draws, so
the comparison is interpreted through sign and relative scale rather than as
a population-level magnitude estimate.
}
\label{tab:c_only_impl_check}
\begin{tabular}{lll}
\toprule
Quantity & ViT-S ($8$ draws, baseline) & ViT-B ($3$ seeds, collapse) \\
\midrule
Entailment gradient sign & $8/8$ c-down & c-down, $100\%$ steps \\
$|\partial L_{\mathrm{e}}/\partial\log c|$ (raw entailment) & $0.315$--$0.551$, med.\ $0.315$ & $0.35$--$0.38$ (mean) \\
Depth gradient sign & $8/8$ c-up & net c-up; $34$--$37\%$ steps c-down \\
$|\partial L_{\mathrm{depth}}/\partial\log c|$ (raw depth) & $1.2{\times}10^{-4}$--$3.5{\times}10^{-3}$, med.\ $2.3{\times}10^{-3}$ & $(1.1$--$1.3)\times10^{-3}$ (mean) \\
$|$entailment$|\,/\,|$depth$|$ & $89$--$4500\times$, med.\ $157\times$ & $289$--$330\times$ \\
\bottomrule
\end{tabular}
\end{table}

\FloatBarrier
\section{Reproducibility Notes}
\label{app:reproducibility}

The supplementary audit suite contains the measurement scripts and raw per-run outputs underlying the diagnostic tables, together with convention documents for $\rho$, $\sqrt{c}\rho$, Lorentz distance, and cone aperture in each model family. Its manifest records the provenance of the included artifacts. The release also records checkpoint SHA-256 hashes, current-GRIT training configurations and final curvatures, reference-metric load checks, and analytic-versus-code tests for the radius and aperture formulas. Evaluation datasets and subsets are specified in the corresponding table captions.

\subsection{Lorentz fp32 numerical patch}

During reproduction, fp16 Lorentz operations under AMP caused numerical
instability and contrastive-loss collapse, most visibly in PHyCLIP's
product-space batched operations. We therefore added explicit fp32 casts to
all Lorentz operations, including the scalar-curvature versions used by MERU
and HyCoCLIP. All reported from-scratch current-GRIT runs use this patch. The
numerical failure itself is not treated as evidence for the hierarchy
conclusions.

\FloatBarrier
\subsection{MERU current-GRIT configuration compatibility}

Some MERU ViT-B checkpoints store the optional configuration keys
\texttt{curv\_min} and \texttt{curv\_max}. Evaluation accepts these keys for
backward-compatible loading; this does not alter model computations.

\FloatBarrier
\subsection{ATMG implementation caveat}

We exclude ATMG from the primary checkpoint audit because we could not
unambiguously match its public implementation to the objective stated in the
paper. ATMG is angle-based and complementary to our diagnosis, so we discuss
it as related work without attributing public-checkpoint findings to the
published method.

\subsection{External spot check: a concurrent model inheriting the cone machinery}
\label{app:uncha_spotcheck}

\rev{We apply the operating-point and cone-saturation diagnostics, together with
each trained relation's violation rate at $\eta=1.0$ (a diagnostics subset, not a
full audit), to the released ViT-S/B checkpoints of concurrent work
\citep{kim2026uncha}, which inherits the cone machinery audited here ($K=0.1$,
threshold $\eta$, $\lambda_{\mathrm{e}}=0.2$) and self-reports curvature at the clamp
floor. Measurement: 2{,}048 GRIT samples with the released configs; the ViT-S cells
match an earlier independent run of the same measurement (n=1{,}024; $\Delta u_{p50}\le0.0002$).}

\begin{table}[htbp]
\centering
\caption{\rev{UNCHA spot check: per-band operating point, cone saturation, and trained-relation violation. Bands
marked $^{*}$ are trained-relation parents; $c=0.100$ (clamp floor) in both scales,
matching the authors' own supplementary. The last column gives the violation rate at
$\eta=1.0$ of the relation(s) whose parent is that band ($^{\mathsf{b}}$box-text
parents two trained relations, bt$\to$bi\,/\,bt$\to$t; $^{\mathsf{d}}$diagnostic
reverse direction i$\to$t, not trained).}}
\label{tab:uncha_spotcheck}
\rev{
\begin{tabular}{llcccccc}
\toprule
Scale & Band & $u_{p50}$ & $H(u)$ & Distortion & vs $2K{=}0.2$ & Cone sat.\ & Viol.\ as parent [\%] \\
\midrule
ViT-S & text$^{*}$      & 0.196 & 1.0064 & 0.64\% & below & 94.1\% & 0.0 \\
      & box\_text$^{*}$ & 0.193 & 1.0061 & 0.61\% & below & 98.3\% & 0.5\,/\,99.3$^{\mathsf{b}}$ \\
      & image           & 0.392 & 1.0246 & 2.46\% & above & 0.0\%  & 100.0$^{\mathsf{d}}$ \\
      & box\_image$^{*}$& 0.391 & 1.0245 & 2.45\% & above & 0.0\%  & 100.0 \\
\midrule
ViT-B & text$^{*}$      & 0.196 & 1.0064 & 0.64\% & below & 97.0\% & 0.0 \\
      & box\_text$^{*}$ & 0.195 & 1.0063 & 0.63\% & below & 98.0\% & 0.3\,/\,99.5$^{\mathsf{b}}$ \\
      & image           & 0.408 & 1.0265 & 2.65\% & above & 0.0\%  & 100.0$^{\mathsf{d}}$ \\
      & box\_image$^{*}$& 0.408 & 1.0265 & 2.65\% & above & 0.0\%  & 100.0 \\
\bottomrule
\end{tabular}
}
\end{table}

\rev{Text-side parents sit below the $2K=0.2$ edge and are $\pi/2$-saturated; box-image parents sit above it and are unsaturated, as the parameter-free identity predicts. The image-side band medians are the highest operating points measured in this paper (median distortion up to $2.65\%$; 95th-percentile $u\le0.417$, ${\approx}2.8\%$) yet remain far below the $5\%$ marker. The spot check does not characterize the full upper tail. On the trained relations, the saturated text-side parents show near-zero violation, while the unsaturated box-image parents' finite cones contain no children, matching the non-operative pattern of the audited cohort.}

\rev{Shuffle-null directed radial checks on the same samples: ViT-S
box\_text$\to$text $z=+3.18$ (replicating at both sample sizes), box\_image$\to$image
$z=-1.55$; ViT-B $z=+0.08$ and $z=+1.32$. The one positive cell mirrors the small
native-relation directed bit we report for the audited box-supervised models
(Section~\ref{sec:parent_child}) and was not accompanied by any traversal measurement
here.}

\input{sections/10_Delta_cone_table}

%% file: sections/10_Delta_cone_table.tex
\clearpage  %
\section{Matched and Re-Paired Containment}
\label{app:delta_cone}

\begin{table}[!htbp]
\centering
\caption{\rev{Containment for matched and randomly re-paired examples ($\eta=1$; 27 comparisons). Rates use 256 GRIT examples; re-paired rates average 9,999 random re-pairings. $\Delta$ is matched minus re-paired containment, in percentage points, calculated before rounding. Positive gaps show sensitivity to pairing; they alone do not establish learned hierarchy.}}
\label{tab:delta_cone}
\begingroup
\fontsize{10}{12}\selectfont
\setlength{\tabcolsep}{4pt}
\renewcommand{\arraystretch}{1.10}
\rev{
\begin{tabular}{@{}lllrrr@{}}
\toprule
Model & Setting & Relation & Matched (\%) & Re-paired (\%) & $\Delta$ (pp) \\
\midrule
MERU-B & Released & Text $\to$ image & 35.2 & 4.7 & +30.5 \\
\addlinespace[3pt]
MERU-B & Baseline & Text $\to$ image & 96.9 & 0.7 & +96.2 \\
\addlinespace[3pt]
MERU-B & clampOff & Text $\to$ image & 98.4 & 2.7 & +95.7 \\
\midrule
HyCoCLIP-B & Released & Text $\to$ image & 99.6 & 2.8 & +96.8 \\
HyCoCLIP-B & Released & Box text $\to$ box image & 99.6 & 42.4 & +57.2 \\
HyCoCLIP-B & Released & Box image $\to$ image & 0.0 & 0.0 & 0.0\textsuperscript{$\dagger$} \\
HyCoCLIP-B & Released & Box text $\to$ text & 56.6 & 0.7 & +56.0 \\
\addlinespace[3pt]
HyCoCLIP-B & Baseline & Text $\to$ image & 100.0 & 2.7 & +97.3 \\
HyCoCLIP-B & Baseline & Box text $\to$ box image & 100.0 & 38.3 & +61.7 \\
HyCoCLIP-B & Baseline & Box image $\to$ image & 0.0 & 0.0 & 0.0\textsuperscript{$\dagger$} \\
HyCoCLIP-B & Baseline & Box text $\to$ text & 51.6 & 0.6 & +51.0 \\
\addlinespace[3pt]
HyCoCLIP-B & clampOff & Text $\to$ image & 100.0 & 100.0 & 0.0\textsuperscript{$\ddagger$} \\
HyCoCLIP-B & clampOff & Box text $\to$ box image & 100.0 & 100.0 & 0.0\textsuperscript{$\ddagger$} \\
HyCoCLIP-B & clampOff & Box image $\to$ image & 2.0 & 0.0 & +1.9 \\
HyCoCLIP-B & clampOff & Box text $\to$ text & 83.6 & 34.5 & +49.1 \\
\midrule
PHyCLIP-B & Released & Text $\to$ image & 91.5 & 35.5 & +56.0 \\
PHyCLIP-B & Released & Box text $\to$ box image & 94.0 & 57.9 & +36.1 \\
PHyCLIP-B & Released & Box image $\to$ image & 26.3 & 2.5 & +23.8 \\
PHyCLIP-B & Released & Box text $\to$ text & 62.3 & 16.3 & +46.0 \\
\addlinespace[3pt]
PHyCLIP-B & Baseline & Text $\to$ image & 90.5 & 32.2 & +58.3 \\
PHyCLIP-B & Baseline & Box text $\to$ box image & 93.9 & 55.2 & +38.6 \\
PHyCLIP-B & Baseline & Box image $\to$ image & 30.7 & 2.1 & +28.6 \\
PHyCLIP-B & Baseline & Box text $\to$ text & 61.4 & 15.4 & +46.0 \\
\addlinespace[3pt]
PHyCLIP-B & clampOff & Text $\to$ image & 99.9 & 89.8 & +10.1 \\
PHyCLIP-B & clampOff & Box text $\to$ box image & 100.0 & 97.4 & +2.6 \\
PHyCLIP-B & clampOff & Box image $\to$ image & 24.5 & 1.8 & +22.7 \\
PHyCLIP-B & clampOff & Box text $\to$ text & 72.9 & 36.9 & +36.0 \\
\bottomrule
\end{tabular}
}
\par\vspace{5pt}
\rev{\parbox{\linewidth}{\raggedright
\textsuperscript{$\dagger$} Both rates are exactly $0\%$ (two comparisons);
\textsuperscript{$\ddagger$} both are exactly $100\%$ (two comparisons).
Box text and box image denote region captions and image crops.
Baseline and clampOff are matched retrainings with the original and relaxed curvature bounds; all retrained comparisons use seed~0.
PHyCLIP rates average its component spaces.}}
\endgroup
\end{table}